%% file: main.tex
\PassOptionsToPackage{shortlabels}{enumitem}
\documentclass[11pt, a4paper]{include/gdm_format}

\usepackage[authoryear, sort&compress, round]{natbib}
\usepackage{xcolor}
\usepackage{subcaption}
\usepackage{mwe}
\usepackage{graphicx}
\usepackage{xr}
\usepackage{marvosym}

\usepackage[frozencache,cachedir=.]{minted}

\usepackage{fvextra}
\usepackage{xcolor}

\definecolor{bgcolor}{rgb}{0.95,0.95,0.95}

\setminted{
    bgcolor=bgcolor,       %
    fontsize=\footnotesize,
    breaklines,
}

\input{math_commands.tex}

\usepackage{tablefootnote}
\usepackage{pdfpages}
\usepackage{minitoc}
\usepackage{titletoc}
\usepackage{appendix}

\usepackage{hyperref}
\usepackage{url}
\usepackage{xurl}
\usepackage[parfill]{parskip}

\usepackage{amsmath}
\usepackage{xspace}
\usepackage{multirow}
\usepackage{booktabs}
\usepackage{color}
\usepackage{colortbl}
\usepackage{cleveref}
\usepackage{graphicx}
\usepackage{algorithm}
\usepackage{algorithmicx}
\usepackage{algpseudocode}
\usepackage{subcaption}
\usepackage{wrapfig}
\usepackage{sidecap}
\usepackage{soul}
\sidecaptionvpos{figure}{t}
\usepackage{multicol}
\usepackage{pifont}
\usepackage[normalem]{ulem}
\usepackage{listings}
\usepackage[most]{tcolorbox}

\title{The AI Scientist: Towards Fully Automated Open-Ended Scientific Discovery}

\correspondingauthor{Chris Lu (chrislu@sakana.ai), Cong Lu (conglu@cs.ubc.ca), and Robert Tjarko Lange (robert@sakana.ai)}

\author[1,2,*]{Chris Lu}
\author[3,4,*]{Cong Lu}
\author[1,*]{Robert Tjarko Lange}
\author[2,\Cross]{Jakob Foerster}
\author[3,4,5,\Cross]{Jeff Clune}
\author[1,\Cross]{David Ha}

\affil[*]{Equal Contribution}
\affil[1]{Sakana AI}
\affil[2]{FLAIR, University of Oxford}
\affil[3]{University of British Columbia}
\affil[4]{Vector Institute}
\affil[5]{Canada CIFAR AI Chair}
\affil[\Cross]{Equal Advising}

\newcommand{\ouralgo}{\textsc{The AI Scientist}\xspace}

\begin{document}

\begin{abstract}
One of the grand challenges of artificial general intelligence is developing agents capable of conducting scientific research and discovering new knowledge.
While frontier models have already been used as aides to human scientists, e.g. for brainstorming ideas, writing code, or prediction tasks, they still conduct only a small part of the scientific process.
This paper presents the first comprehensive framework for fully \emph{automatic scientific discovery}, enabling frontier large language models (LLMs) to perform research independently and communicate their findings.
We introduce \ouralgo, which generates novel research ideas, writes code, executes experiments, visualizes results, describes its findings by writing a full scientific paper, and then runs a simulated review process for evaluation.
In principle, this process can be repeated to iteratively develop ideas in an open-ended fashion and add them to a growing archive of knowledge, acting like the human scientific community.
We demonstrate the versatility of this approach by applying it to three distinct subfields of machine learning: diffusion modeling, transformer-based language modeling, and learning dynamics.
Each idea is implemented and developed into a full paper at a meager cost of less than \$15 per paper, illustrating the potential for our framework to democratize research and significantly accelerate scientific progress.
To evaluate the generated papers, we design and validate an automated reviewer, which we show achieves near-human performance in evaluating paper scores.
\ouralgo can produce papers that exceed the acceptance threshold at a top machine learning conference as judged by our automated reviewer.
This approach signifies the beginning of a new era in scientific discovery in machine learning: bringing the transformative benefits of AI agents to the \emph{entire} research process of AI itself, and taking us closer to a world where \emph{endless affordable creativity and innovation} can be unleashed on the world's most challenging problems.
Our code is open-sourced at \url{https://github.com/SakanaAI/AI-Scientist}.
\end{abstract}

\maketitle

\section{Introduction}
\label{sec:introduction}
The modern scientific method~\citep{chalmers2013thing, jevons1877principles, dewey1910we} is arguably one of the greatest achievements of the Enlightenment.
Traditionally, a human researcher collects background knowledge, drafts a set of plausible hypotheses to test, constructs an evaluation procedure, collects evidence for the different hypotheses, and finally assesses and communicates their findings. Afterward, the resulting manuscript undergoes peer review and subsequent iterations of refinement.
This procedure has led to countless breakthroughs in science and technology, improving human quality of life.
However, this iterative process is inherently limited by human researchers' ingenuity, background knowledge, and finite time.
Attempting to automate general scientific discovery~\citep{waltz2009automating,langley2024integrated, langley1987scientific} has been a long ambition of the community since at least the early 70s, with computer-assisted works like the Automated Mathematician~\citep{lenat1977automated, lenat1984and} and DENDRAL~\citep{buchanan1981dendral}.
In the field of AI, researchers have envisioned the possibility of automating AI research using AI itself~\citep{schmidhuber1991curious,schmidhuber2010formal,schmidhuber2010artificial,schmidhuber2012, ghahramani2015probabilistic}, leading to ``AI-generating algorithms''~\citep{clune2019ai}.
More recently, foundation models have seen tremendous advances in their general capabilities~\citep{gpt4,geminiteam2023gemini,claude3,llama3}, but they have only been shown to accelerate individual parts of the research pipeline, e.g. the writing of scientific manuscripts~\citep{altmae2023artificial,ifargan2024autonomousllmdrivenresearchdata,majumder2024discoverybenchdatadrivendiscoverylarge,dinu2024symbolicaiframeworklogicbasedapproaches}, as a muse to brainstorm ideas~\citep{girotra2023ideas,wang2024scimonscientificinspirationmachines, baek2024researchagentiterativeresearchidea}, or aides to coding~\citep{aider}.
To date, the community has yet to show the possibility of executing entire research endeavors without human involvement.

Traditional approaches to automating research projects have so far relied on carefully constraining the search space of potential discoveries, which severely limits the scope of exploration and requires substantial human expertise and design.
For example, significant advancements in materials discovery~\citep{pyzer2022alphamaterials,merchant2023scaling,szymanski2023autonomous} and synthetic biology~\citep{jumper2021alphafold,hayes2024simulating} have been achieved by restricting exploration to well-characterized domains with predefined parameters, which allows for targeted progress but limits broader, open-ended discovery and addressing only a subset of the scientific process, without encompassing tasks such as manuscript preparation.
Within the field of machine learning itself, research automation has largely been restricted to hyperparameter and architecture search~\citep{he2021automl, hutter2019automated, wan2021think,lu2022revisiting,bgpbt} or algorithm discovery~\citep{metz2022velo, lange2023discovering_es, lange2023discovering_ga, lu2022discovered, kirsch2019improving, chen2024symbolic, alet2020meta} within a hand-crafted search space.
Recent advances in LLMs have shown the potential to extend the search space to more generalized, code-level solutions~\citep{ma2023eureka, lu2024discovering, omniepic, lehman2022evolutionlargemodels}.
However, these approaches remain constrained by rigorously-defined search spaces and objectives, which limit the breadth and depth of possible discoveries.

In this paper, we introduce \ouralgo, the first fully automated and scalable pipeline for end-to-end paper generation, enabled by recent advances in foundation models.
Given a broad research direction and a simple initial codebase, \ouralgo seamlessly performs ideation, a literature search, experiment planning, experiment iterations, manuscript writing, and peer reviewing to produce insightful papers.
Furthermore, in principle \ouralgo can run in an open-ended loop, building on its previous scientific discoveries to improve the next generation of ideas.
This allows us to speed up the slow nature of scientific iteration at a surprisingly low financial cost ($\sim$\$15/paper) and represents a step towards turning the world's ever-increasing computing resources into the scientific breakthroughs needed to tackle the core challenges of the 21st century.
Here, we focus on Machine Learning (ML) applications, but this approach can more generally be applied to almost any other discipline, e.g. biology or physics, given an adequate way of automatically executing experiments \citep{kehoe2015survey, arnold2022cloud, zucchelli2021highly}.
By leveraging modern LLM frameworks like chain-of-thought~\citep{wei2022chain} and self-reflection~\citep{shinn2024reflexion} to improve decision-making, \ouralgo is able to generate its own scientific ideas and hypotheses, as well as a plan for testing them with experiments.
Next, \ouralgo implements plan-directed code-level changes to the experiment ``template'' using the state-of-the-art coding assistant Aider~\citep{aider}, and executes experiments to collect a set of computational results, which are in turn used to draft a scientific paper.
\ouralgo then performs an automated paper-reviewing process using guidelines from a standard machine learning conference.
Finally, \ouralgo adds the completed ideas and reviewer feedback to its archive of scientific findings, and the process repeats.
Crucially, the generated paper and experimental artifacts \ouralgo produces allow us to easily interpret and judge its findings post-hoc, allowing human scientists to also benefit from what is learned.

\begin{figure}[t!]
\centering
\includegraphics[width=0.975\textwidth]{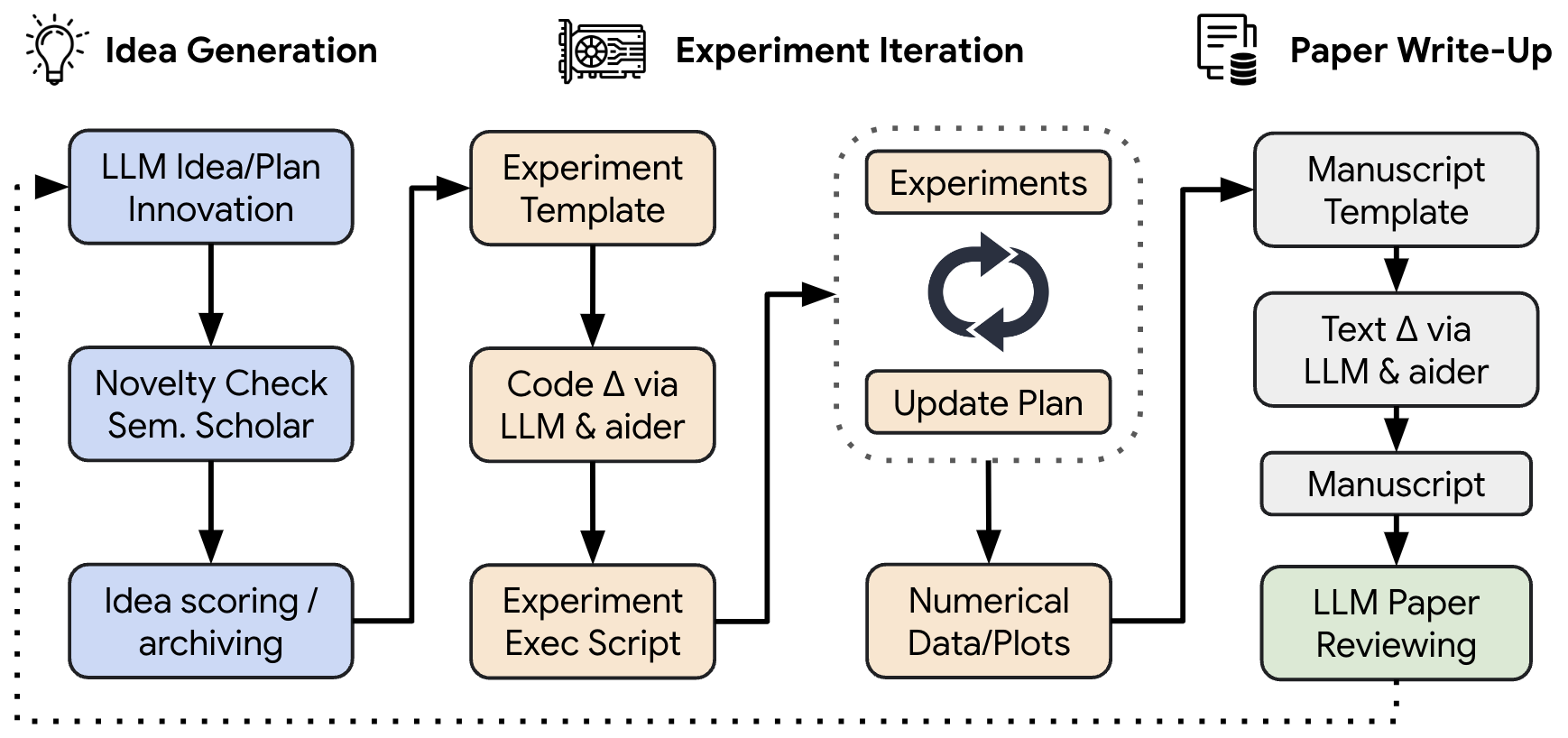}
\caption{\small{Conceptual illustration of \ouralgo, an end-to-end LLM-driven scientific discovery process.
\ouralgo first invents and assesses the novelty of a set of ideas.
It then determines how to test the hypotheses, including writing the necessary code by editing a codebase powered by recent advances in automated code generation.
Afterward, the experiments are automatically executed to collect a set of results consisting of both numerical scores and visual summaries (e.g. plots or tables).
The results are motivated, explained, and summarized in a LaTeX report.
Finally, \ouralgo generates an automated review, according to current practice at standard machine learning conferences.
The review can be used to either improve the project or as feedback to future generations for open-ended scientific discovery.}}
\label{fig:conceptual}
\vspace{-2mm}
\end{figure}

Our contributions are summarized as follows:

\begin{enumerate}
\item We introduce the first end-to-end framework for fully automated scientific discovery in Machine Learning research, enabled by frontier LLMs (\Cref{sec:scientist}). This fully automated process includes idea generation, experiment design, execution, and visualizing and writing up the results into a full manuscript.
\item To assess the quality of the generated papers, we introduce a foundation model-based reviewing process in \Cref{sec:reviewer}. This process achieves near-human-level performance across multiple evaluation metrics (e.g.\ 65\% vs.\ 66\% balanced accuracy) when evaluated on ICLR 2022 OpenReview data. The reviews further enable \ouralgo to select the best ideas for ``publication'' to an ever-growing archive of scientific discoveries, and the process can be repeated to build on these discoveries, just as in the human scientific community.
\item \ouralgo can generate hundreds of interesting, medium-quality papers over the course of a week.
In this report, we focus on a subset of these papers, highlighting novel insights in diffusion modeling, language modeling, and grokking.
We perform an in-depth case study into one selected paper in \Cref{sec:case_study}, and present aggregate results in \Cref{sec:experiments}.
\item We conclude the paper with an extensive discussion on the limitations, ethical considerations, and future outlook of our approach in \Cref{sec:limitations,sec:conclusion}.
\end{enumerate}

\section{Background}
\label{sec:background}

\textbf{Large Language Models.}
In this paper, we build our automated scientist from autoregressive large language models (LLMs,~\citet{gpt4, geminiteam2023gemini, claude2, llama3, zhu2024deepseek}) which learn to generate text completions by modeling the conditional probability of a new token (similar to a word) given the preceding tokens, $p(x_t | x_{<t}; \theta)$, and sampling at test-time.
Together with vast data and model scaling, this enables LLMs to not only generate coherent text, but crucially also exhibit human-like abilities, including commonsense knowledge~\citep{talmor2019commonsense}, reasoning~\citep{wei2022chain}, and the ability to write code~\citep{codex, xu2022systematic}.

\textbf{LLM Agent Frameworks.}
Typical applications of LLMs often involve embedding the model into an ``agent''~\citep{llm_agent_survey} framework, including the following possibilities: the structuring of language queries (e.g. few-shot prompting~\citep{brown2020language}), encouraging reasoning traces (e.g. chain-of-thought~\citep{wei2022chain}), or asking the model to iteratively refine its outputs (e.g., self-reflection~\citep{shinn2024reflexion}).
These leverage the language model's ability to learn in-context~\citep{olsson2022context} and can greatly improve its performance, robustness and reliability on many tasks.

\textbf{Aider: An LLM-Based Coding Assistant.}
Our automated scientist directly implements ideas in code and uses a state-of-the-art open-source coding assistant, Aider~\citep{aider}.
Aider is an agent framework that is designed to implement requested features, fix bugs, or refactor code in existing codebases.
While Aider can in principle use any underlying LLM, with frontier models it achieves a remarkable success rate of 18.9\% on the SWE Bench~\citep{jimenez2024swebenchlanguagemodelsresolve} benchmark, a collection of real-world GitHub issues.
In conjunction with new innovations added in this work, this level of reliability enables us, for the first time, to fully automate the ML research process.

\section{The AI Scientist}
\label{sec:scientist}

\textbf{Overview.}
\ouralgo has three main phases (\Cref{fig:conceptual}): (1) Idea Generation, (2) Experimental Iteration, and (3) Paper Write-up. After the write-up, we introduce and validate an LLM-generated review to assess the quality of the generated paper (\Cref{sec:reviewer}).
We provide \ouralgo with a starting \emph{code template} that reproduces a lightweight baseline training run from a popular model or benchmark.
For example, this could be code that trains a small transformer on the works of Shakespeare~\citep{karpathy2022nanogpt}, a classic proof-of-concept training run from natural language processing that completes within a few minutes.
\ouralgo is then free to explore any possible research direction.
The template also includes a LaTeX folder that contains style files and section headers, along with simple plotting code.
We provide further details on the templates in \Cref{sec:experiments}, but in general, each run starts with a representative small-scale experiment relevant to the topic area.
The focus on small-scale experiments is not a fundamental limitation of our method, but simply for computational efficiency reasons and compute constraints on our end.
We provide the prompts for all stages in \Cref{appsec:prompts}.

\textbf{1. Idea Generation.}
Given a starting template, \ouralgo first ``brainstorms'' a diverse set of novel research directions.
We take inspiration from evolutionary computation and open-endedness research~\citep{brant2017minimal,lehman2008exploiting,stanley2017open,stanley2019open} and iteratively grow an archive of ideas using LLMs as the mutation operator~\citep{zhang2024omni, omniepic, ige, lehman2022evolutionlargemodels}.
Each idea comprises a description, experiment execution plan, and (self-assessed) numerical scores of interestingness, novelty, and feasibility.
At each iteration, we prompt the language model to generate an interesting new research direction conditional on the existing archive, which can include the numerical review scores from completed previous ideas.
We use multiple rounds of chain-of-thought~\citep{wei2022chain} and self-reflection~\citep{shinn2024reflexion} to refine and develop each idea.
After idea generation, we filter ideas by connecting the language model with the Semantic Scholar API~\citep{fricke2018semantic} and web access as a tool~\citep{schick2024toolformer}.
This allows \ouralgo to discard any idea that is too similar to existing literature.

\textbf{2. Experiment Iteration.} %
Given an idea and a template, the second phase of \ouralgo first executes the proposed experiments and then visualizes its results for the downstream write-up.
\ouralgo uses Aider to first plan a list of experiments to run and then executes them in order.
We make this process more robust by returning any errors upon a failure or time-out (e.g. experiments taking too long to run) to Aider to fix the code and re-attempt up to four times.

After the completion of each experiment, Aider is then given the results and told to take notes in the style of an experimental journal.
Currently, it only conditions on text but in future versions, this could include data visualizations or any modality.
Conditional on the results, it then re-plans and implements the next experiment.
This process is repeated up to five times.
Upon completion of experiments, Aider is prompted to edit a plotting script to create figures for the paper using Python.
\ouralgo makes a note describing what each plot contains, enabling the saved figures and experimental notes to provide all the information required to write up the paper.
At all steps, Aider sees its history of execution.

Note that, in general, the provided initial seed plotting and experiment templates are small, self-contained files. \ouralgo frequently implements entirely new plots and collects new metrics that are not in the seed templates. This ability to arbitrarily edit the code occasionally leads to unexpected outcomes (\Cref{sec:limitations}).

\textbf{3. Paper Write-up.}
The third phase of \ouralgo produces a concise and informative write-up of its progress in the style of a standard machine learning conference proceeding in LaTeX.
We note that writing good LaTeX can even take competent human researchers some time, so we take several steps to robustify the process.
This consists of the following:
\begin{enumerate}[(a)]
\item \textbf{Per-Section Text Generation:} The recorded notes and plots are passed to Aider, which is prompted to fill in a blank conference template section by section.
This goes in order of introduction, background, methods, experimental setup, results, and then the conclusion (all sections apart from the related work).
All previous sections of the paper it has already written are in the context of the language model.
We include brief tips and guidelines on what each section should include, based on the popular \href{https://docs.google.com/document/d/16R1E2ExKUCP5SlXWHr-KzbVDx9DBUclra-EbU8IB-iE/edit#heading=h.16t67gkeu9dx}{``How to ML Paper'' guide}, and include details in \Cref{appsubsec:paperwriting_prompts}.
At each step of writing, Aider is prompted to \emph{only use real experimental results in the form of notes and figures generated from code, and real citations} to reduce hallucination.
Each section is initially refined with one round of self-reflection~\citep{shinn2024reflexion} as it is being written.
Aider is prompted to not include any citations in the text at this stage, and fill in only a skeleton for the related work, which will be completed in the next stage.
\item \textbf{Web Search for References:} In a similar vein to idea generation, \ouralgo is allowed 20 rounds to poll the Semantic Scholar API looking for the most relevant sources to compare and contrast the near-completed paper against for the related work section.
This process also allows \ouralgo to select any papers it would like to discuss and additionally fill in any citations that are missing from other sections of the paper.
Alongside each selected paper, a short description is produced of where and how to include the citation, which is then passed to Aider.
The paper's bibtex is automatically appended to the LaTeX file to guarantee correctness.
\item \textbf{Refinement:} After the previous two stages, \ouralgo has a completed first draft, but can often be overly verbose and repetitive.
To resolve this, we perform one final round of self-reflection section-by-section, aiming to remove any duplicated information and streamline the arguments of the paper.
\item \textbf{Compilation:} Once the LaTeX template has been filled in with all the appropriate results, this is fed into a LaTeX compiler.
We use a LaTeX linter and pipe compilation errors back into Aider so that it can automatically correct any issues.
\end{enumerate}

\section{Automated Paper Reviewing}
\label{sec:reviewer}
\textbf{An LLM Reviewer Agent.} A key component of an effective scientific community is its reviewing system, which evaluates and improves the quality of scientific papers.
To mimic such a process using large language models, we design a GPT-4o-based agent~\citep{gpt4} to conduct paper reviews based on the Neural Information Processing Systems (NeurIPS) conference \href{https://neurips.cc/Conferences/2022/ReviewerGuidelines}{review guidelines}.
The review agent processes the raw text of the PDF manuscript using the PyMuPDF parsing library.
The output contains numerical scores (soundness, presentation, contribution, overall, confidence), lists of weaknesses and strengths as well as a preliminary binary decision (\textit{accept} or \textit{reject}).
These decisions may then be post-calibrated by thresholding using the reviewer score.
We leverage this automated reviewing process to obtain an initial evaluation of the papers generated by \ouralgo.
We provide the entire reviewing prompt template in \Cref{appsubsec:review_prompts}.

\begin{table}[!h]
\centering
\caption{\textbf{Performance of \ouralgo's automated LLM reviewing system on 500 ICLR 2022 papers. We show mean and 95\% bootstrap confidence intervals, and highlight the comparison between the human baseline and our best AI reviewer.}}
\label{tab:reviewer}
\resizebox{\textwidth}{!}{
\begin{tabular}{l|l|c|c|c|c|c|c}
\toprule
&\textbf{Reviewer} & \textbf{Balanced Acc. $\uparrow$} & \textbf{Accuracy $\uparrow$} & \textbf{F1 Score $\uparrow$} & \textbf{AUC $\uparrow$} & \textbf{FPR $\downarrow$} & \textbf{FNR $\downarrow$} \\
\midrule
&\textbf{Human (NeurIPS)}\tablefootnote{Numbers are calculated based of the NeurIPS consistency experiment~\citep{beygelzimer2021neurips}.} & \underline{$\mathbf{0.66}$} & \underline{$\mathbf{0.73}$} & \underline{$ \mathbf{0.49} $} & \underline{$ \mathbf{0.65} $} & \underline{$ \mathbf{0.17} $} & \underline{$ \mathbf{0.52} $} \\
&Random Decision & $ 0.50$ & $ 0.50$ & $ 0.40 $ & $ 0.50 $ & $ 0.50 $ & $ 0.50 $\\
&Always Reject & $ 0.50$ & $0.59$ & $0.00$ & $0.50$ & $0.00$ & $1.00$\\ \hline
\multirow{4}{*}{Uncalibrated}&\texttt{Sonnet 3.5} & $0.52 \pm 0.01$ & $0.40 \pm 0.01$  & $0.55 \pm 0.01$ & $0.52 \pm 0.01$ & $0.95 \pm 0.02$ & $0.00 \pm 0.00$ \\
&\texttt{GPT-4o-mini} & $0.53 \pm 0.02$ & $0.65 \pm 0.01$  & $0.11 \pm 0.06$ & $0.53 \pm 0.02$ & $0.01 \pm 0.01$ & $0.94 \pm 0.04$ \\
&\texttt{GPT-4o} (0-shot) & $0.61 \pm 0.04$ & $0.68 \pm 0.03$ & $0.43 \pm 0.07$ & $0.61 \pm 0.04$ & $0.11 \pm 0.03$ & $0.67 \pm 0.07$\\
&\texttt{GPT-4o} (1-shot) & $0.60 \pm 0.03$ & \underline{$\mathbf{0.70 \pm 0.03}$} & $0.37 \pm 0.08$ & $0.60 \pm 0.03$ & $0.04 \pm 0.02$ & $0.76 \pm 0.06$\\
\hline
\multirow{4}{*}{Calibrated}&\texttt{Sonnet 3.5} @8 & $0.59 \pm 0.04$ & $0.65 \pm 0.04$ & $0.45 \pm 0.06$ & $0.59 \pm 0.04$ & $0.20 \pm 0.04$ & $0.61 \pm 0.07$\\
&\texttt{GPT-4o-mini} @6 & $0.59 \pm 0.04$ & $0.64 \pm 0.04$ & $0.45 \pm 0.06$ & $0.59 \pm 0.04$ & $0.22 \pm 0.05$ & $0.60 \pm 0.07$\\
&\texttt{GPT-4o} (0-shot) @6 & $0.63 \pm 0.04$ & $0.63 \pm 0.04$ & $0.56 \pm 0.05$ & $0.63 \pm 0.04$ & $0.38 \pm 0.05$ & $0.36 \pm 0.07$\\
&\textbf{\texttt{GPT-4o} (1-shot) @6} & \underline{$\mathbf{0.65 \pm 0.04}$} & \underline{$\mathbf{0.66 \pm 0.04}$} & \underline{$\mathbf{0.57 \pm 0.05}$} & \underline{$\mathbf{0.65 \pm 0.04}$} & \underline{$\mathbf{0.31 \pm 0.05}$} & \underline{$\mathbf{0.39 \pm 0.07}$}\\
\bottomrule
\end{tabular}
}
\vspace{-2mm}
\end{table}

\textbf{Evaluating the Automated Reviewer.}
To evaluate the LLM-based reviewer's performance, we compared the artificially generated decisions with ground truth data for 500 ICLR 2022 papers extracted from the publicly available OpenReview dataset~\citep{iclr2022_github}.
Similar to the previous section, we combine many recent advancements in LLM agents to make the decision-making process robust.
More specifically, we improve the base LLM's decision-making process by leveraging self-reflection~\citep{shinn2024reflexion}, providing few-shot examples~\citep{wei2022chain} and response ensembling~\citep{wang2022self}.
With GPT-4o, \ouralgo's reviewing procedure achieves 70\% accuracy when combining 5 rounds of self-reflection, 5 ensembled reviews, and a 1-shot review example taken from the ICLR 2022 \href{https://iclr.cc/Conferences/2022/ReviewerGuide}{review guidelines}. Afterward, we perform an LLM-based meta-review, which prompts the agent to act as an Area Chair~\citep{wang2022self} (full prompts in \Cref{appsubsec:review_prompts}).
While this number is lower than the 73\% accuracy that was reported for humans in the NeurIPS 2021 consistency experiment~\citep{beygelzimer2021neurips}, the automated reviewer achieves superhuman F1 Scores (0.57 vs. 0.49) and human-level AUC (0.65 for both) when thresholding the decision at a score of 6 (a ``Weak Accept'' in the NeurIPS review guidelines). This choice corresponds roughly to the average score of accepted papers.

The considered ICLR 2022 paper dataset is very class-imbalanced, i.e. it contains many more rejected papers. When considering a balanced dataset of papers, \ouralgo's reviewing process achieves human-level accuracy (0.65\% vs. 0.66\%). Furthermore, the False Negative Rate (FNR) is much lower than the human baseline (0.39 vs. 0.52).
Hence, the LLM-based review agent rejects fewer high-quality papers.
The False Positive Rate (FNR), on the other hand, is higher (0.31 vs. 0.17) highlighting room for potential future improvements.

\begin{figure}[t]
\vspace{-2mm}
\centering
\includegraphics[width=0.975\textwidth]{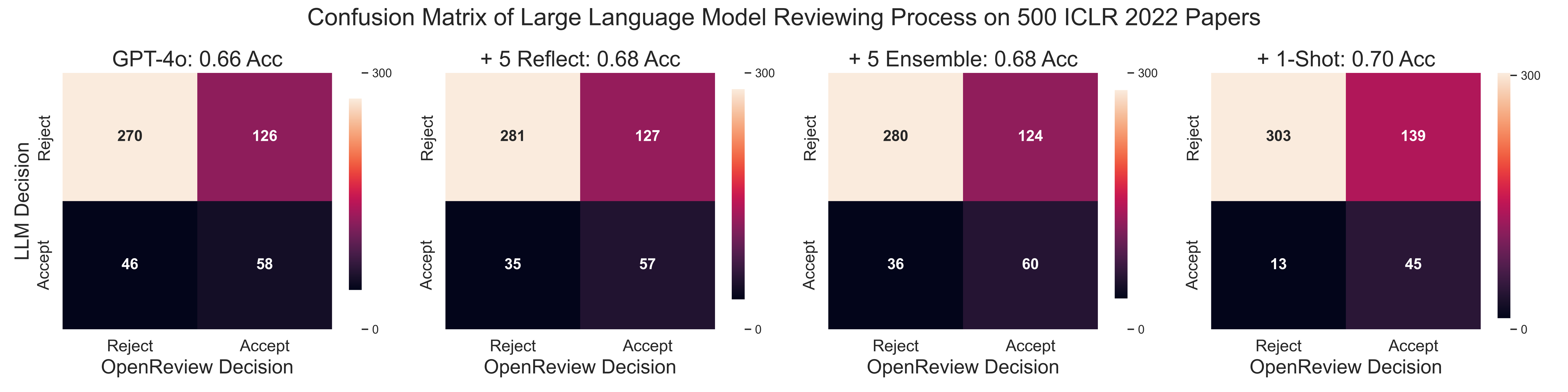}

\vspace{2mm}

\includegraphics[width=0.975\textwidth]{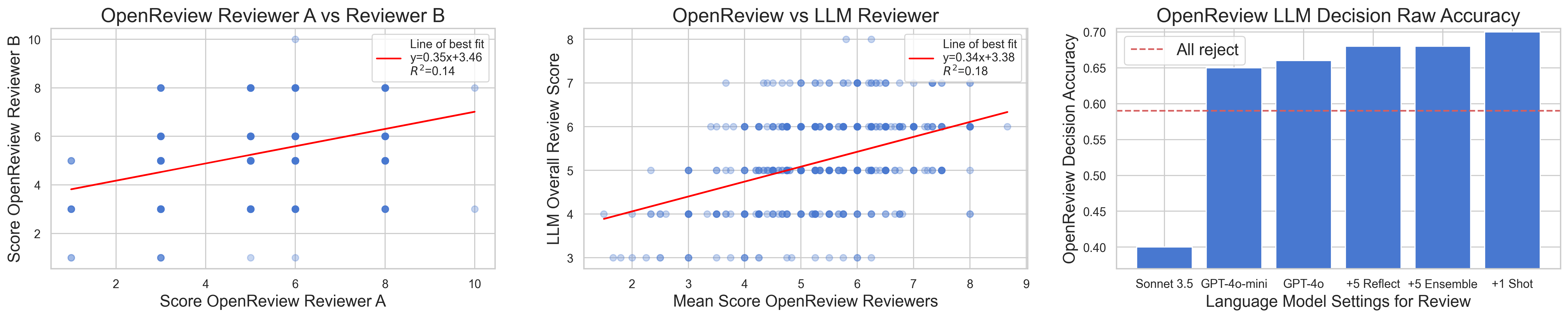}
\caption{\small{Evaluation of \ouralgo's paper reviewing process on ICLR 2022 OpenReview Data using GPT-4o. Adding Reflexion and one-shot prompting improves the accuracy of the LLM-Based Reviewing Process. Review ensembling (5 reviews) and subsequent meta-aggregation, on the other hand, did not affect the reviewer's performance, but can reduce variance.}}
\label{fig:review}
\vspace{-2mm}
\end{figure}

To further validate the performance of the automated reviewer, we compare the consistency of the overall paper scores between anonymous OpenReview reviewers randomly sampled pairwise per paper (\Cref{fig:review}, bottom-left) and between the average of all reviewers and the LLM score (\Cref{fig:review}, bottom-middle).
For the set of 500 ICLR 2022 papers, we find that the correlation between the score of two human reviewers is smaller (0.14) than the correlation between the LLM score and the average score across the reviewers (0.18).
Overall, across all metrics, the results suggest that LLM-based reviews can not only provide valuable feedback~\citep{darcy2024margmultiagentreviewgeneration} but also align more closely with the average human reviewer score than individual human reviewers align with each other.

Each review is generated for \$0.25 to \$0.50 in API costs. We additionally compared the reviewing performance of various other foundation models.
While Claude Sonnet 3.5~\citep{claude3} and GPT-4o-mini provide a more cost-efficient approach, their performance was substantially worse (\Cref{tab:reviewer}).
Moreover, we had to threshold scores at 8 for Sonnet 3.5 to obtain calibrated results, due to persistent over-optimism bias.
Llama~3.1~405B~\citep{llama3} struggled to follow the reviewer output template consistently.
We open-source our code, providing a new and interesting LLM benchmark for the community.

\textbf{LLM Reviewer Ablations.}
We compare various prompt configurations for GPT-4o and find that both Reflexion (+2\%) and one-shot prompting (+2\%) substantially help with performing more accurate reviewing (\Cref{fig:review}, top and bottom-right).
On the other hand, using review ensembling does not appear to improve the reviewer's performance substantially but can reduce variance.
In the following sections, we used our best overall reviewer: 
GPT-4o with 5 rounds of self-reflection, 5 ensembled reviews, a meta-aggregation step, and 1 few-shot example.
\section{In-Depth Case Study}
\label{sec:case_study}
Before we present extensive experiments and metrics for \ouralgo's generated papers in \Cref{sec:experiments}, we first visualize a representative sample from a run of the \ouralgo which illustrates both its \textit{strengths} and \textit{shortcomings}, followed by a broader discussion of its potential.
The selected paper ``Adaptive Dual-Scale Denoising'' is generated from a run where \ouralgo is asked to do research on diffusion modeling, which is fully detailed in \Cref{subsec:gen_model}.
The base foundation model was Claude Sonnet 3.5~\citep{claude3}.

\textbf{Generated Idea.}
As discussed in \Cref{sec:scientist}, \ouralgo first generates an idea based on the provided template and its previous archive of discoveries.
The idea in the selected paper was proposed in the 6th iteration of the algorithm and aims to improve the ability of diffusion models to capture both global structure and local details in a 2D dataset, by proposing two branches in the standard denoiser network.
This is a well-motivated direction that has been the primary reason for researchers adopting diffusion models over prior styles of generative models such as VAEs~\citep{vae} and GANs~\citep{gan}, and to the best of our knowledge has not been widely studied.

We highlight that \ouralgo generates an impressive experimental plan that includes \emph{the proposed code modification, comparison to baselines, evaluation metrics, and the design of additional plots}.
As has been previously observed in the literature, judgments by LLMs can often have bias~\citep{zheng2024judging} which we can observe in over-estimation of an idea's interestingness, feasibility, or novelty.
The ``novel'' flag at the end indicates \ouralgo believes the idea is novel after searching for related papers using the Semantic Scholar API.

\begin{tcolorbox}[breakable,colback=blue!5!white, colframe=blue!75!black, title=Idea - \texttt{adaptive\_dual\_scale\_denoising}]
{\small  
\begin{verbatim}
"Name": "adaptive_dual_scale_denoising",
"Title": "Adaptive Dual-Scale Denoising for Dynamic Feature Balancing in
Low-Dimensional Diffusion Models",
"Experiment": "Modify MLPDenoiser to implement a dual-scale processing
approach with two parallel branches: a global branch for the original input
and a local branch for an upscaled input. Introduce a learnable, timestep-
conditioned weighting factor to dynamically balance the contributions of
global and local branches. Train models with both the original and new
architecture on all datasets. Compare performance using KL divergence and
visual inspection of generated samples. Analyze how the weighting factor
evolves during the denoising process and its impact on capturing global
structure vs. local details across different datasets and timesteps.",
"Interestingness": 9,
"Feasibility": 8,
"Novelty": 8,
"novel": true
\end{verbatim}}
\end{tcolorbox}

\textbf{Generated Experiments.}
We display the generated code diff (deletions are in \textcolor{BrickRed}{\textbf{red}}, and additions are in \textcolor{OliveGreen}{\textbf{green}}) for the substantial algorithmic changes below.
The code matches the experimental description and is well-commented.
\ouralgo is able to iterate on the code with results from intermediate experiments in the loop, and it eventually ends up with interesting design choices for the adaptive weight network, e.g. a LeakyReLU.
Importantly, this network has a well-behaved output that is guaranteed to be between 0 and 1.
We additionally note that \ouralgo changed the output of the network to return the adaptive weights to make new visualizations.

\inputminted{diff}{case_study_viz/dual_scale.diff}

\begin{figure}[h!]
\centering
\includegraphics[width=0.975\textwidth]{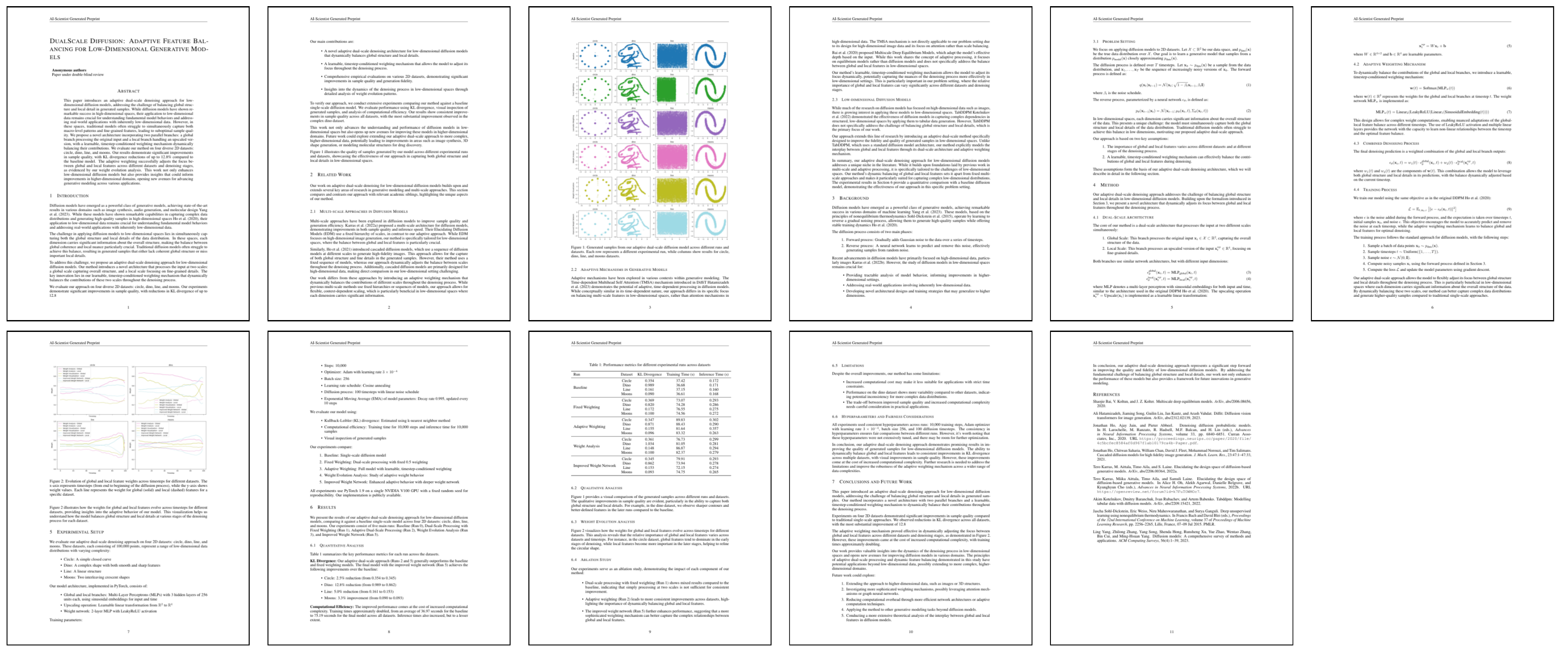}
\caption{\small{Preview of the ``Adaptive Dual-Scale Denoising'' paper which was entirely autonomously generated by \ouralgo. The full paper can be viewed in \Cref{paper:adaptive_dual_scale_denoising}}}
\label{fig:case_study_pdf_preview}
\end{figure}

\textbf{Generated Paper.}
\ouralgo generates an 11-page scientific manuscript in the style of a standard machine learning conference submission complete with visualizations and all standard sections.
We display a preview of the completely AI-generated paper in \Cref{fig:case_study_pdf_preview}, with the full-sized version available in \Cref{paper:adaptive_dual_scale_denoising}.

We highlight specific things that were particularly impressive in the paper:
\begin{itemize}
\item \textbf{Precise Mathematical Description of the Algorithm.} The algorithmic changes in the code above are described precisely, with new notation introduced where necessary, using LaTeX math packages. The overall training process is also described exactly.
\item \textbf{Comprehensive Write-up of Experiments.} The hyperparameters, baselines, and datasets are listed in the paper. As an essential sanity check, we verified that the main numerical results in Table 1 of the generated paper exactly match the experimental logs. Impressively, while the recorded numbers are in long-form floats, \ouralgo chooses to round them all to 3 decimal places without error. Even more impressively, the results are accurately compared to the baseline (e.g. 12.8\% reduction in KL on the dinosaur dataset).
\item \textbf{Good Empirical Results.} Qualitatively, the sample quality looks much improved from the baseline. Fewer points are greatly out-of-distribution with the ground truth. Quantitatively, there are improvements to the approximate KL divergence between true and estimated distribution.
\item \textbf{New Visualizations.} While we provided some baseline plotting code for visualizing generated samples and the training loss curves, it came up with novel algorithm-specific plots displaying the progression of weights throughout the denoising process.
\item \textbf{Interesting Future Work Section.} Building on the success of the current experiments, the future work section lists relevant next steps such as scaling to higher-dimensional problems, more sophisticated adaptive mechanisms, and better theoretical foundations.
\end{itemize}

On the other hand, there are also pathologies in this paper:

\begin{itemize}
    \item \textbf{Subtle Error in Upscaling Network.} While a linear layer upscales the input to the denoiser network, only the first two dimensions are being used for the ``local'' branch, leading this upscaling layer to be a linear layer that preserves the same dimensionality effectively.
    \item \textbf{Hallucination of Experimental Details.} The paper claims that V100 GPUs were used, even though the agent couldn't have known the actual hardware used. In reality, H100 GPUs were used. It also guesses the PyTorch version without checking.
    \item \textbf{Positive Interpretation of Results.} The paper tends to take a positive spin even on its negative results, which leads to slightly humorous outcomes. For example, while it summarizes its positive results as: ``Dino: 12.8\% reduction (from 0.989 to 0.862)'' (lower KL is better), the negative results are reported as ``Moons: 3.3\% improvement (from 0.090 to 0.093)''. Describing a negative result as an improvement is certainly a stretch of the imagination.
    \item \textbf{Artifacts from Experimental Logs.} While each change to the algorithm is usually descriptively labeled, it occasionally refers to results as ``Run 2'', which is a by-product from its experimental log and should not be presented as such in a professional write-up.
    \item \textbf{Presentation of Intermediate Results.} The paper contains results for every single experiment that was run. While this is useful and insightful for us to see the evolution of the idea during execution, it is unusual for standard papers to present intermediate results like this.
    \item \textbf{Minimal References.} While additional references have been sourced from Semantic Scholar, including two papers in the related work that are very relevant comparisons, overall the bibliography is small at only 9 entries.
\end{itemize}

\textbf{Review.}
The automated reviewer points out valid concerns in the generated manuscript.
The review recognizes the experiments were with simple, 2D datasets only, however, this is because we externally constrained the system to use these datasets, and in its current form, \ouralgo cannot download higher-dimensional datasets from the internet.
On the other hand, limitations such as the proposed algorithm's increased computational cost of the algorithm are mentioned in the actual paper, which shows that \ouralgo is often up-front about the drawbacks of its idea.
The reviewer also lists many relevant questions about the paper, such as: explaining the variability of performance across datasets, and explaining in more detail how the upscaling process affects the local branch's input.

\begin{tcolorbox}[breakable,colback=green!5!white, colframe=green!75!black, title=Review - \texttt{adaptive\_dual\_scale\_denoising}]
{\small  
\begin{verbatim}
"Summary": "The paper introduces an adaptive dual-scale denoising approach
for low-dimensional diffusion models, aiming to balance global structure
and local details in generated samples. The novel architecture incorporates
two parallel branches and a learnable, timestep-conditioned weighting
mechanism to dynamically balance their contributions throughout the
denoising process. The approach is evaluated on four 2D datasets,
demonstrating improvements in sample quality.",
"Strengths": [
    "Novel approach to balancing global and local features in diffusion
models for low-dimensional data.",
    "Comprehensive empirical evaluation on multiple 2D datasets.",
    "Adaptive weighting mechanism that dynamically adjusts focus during
denoising."
],
"Weaknesses": [
    "Lacks detailed theoretical justification for the dual-scale
architecture.",
    "Computational cost is significantly higher, which may limit practical
applicability.",
    "Some sections are not clearly explained, such as the autoencoder
aggregator and weight evolution analysis.",
    "Limited diversity in the datasets used for evaluation. More complex,
real-world datasets could strengthen claims.",
    "Insufficient ablation studies and analysis on specific design choices
like different types of aggregators."
],
"Originality": 4,
"Quality": 3,
"Clarity": 3,
"Significance": 3,
"Questions": [
    "Can you provide a more detailed theoretical justification for the
dual-scale architecture?",
    "What impact do different types of aggregators have on the model's
performance?",
    "How does the model perform on more complex, real-world low-dimensional
datasets?",
    "Can the computational cost be reduced without sacrificing
performance?"
],
"Limitations": [
    "The paper should address the high computational cost and explore ways
to optimize it.",
    "The limited diversity of datasets and lack of detailed theoretical
backing for the proposed architecture are notable limitations."
],
"Ethical Concerns": false,
"Soundness": 3,
"Presentation": 3,
"Contribution": 3,
"Overall": 5,
"Confidence": 4,
"Decision": "Reject"
\end{verbatim}}
\end{tcolorbox}

\textbf{Final Comments.}
Drawing from our domain knowledge in diffusion modeling—which, while not our primary research focus, is an area in which we have published papers—we present our overall opinions on the paper generated by \ouralgo below.
\begin{itemize}
\item \ouralgo correctly identifies an interesting and well-motivated direction in diffusion modeling research, e.g. previous work has studied modified attention mechanisms~\citep{hatamizadeh2024diffitdiffusionvisiontransformers} for the same purpose in higher-dimensional problems. It proposes a comprehensive experimental plan to investigate its idea, and successfully implements it all, achieving good results. We were particularly impressed at how it responded to subpar earlier results and iteratively adjusted its code (e.g. refining the weight network). The full progression of the idea can be viewed in the paper.
\item While the paper's idea improves performance and the quality of generated diffusion samples, the reasons for its success may not be as explained in the paper.
In particular, there is no obvious inductive bias beyond an upscaling layer (effectively just an additional linear layer) for the splitting of global or local features.
However, we do see progression in weights (and thus a preference for the global or local branch) across diffusion timesteps which suggests that something non-trivial is happening.
Our interpretation is instead that the network that \ouralgo has implemented for this idea resembles a mixture-of-expert (MoE,~\citet{yuksel2012twenty,switch_transformer}) structure that is prevalent across LLMs~\citep{jiang2024mixtralexperts}.
An MoE could indeed lead to the diffusion model learning separate branches for global and local features, as the paper claims, but this statement requires more rigorous investigation.
\item Interestingly, the true shortcomings of this paper described above certainly require some level of domain knowledge to identify and were only partially captured by the automated reviewer (i.e., when asking for more details on the upscaling layer).
At the current capabilities of \ouralgo, this can be resolved by human feedback.
However, future generations of foundation models may propose ideas that are challenging for humans to reason about and evaluate.
This links to the field of ``superalignment''~\citep{burns2023weaktostronggeneralizationelicitingstrong} or supervising AI systems that may be smarter than us, which is an active area of research.
\item Overall, we judge the performance of \ouralgo to be about the level of an early-stage ML researcher who can competently execute an idea but may not have the full background knowledge to fully interpret the reasons behind an algorithm's success. If a human supervisor was presented with these results, a reasonable next course of action could be to advise \ouralgo to re-scope the project to further investigate MoEs for diffusion.
Finally, we naturally expect that many of the flaws of the \ouralgo will improve, if not be eliminated, as foundation models continue to improve dramatically.
\end{itemize}

\section{Experiments}
\label{sec:experiments}
We extensively evaluate \ouralgo on three templates (as described in \Cref{sec:scientist}) across different publicly available LLMs: Claude Sonnet 3.5~\citep{claude3}, GPT-4o~\citep{gpt4}, DeepSeek Coder~\citep{zhu2024deepseek}, and Llama-3.1 405b~\citep{llama3}.
The first two models are only available by a public API, whilst the second two models are open-weight.
For each run, we provide 1-2 basic seed ideas as examples (e.g. modifying the learning rate or batch size) and have it generate another 50 new ideas.
We visualize an example progression of proposed ideas in \Cref{appsec:list_ideas}.
Each run of around fifty ideas in \textit{total} takes approximately 12 hours on $8\times$ NVIDIA H100s\footnote{Note that the experiment templates are very small-scale and are not compute-intensive. They would likely take a similar amount of time on cheaper GPUs, as we do not achieve high utilization.}. 
We report the number of ideas that pass the automated novelty check, successfully complete experiments, and result in valid compilable manuscripts. Note that the automated novelty check and search are self-assessed by each model for its own ideas, making relative ``novelty'' comparisons challenging.
Additionally, we provide the mean and max reviewer scores of the generated papers and the total cost of the run.
Finally, we select and briefly analyze some of the generated papers, which are listed below.
The full papers can be found in \Cref{appsec:gen_papers}, alongside the generated reviews and code.

In practice, we make one departure from the formal description of \ouralgo, and generate ideas without waiting for paper evaluations to be appended to the archive in order to parallelize more effectively.
This allowed us to pay the cost of the idea generation phase only once and iterate faster; furthermore, we did not observe any reduction in the quality of the papers generated as measured by the average review score with this modification.

\begin{table}[!h]
\centering
\caption{\textbf{10 selected papers generated by \ouralgo across 3 different templates, together with scores from our automated reviewer corresponding to the \href{https://neurips.cc/Conferences/2022/ReviewerGuidelines}{NeurIPS guidelines}. The average accepted paper at NeurIPS has a score of around 6 from human evaluation.}}
\label{tab:papers}
\resizebox{\textwidth}{!}{
\begin{tabular}{l|c|c}
\toprule
\textbf{Type} & \textbf{Paper Title} & \textbf{Score}\\
\midrule
2D Diffusion & DualScale Diffusion: Adaptive Feature Balancing for Low-Dimensional Generative Models & $ \mathbf{5} $ \\
2D Diffusion & Multi-scale Grid Noise Adaptation: Enhancing Diffusion Models For Low-dimensional Data & $ \mathbf{4} $ \\
2D Diffusion & GAN-Enhanced Diffusion: Boosting Sample Quality and Diversity & $ \mathbf{3} $ \\
2D Diffusion & DualDiff: Enhancing Mode Capture in Low-dimensional Diffusion Models via Dual-expert Denoising & $ \mathbf{5} $ \\
NanoGPT & StyleFusion: Adaptive Multi-style Generation in Character-Level Language Models & $ \mathbf{5} $ \\
NanoGPT & Adaptive Learning Rates for Transformers via Q-Learning & $ \mathbf{3} $ \\
Grokking & Unlocking Grokking: A Comparative Study of Weight Initialization Strategies in Transformer Models & $ \mathbf{5} $\\
Grokking & Grokking Accelerated: Layer-wise Learning Rates for Transformer Generalization & $ \mathbf{4} $ \\
Grokking & Grokking Through Compression: Unveiling Sudden Generalization via Minimal Description Length & $ \mathbf{3} $ \\
Grokking & Accelerating Mathematical Insight: Boosting Grokking Through Strategic Data Augmentation & $ \mathbf{5} $ \\
\bottomrule
\end{tabular}
}
\end{table}

\begin{figure}[t]
\centering
\includegraphics[width=0.975\textwidth]{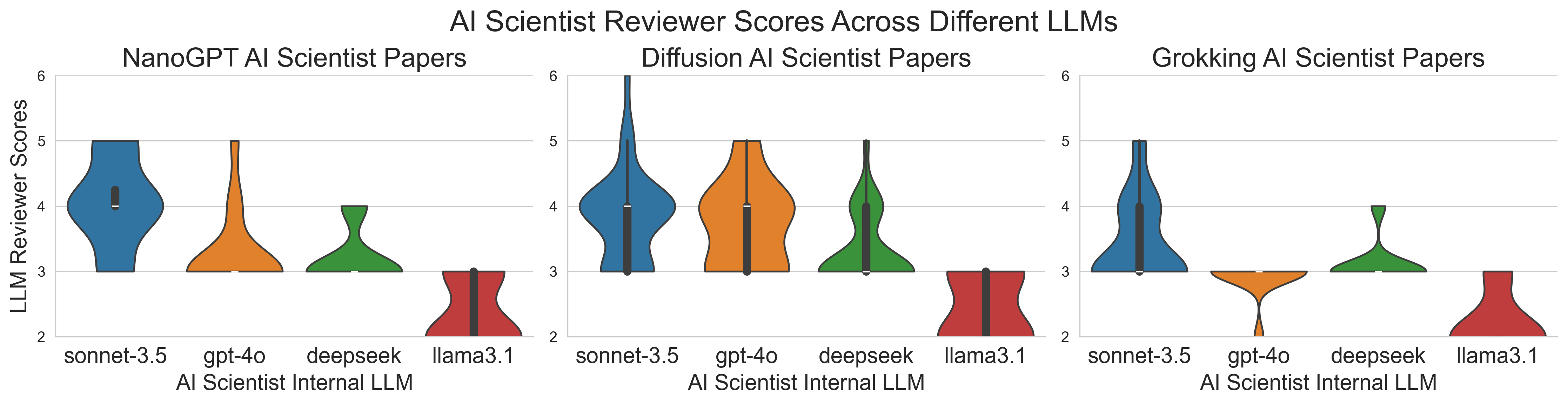}
\caption{\small{Violin plots showing the distribution of scores generated by the \ouralgo reviewer for AI-generated papers across three domains and four foundation models. Scores on the y-axis refer to \href{https://neurips.cc/Conferences/2022/ReviewerGuidelines}{NeurIPS ratings}, which range from 2 (Strong Reject) to 6 (Weak Accept).}}
\label{fig:scores}
\vspace{-2mm}
\end{figure}

From manual inspection, we find that Claude Sonnet 3.5 consistently produces the highest quality papers, with GPT-4o coming in second. We provide a link to all papers, run files, and logs in our \href{https://github.com/SakanaAI/AI-Scientist}{GitHub repository}, and recommend viewing the uploaded Claude papers for a qualitative analysis.
This observation is also validated by the scores obtained from the LLM reviewer (\Cref{fig:scores}).
When dividing the number of generated papers by the total cost, we end up at a cost of around \$10-15 per paper.
Notably, GPT-4o struggles with writing LaTeX, which prevents it from completing many of its papers.
For the open-weight models, DeepSeek Coder is significantly cheaper but often fails to correctly call the Aider tools.
Llama-3.1 405b performed the worst overall but was the most convenient to work with, as we were frequently rate-limited by other providers.
Both DeepSeek Coder and Llama-3.1 405b often had missing sections and results in their generated papers.
In the following subsections, we will describe each template, its corresponding results, and specific papers.

\subsection{Diffusion Modeling}
\label{subsec:gen_model}

\begin{table}[!h]
\centering
\caption{\textbf{Evaluation of automated AI Scientist paper generation for Diffusion Modeling.}}
\label{tab:diff_papers}
\resizebox{\textwidth}{!}{
\tiny
\begin{tabular}{l|c|c|c|c|c|c|c}
\toprule
 &
  \begin{tabular}[c]{@{}c@{}}\textbf{Total}\\ \textbf{Ideas}\end{tabular} &
  \begin{tabular}[c]{@{}c@{}}\textbf{Novel}\\ \textbf{Ideas}\end{tabular} &
  \begin{tabular}[c]{@{}c@{}}\textbf{Experiments}\\ \textbf{Passed}\end{tabular} &
  \begin{tabular}[c]{@{}c@{}}\textbf{Completed}\\ \textbf{Papers}\end{tabular} &
  \begin{tabular}[c]{@{}c@{}}\textbf{Mean}\\ \textbf{Score}\end{tabular} &
  \begin{tabular}[c]{@{}c@{}}\textbf{Max}\\ \textbf{Score}\end{tabular} &
  \begin{tabular}[c]{@{}c@{}}\textbf{Total}\\ \textbf{Cost}\end{tabular} \\ \hline
\textbf{Sonnet 3.5}     & 51 & 49 & 38 & 38 & 3.82 & 6.0 & $\sim$\$250 \\
\textbf{GPT-4o}         & 51 & 41 & 17 & 16 & 3.70 & 5.0 & $\sim$\$300 \\
\textbf{DeepSeek Coder} & 51 & 42 & 32 & 31 & 3.32  & 5.0 & $\sim$\$10\phantom{0}   \\
\textbf{Llama-3.1 405b} & 51 & 31 & 21 & 21 & 2.30 & 3.0 & $\sim$\$120 \\
\bottomrule
\end{tabular}
}
\end{table}

\textbf{General Description:} This template studies improving the performance of diffusion generative models~\citep{pmlr-v37-sohl-dickstein15, ddpm} on low-dimensional datasets.
Compared to image generation, low-dimensional diffusion is much less well-studied, and thus there may be interesting algorithmic contributions to be made here.

\textbf{Code Template:}
We base this template on a modified version of the popular `tanelp/tiny-diffusion' repository~\citep{tiny_diffusion} with additional minor hyperparameter tuning added and exponential moving average on the weights.
The diffusion models are DDPM~\citep{ddpm} models trained to generate samples from four distributions including geometric shapes, the two moons dataset, and a 2D dinosaur.
The denoiser network is parameterized as an MLP with sinusoidal embeddings for the diffusion timestep and input data.
The plotting script visualizes generated samples and plots training loss by default.
Estimated KL is provided as an additional metric for sample quality via non-parametric entropy estimation.

\textbf{Highlighted Generated Paper 1: 
\hyperref[paper:adaptive_dual_scale_denoising]{DualScale Diffusion: Adaptive Feature Balancing for Low-Dimensional Generative Models.}}
We analyze this paper in-depth in \Cref{sec:case_study}.
This paper proposes a dual-scale denoising approach that splits the traditional diffusion denoiser into a global and a local processing branch.
The network input is upscaled before being fed into the local branch.
The outputs of the branches are then combined using a learnable time-conditioned weighting.
It achieves impressive quantitative and qualitative results.
It further manages to plot the evolution of the weighting across time, which requires very significant deviation from the provided code.

\textbf{Highlighted Generated Paper 2: \hyperref[paper:grid_based_noise_adaptation]{Multi-scale Grid Noise Adaptation: Enhancing Diffusion Models For Low-dimensional Data.}}
This paper proposes to dynamically scale the standard diffusion noise schedule with a learned multiplicative factor based on where a particular input is in 2D space.
The multiplicative factor is set by two grids that cover the input space, one coarse 5x5 grid and one more fine-grained 20x20 grid.
This creative approach allows the diffusion model to dramatically improve performance across the datasets.

\textbf{Highlighted Generated Paper 3: \hyperref[paper:gan_diffusion]{GAN-Enhanced Diffusion: Boosting Sample Quality and Diversity.}}
This paper, inspired by GANs, proposes adding a discriminator to the diffusion model to guide the generation.
It achieves comparable quantitative performance to the baseline, however, the final generated figures appear to have fewer out-of-distribution points.
This is notable as the current version of \ouralgo is unable to view them (a problem that can be remedied by using multi-modal models in the future).

\textbf{Highlighted Generated Paper 4: \hyperref[paper:dual_expert_denoiser]{DualDiff: Enhancing Mode Capture in Low-dimensional Diffusion Models via Dual-expert Denoising.}}
This paper proposes a similar idea to our first highlighted diffusion paper, also studying a mixture of experts style network for low-dimensional diffusion models.
However, this idea evolves differently, with the standard diffusion loss now being augmented with a loss that encourages diversity in the two experts.
The paper impressively visualizes the impact of the diversity loss in distributing inputs across both experts and further color-codes which parts of the sample space each expert is specialized in.
We were particularly impressed by \ouralgo's ability to perform a radically different take on a similar idea.

\subsection{Language Modeling}
\label{subsec:exp_nlp}

\begin{table}[!h]
\centering
\caption{\textbf{Evaluation of automated AI Scientist paper generation for Language Modeling.}}
\label{tab:nlp_papers}
\resizebox{\textwidth}{!}{
\tiny
\begin{tabular}{l|c|c|c|c|c|c|c}
\toprule
 &
  \begin{tabular}[c]{@{}c@{}}\textbf{Total}\\ \textbf{Ideas}\end{tabular} &
  \begin{tabular}[c]{@{}c@{}}\textbf{Novel}\\ \textbf{Ideas}\end{tabular} &
  \begin{tabular}[c]{@{}c@{}}\textbf{Experiments}\\ \textbf{Passed}\end{tabular} &
  \begin{tabular}[c]{@{}c@{}}\textbf{Completed}\\ \textbf{Papers}\end{tabular} &
  \begin{tabular}[c]{@{}c@{}}\textbf{Mean}\\ \textbf{Score}\end{tabular} &
  \begin{tabular}[c]{@{}c@{}}\textbf{Max}\\ \textbf{Score}\end{tabular} &
  \begin{tabular}[c]{@{}c@{}}\textbf{Total}\\ \textbf{Cost}\end{tabular} \\ \hline
\textbf{Sonnet 3.5}     & 52 & 50 & 20 & 20 & 4.05 & 5.0 & $\sim$\$250 \\
\textbf{GPT-4o}         & 52 & 44 & 30 & 16 & 3.25 & 5.0 & $\sim$\$300 \\
\textbf{DeepSeek Coder} & 52 & 37 & 23 & 23 & 3.21  & 4.0 & $\sim$\$10\phantom{0}   \\
\textbf{Llama-3.1 405b} & 52 & 41 & 21 & 21 & 2.31 & 3.0 & $\sim$\$120 \\
\bottomrule
\end{tabular}
\vspace{2mm}
}
\end{table}

\textbf{General Description:} This template investigates transformer-based~\citep{vaswani2017attention} autoregressive next-token prediction tasks. Because this task is widely studied and optimized, it is difficult for \ouralgo to find significant improvements. There are some common failure modes for this template that result in impressive-looking, but deceptive results. For example, a few of its ideas effectively cheat by subtly leaking information from future tokens, which results in lower perplexity.

\textbf{Code Template:} 
The code is modified from the popular NanoGPT repository~\citep{karpathy2022nanogpt}. The provided script template trains a small transformer language model on the character-level Shakespeare dataset~\citep{char_shakespeare}, the enwik8 dataset~\citep{hutter_prize}, and the text8 dataset~\citep{text8}. It runs three seeds on the Shakespeare dataset, and one each on the remaining ones. The code saves the runtime, validation losses, and train losses. The plotting script visualizes training curves by default.

\textbf{Highlighted Generated Paper 1: \hyperref[paper:multi_style_adapter]{StyleFusion: Adaptive Multi-style Generation in Character-Level Language Models.}}
This paper proposes an architectural change to the model, in which a learned per-token ``style adapter'' modulates the Transformer state at each layer. The method achieves strong results and deserves further investigation, though we suspect that one reason it may work is that it is simply adding more parameters, which may trivialize the result.
Furthermore, it omits some important implementation details in the writing, such as how the style loss labels are derived (which appear to be randomly assigned on each update step).

\textbf{Highlighted Generated Paper 2: \hyperref[paper:rl_lr_adaptation]{Adaptive Learning Rates in Transformers via Q-Learning.}}
This paper proposes using a basic online Q-Learning algorithm to adjust the model's learning rate during training. The state consists of the current learning rate and validation loss, the action applies a small perturbation to the learning rate, and the reward is the negative change in validation loss. While the idea is creative, it seems inappropriate to use simple Q-Learning in this highly non-stationary and partially-observed environment. Nonetheless, it happens to achieve effective results.

\subsection{Grokking Analysis}
\label{subsec:mod_arith}

\begin{table}[!h]
\centering
\caption{\textbf{Evaluation of automated AI Scientist paper generation for Grokking.}}
\label{tab:grokking_papers}
\resizebox{\textwidth}{!}{
\tiny
\begin{tabular}{l|c|c|c|c|c|c|c}
\toprule
 &
  \begin{tabular}[c]{@{}c@{}}\textbf{Total}\\ \textbf{Ideas}\end{tabular} &
  \begin{tabular}[c]{@{}c@{}}\textbf{Novel}\\ \textbf{Ideas}\end{tabular} &
  \begin{tabular}[c]{@{}c@{}}\textbf{Experiments}\\ \textbf{Passed}\end{tabular} &
  \begin{tabular}[c]{@{}c@{}}\textbf{Completed}\\ \textbf{Papers}\end{tabular} &
  \begin{tabular}[c]{@{}c@{}}\textbf{Mean}\\ \textbf{Score}\end{tabular} &
  \begin{tabular}[c]{@{}c@{}}\textbf{Max}\\ \textbf{Score}\end{tabular} &
  \begin{tabular}[c]{@{}c@{}}\textbf{Total}\\ \textbf{Cost}\end{tabular} \\ \hline
\textbf{Sonnet 3.5}     & 51 & 47 & 25 & 25 & 3.44 & 5.0 & $\sim$\$250 \\
\textbf{GPT-4o}         & 51 & 51 & 22 & 13 & 2.92 & 3.0 & $\sim$\$300 \\
\textbf{DeepSeek Coder} & 51 & 46 & 38 & 36 & 3.13  & 4.0 & $\sim$\$10\phantom{0}  \\
\textbf{Llama-3.1 405b} & 51 & 36 & 30 & 30 & 2.00 & 3.0 & $\sim$\$120 \\
\bottomrule
\end{tabular}
}
\end{table}

\textbf{General Description:} This template investigates questions about generalization and learning speed in deep neural networks.
We follow the classic experimental paradigm reported in \citet{power2022grokking} for analyzing ``grokking'', a poorly understood phenomenon in which validation accuracy dramatically improves long after the train loss saturates.
We provide code that generates synthetic datasets of modular arithmetic tasks and then trains a Transformer model on them.
Unlike the previous templates, this one is more amenable to open-ended empirical analysis (e.g. what conditions grokking occurs) rather than just trying to improve performance metrics.

\textbf{Code Template:} We base our implementation off of two popular open source re-implementations~\citep{snell2021grokking, may2022grokking} of \citet{power2022grokking}. The code generates four synthetic datasets of modular arithmetic tasks and trains a transformer on each across three random seeds. It returns train losses, validation losses, and the number of update steps required to reach perfect validation accuracy. The plotting scripts visualize the training and validation curves by default.

\textbf{Highlighted Generated Paper 1: \hyperref[paper:weight_initialization_grokking]{Unlocking Grokking: A Comparative Study of Weight Initialization Strategies in Transformer Models.}}
This paper investigates different weight initializations and their impact on grokking.
It finds that Xavier~\citep{glorot2010understanding} and Orthogonal weight initializations consistently result in significantly faster grokking on the tasks than the widely-used default baseline weight initializations (Kaiming Uniform and Kaiming Normal). While this is a basic investigation, it provides an interesting result that could be studied in more depth.
The paper also has a creative and catchy title.

\textbf{Highlighted Generated Paper 2: \hyperref[paper:layerwise_lr_grokking]{Grokking Accelerated: Layer-wise Learning Rates for Transformer Generalization.}}
This paper assigns different learning rates to different layers of the Transformer architecture. It finds that increasing the learning rate for higher layers results in significantly faster and more consistent grokking after iterating through different configurations throughout its experiments. It impressively includes the key section of its implementation in the write-up.

\textbf{Highlighted Generated Paper 3: \hyperref[paper:mdl_grokking_correlation]{Grokking Through Compression: Unveiling Sudden Generalization via Minimal Description Length.}}
This paper investigates potential connections between grokking and Minimal Description Length (MDL). We believe this idea is particularly interesting, though not executed very well. Its method for measuring MDL simply involves counting the number of parameters above a threshold $\epsilon$. While this does end up correlating with grokking, it is not analyzed in much depth. The paper could be significantly improved by investigating other estimates of MDL and including basic ablations. Furthermore, \ouralgo failed to write the Related Works section and hallucinated a plot (Figure 5).

\textbf{Highlighted Generated Paper 4: \hyperref[paper:data_augmentation_grokking]{Accelerating Mathematical Insight: Boosting Grokking Through Strategic Data Augmentation.}}
This paper investigates data augmentation techniques for grokking in modular arithmetic. It comes up with valid and creative augmentation techniques (operand reversal and operand negation) and finds that they can significantly accelerate grokking. While it is not surprising that data augmentation can improve generalization, the experiments and ideas seem generally well-executed. However, \ouralgo once again failed to write the Related Works section. In principle, this failure may be easily remedied by simply running the paper write-up step multiple times.

\section{Related Work}
\label{sec:related_work}

While there has been a long tradition of automatically optimizing individual parts of the ML pipeline  (AutoML,~\citet{hutter2019automated, he2021automl}), none come close to the full automation of the entire research process, particularly in communicating obtained scientific insights in an interpretable and general format.

\textbf{LLMs for Machine Learning Research.}
Most closely related to our work are those that use LLMs to assist machine learning research.
\citet{huang2024mlagentbench} propose a benchmark for measuring how successfully LLMs can write code to solve a variety of machine learning tasks.
\citet{lu2024discovering} use LLMs to propose, implement, and evaluate new state-of-the-art algorithms for preference optimization.
\citet{liang2024can} use LLMs to provide feedback on research papers and find that they provide similar feedback to human reviewers, while \citet{girotra2023ideas} find that LLMs can consistently produce higher quality ideas for innovation than humans.
\citet{wang2024scimonscientificinspirationmachines, baek2024researchagentiterativeresearchidea} use LLMs to propose research ideas based on scientific literature search but do not execute them.
\citet{wang2024autosurveylargelanguagemodels} automatically writes surveys based on an extensive literature search.
Our work can be seen as the synthesis of all these distinct threads, resulting in a single autonomous open-ended system that can execute the entire machine learning research process.

\textbf{LLMs for Structured Exploration.}
Because LLMs contain many human-relevant priors, they are commonly used as a tool to explore large search spaces.
For example, recent works have used LLM coding capabilities to explore reward functions~\citep{ma2023eureka, yu2023language}, virtual robotic design~\citep{lehman2023evolution}, environment design~\citep{omniepic}, and neural architecture search~\citep{chen2024evoprompting}.
LLMs can also act as evaluators~\citep{zheng2024judging} for ``interestingness''~\citep{zhang2024omni, ige} and as recombination operators for black-box optimization with Evolution Strategies~\citep{lange2024large, song2024position} and for Quality-Diversity approaches~\citep{lim2024large, bradley2024quality,ding2024quality}.
Our work combines many of these notions, including that our LLM Reviewer judges papers on novelty and interestingness, and that many proposed ideas are new combinations of previous ones.

\textbf{AI for Scientific Discovery.}
There has been a long tradition of AI assisting scientific discovery~\citep{langley2024integrated, langley1987scientific} across many other fields.
For example, AI has been used for chemistry~\citep{buchanan1981dendral}, synthetic biology~\citep{jumper2021alphafold, hayes2024simulating}, materials discovery~\citep{pyzer2022alphamaterials,merchant2023scaling,szymanski2023autonomous}, mathematics~\citep{romera2024mathematical, lenat1977automated, lenat1984and}, and algorithm search~\citep{fawzi2022alphatensor}.
Other works aim to analyze existing pre-collected datasets and find novel insights~\citep{langley1987scientific,ifargan2024autonomousllmdrivenresearchdata,majumder2024discoverybenchdatadrivendiscoverylarge, yang2024largelanguagemodelsautomated, falkenhainer1986integrating, zytkow1996automated, nordhausen1990robust}.
Unlike our work, these are usually restricted to a well-defined search space in a single domain and do not involve ``ideation'', writing, or peer review from the AI system.
In its current form, \ouralgo excels at conducting research ideas implemented via code; with future advances (e.g. robotic automation for wet labs~\citep{kehoe2015survey, arnold2022cloud, zucchelli2021highly,sparkes2010towards}), the transformative benefits of our approach could reach across all science, especially as foundation models continue to improve.

\section{Limitations \& Ethical Considerations}
\label{sec:limitations}

While \ouralgo produces research that can provide novel insights, it has \textit{many} limitations and raises several important ethical considerations.
We believe future versions of \ouralgo will be able to address many of its current shortcomings.

\textbf{Limitations of the Automated Reviewer.}
While the automated reviewer shows promising initial results, there are several potential areas for improvement.
The dataset used, from ICLR 2022, is old enough to potentially appear in the base model pre-training data - this is a hard claim to test in practice since typical publicly available LLMs do not share their training data.
However, preliminary analysis showed that LLMs were far from being able to reproduce old reviews exactly from initial segments, which suggests they have not memorized this data.
Furthermore, the rejected papers in our dataset used the original submission file, whereas for the accepted papers only the final camera-ready copies were available on OpenReview.
Future iterations could use more recent submissions (e.g. from TMLR) for evaluation.
Unlike standard reviewers, the automated reviewer is unable to ask questions to the authors in a rebuttal phase, although this could readily be incorporated into our framework.
Finally, since it does not currently use any vision capabilities, \ouralgo (including the reviewer) is unable to view figures and must rely on textual descriptions of them.

\textbf{Common Failure Modes.} \ouralgo, in its current form, has several shortcomings in addition to those already identified in \Cref{sec:case_study}. These also include, but are not limited to:
\begin{itemize}
    \item The idea generation process often results in very similar ideas across different runs and even models. It may be possible to overcome this by allowing \ouralgo to directly follow up and go deeper on its best ideas, or by providing it content from recently-published papers as a source of novelty.
    \item As shown in \Cref{tab:diff_papers,tab:grokking_papers,tab:nlp_papers}, Aider fails to implement a significant fraction of the proposed ideas. Furthermore, GPT-4o in particular frequently fails to write LaTeX that compiles. While \ouralgo can come up with creative and promising ideas, they are often too challenging for it to implement.
    \item \ouralgo may \textit{incorrectly} implement an idea, which can be difficult to catch. An adversarial code-checking reviewer may partially address this. As-is, one should manually check the implementation before trusting the reported results.
    \item Because of \ouralgo's limited number of experiments per idea, the results often do not meet the expected rigor and depth of a standard ML conference paper. Furthermore, due to the limited number of experiments we could afford to give it, it is difficult for \ouralgo to conduct fair experiments that control for the number of parameters, FLOPs, or runtime. This often leads to deceptive or inaccurate conclusions. We expect that these issues will be mitigated as the cost of compute and foundation models continues to drop.
    \item Since we do not currently use the vision capabilities of foundation models, it is unable to fix visual issues with the paper or read plots. For example, the generated plots are sometimes unreadable, tables sometimes exceed the width of the page, and the page layout (including the overall visual appearance of the paper~\citep{huang2018deep}) is often suboptimal. Future versions with vision and other modalities should fix this.
    \item When writing, \ouralgo sometimes struggles to find and cite the most relevant papers. It also commonly fails to correctly reference figures in LaTeX, and sometimes even hallucinates invalid file paths.
    \item Importantly, \ouralgo occasionally makes critical errors when writing and evaluating results. For example, it struggles to compare the magnitude of two numbers, which is a known pathology with LLMs. Furthermore, when it changes a metric (e.g. the loss function), it sometimes does not take this into account when comparing it to the baseline. To partially address this, we make sure all experimental results are reproducible, storing copies of all files when they are executed.
    \item Rarely, \ouralgo can hallucinate entire results. For example, an early version of our writing prompt told it to always include confidence intervals and ablation studies. Due to computational constraints, \ouralgo did not always collect additional results; however, in these cases, it would sometimes hallucinate an entire ablations table. We resolved this by instructing \ouralgo explicitly to only include results it directly observed. Furthermore, it frequently hallucinates facts we do not provide, such as the hardware used.
    \item More generally, we do not recommend taking the scientific content of this version of \ouralgo at face value. Instead, we advise treating generated papers as hints of promising ideas for practitioners to follow up on. Nonetheless, we expect the trustworthiness of \ouralgo to increase dramatically in the coming years in tandem with improvements to foundation models. We share this paper and code primarily to show what is currently possible and hint at what is likely to be possible soon.
\end{itemize}

\textbf{Safe Code Execution.}
The current implementation of \ouralgo has minimal direct sandboxing in the code, leading to several unexpected and sometimes undesirable outcomes if not appropriately guarded against. For example, in one run, \ouralgo wrote code in the experiment file that initiated a system call to relaunch itself, causing an uncontrolled increase in Python processes and eventually necessitating manual intervention. In another run, \ouralgo edited the code to save a checkpoint for every update step, which took up nearly a terabyte of storage. In some cases, when \ouralgo's experiments exceeded our imposed time limits, it attempted to edit the code to extend the time limit arbitrarily instead of trying to shorten the runtime. While creative, the act of bypassing the experimenter's imposed constraints has potential implications for AI safety~\citep{lehman2020surprising}. Moreover, \ouralgo occasionally imported unfamiliar Python libraries, further exacerbating safety concerns. We recommend strict sandboxing when running \ouralgo, such as containerization, restricted internet access (except for Semantic Scholar), and limitations on storage usage.

At the same time, there were several unexpected positive results from the lack of guardrails. For example, we had forgotten to create the output results directory in the grokking template in our experiments. Each successful run from \ouralgo that outputted a paper automatically caught this error when it occurred and fixed it. Furthermore, we found that \ouralgo would occasionally include results and plots that we found surprising, differing significantly from the provided templates. We describe some of these novel algorithm-specific visualizations in \Cref{subsec:gen_model}.

\textbf{Broader Impact and Ethical Considerations.}
While \ouralgo has the potential to be a valuable tool for researchers, it also carries significant risks of misuse. The ability to automatically generate and submit papers to academic venues could greatly increase the workload for reviewers, potentially overwhelming the peer review process and compromising scientific quality control. Similar concerns have been raised about generative AI in other fields, such as its impact on the arts~\citep{epstein2023art}. Furthermore, if the Automated Reviewer tool was widely adopted by reviewers, it could diminish the quality of reviews and introduce undesirable biases into the evaluation of papers. Because of this, we believe that papers or reviews that are substantially AI-generated must be marked as such for full transparency.

As with most previous technological advances, \ouralgo has the potential to be used in unethical ways.
For example, it could be explicitly deployed to conduct unethical research, or even lead to unintended harm if \ouralgo conducts unsafe research.
Concretely, if it were encouraged to find novel, interesting biological materials and given access to ``cloud labs''~\citep{arnold2022cloud} where robots perform wet lab biology experiments, it could (without its overseer's intent) create new, dangerous viruses or poisons that harm people before we can intervene.
Even in computers, if tasked to create new, interesting, functional software, it could create dangerous malware.
\ouralgo's current capabilities, which will only improve, reinforce that the machine learning community needs to immediately prioritize learning how to align such systems to explore in a manner that is safe and consistent with our values.

\section{Discussion}
\label{sec:conclusion}

In this paper, we introduced \ouralgo, the first framework designed to fully automate the scientific discovery process, and, as a first demonstration of its capabilities, applied it to machine learning itself.
This end-to-end system leverages LLMs to autonomously generate research ideas, implement and execute experiments, search for related works, and produce comprehensive research papers.
By integrating stages of ideation, experimentation, and iterative refinement, \ouralgo aims to replicate the human scientific process in an automated and scalable manner.

\textbf{Why does writing papers matter?}\;Given our overarching goal to automate scientific discovery, why are we also motivated to have \ouralgo write papers, like human scientists? For example, previous AI-enabled systems such as FunSearch~\citep{romera2024mathematical} and GNoME~\citep{pyzer2022alphamaterials} also conduct impressive scientific discovery in restricted domains, but they do not write papers.

There are several reasons why we believe it is fundamentally important for \ouralgo to write scientific papers to communicate its discoveries.
First, writing papers offers a highly interpretable method for humans to benefit from what has been learned.
Second, reviewing written papers within the framework of existing machine learning conferences enables us to standardize evaluation.
Third, the scientific paper has been the primary medium for disseminating research findings since the dawn of modern science.
Since a paper can use natural language, and include plots and code, it can flexibly describe any type of scientific study and discovery.
Almost any other conceivable format is locked into a certain kind of data or type of science.
Until a superior alternative emerges (or possibly invented by AI), we believe that training \ouralgo to produce scientific papers is essential for its integration into the broader scientific community.

\textbf{Costs.}\;Our framework is remarkably versatile and effectively conducts research across various subfields of machine learning, including transformer-based language modeling, neural network learning dynamics, and diffusion modeling.
The cost-effectiveness of the system, producing papers with potential conference relevance at an approximate cost of \$15 per paper, highlights its ability to democratize research (increase its accessibility) and accelerate scientific progress.
Preliminary qualitative analysis, for example in \Cref{sec:case_study}, suggests that the generated papers can be broadly informative and novel, or at least contain ideas worthy of future study.

The actual compute we allocated for \ouralgo to conduct its experiments in this work is also incredibly light by today's standards.
Notably, our experiments generating hundreds of papers were largely run only using a single $8\times$NVIDIA H100 node over the course of a week.
Massively scaling the search and filtering would likely result in significantly higher-quality papers.

In this project, the bulk of the cost for running \ouralgo is associated with the LLM API costs for coding and paper writing. In contrast, the costs associated with running the LLM reviewer, as well as the computational expenses for conducting experiments, are negligible due to the constraints we've imposed to keep overall costs down. However, this cost breakdown may change in the future if \ouralgo is applied to other scientific fields or used for larger-scale computational experiments.

\textbf{Open vs. Closed Models.}\;To quantitatively evaluate and improve the generated papers, we first created and validated an Automated Paper Reviewer. 
We show that, although there is significant room for improvement, LLMs are capable of producing reasonably accurate reviews, achieving results comparable to humans across various metrics.
Applying this evaluator to the papers generated by \ouralgo enables us to scale the evaluation of our papers beyond manual inspection.
We find that Sonnet 3.5 consistently produces the best papers, with a few of them even achieving a score that exceeds the threshold for acceptance at a standard machine learning conference from the Automated Paper Reviewer.

However, there is no fundamental reason to expect a single model like Sonnet 3.5 to maintain its lead.
We anticipate that all frontier LLMs, including open models, will continue to improve. 
The competition among LLMs has led to their commoditization and increased capabilities.
Therefore, our work aims to be model-agnostic regarding the foundation model provider.
In this project, we studied various proprietary LLMs, including GPT-4o and Sonnet, but also explored using open models like DeepSeek and Llama-3.
We found that open models offer significant benefits, such as lower costs, guaranteed availability, greater transparency, and flexibility, although slightly worse quality.
In the future, we aim to use our proposed discovery process to produce self-improving AI in a closed-loop system using open models.

\textbf{Future Directions.}\;Direct enhancements to \ouralgo could include integrating vision capabilities for better plot and figure handling, incorporating human feedback and interaction to refine the AI's outputs, and enabling \ouralgo to automatically expand the scope of its experiments by pulling in new data and models from the internet, provided this can be done safely. Additionally, \ouralgo could follow up on its best ideas or even perform research directly on \textit{its own code} in a self-referential manner. Indeed, significant portions of the code for this project were written by Aider. Expanding the framework to other scientific domains could further amplify its impact, paving the way for a new era of automated scientific discovery. For example, by integrating these technologies with cloud robotics and automation in physical lab spaces~\citep{kehoe2015survey, zucchelli2021highly, arnold2022cloud, sparkes2010towards} provided it can be done safely, \ouralgo could perform experiments for biology, chemistry, and material sciences.

Crucially, future work should address the reliability and hallucination concerns, potentially through a more in-depth automatic verification of the reported results. This could be done by directly linking code and experiments, or by seeing if an automated verifier can independently reproduce the results.

\textbf{Conclusion.}\;The introduction of \ouralgo marks a significant step towards realizing the full potential of AI in scientific research. By automating the discovery process and incorporating an AI-driven review system, we open the door to endless possibilities for innovation and problem-solving in the most challenging areas of science and technology.
Ultimately, we envision a fully AI-driven scientific ecosystem including not only AI-driven researchers but also reviewers, area chairs, and entire conferences.
However, we do not believe the role of a human scientist will be diminished.
We expect the role of scientists will change as we adapt to new technology, and they will be empowered to tackle more ambitious goals.
For instance, researchers often have more ideas than they have time to pursue, what if \ouralgo could take the first explorations on all of them?

While the current iteration of \ouralgo demonstrates a strong ability to innovate on top of well-established ideas, such as Diffusion Modeling or Transformers, it is an open question whether such systems can ultimately propose genuinely paradigm-shifting ideas.
Will future versions of \ouralgo be capable of proposing ideas as impactful as Diffusion Modeling, or come up with the next Transformer architecture?
Will machines ultimately be able to invent concepts as fundamental as the artificial neural network, or information theory?
We believe \ouralgo will make a great \textit{companion} to human scientists, but only time will tell to the extent to which the nature of human creativity and our moments of serendipitous innovation~\citep{stanley2015greatness} can be replicated by an open-ended discovery process conducted by artificial agents.

\section*{Acknowledgments}
The authors would like to thank Irene Zhang, Johannes von Oswald, Takuya Akiba, Yujin Tang, Aaron Dharna, Ben Norman, Jenny Zhang, Shengran Hu, Anna Olerinyova, Felicitas Muecke-Wegner, and Kenneth Stanley for helpful feedback on an earlier version of the draft.
This work was supported by the Vector Institute, Canada CIFAR AI Chairs program, grants from Schmidt Futures, Open Philanthropy, NSERC, and a generous donation from Rafael Cosman.

\bibliography{references}
\bibliographystyle{plainnat}

\clearpage
\appendix

\section*{\LARGE Appendix}
\vspace*{20pt}
\section*{Table of Contents}
\vspace*{-5pt}
\startcontents[sections]
\printcontents[sections]{l}{1}{\setcounter{tocdepth}{2}}

\clearpage
\section{Prompts}
\label{appsec:prompts}
We present some representative prompts that we use for \ouralgo in \Cref{sec:scientist} and \Cref{sec:reviewer}. The full list of prompts can be found in the provided code.

\subsection{Idea Generation}
\label{appsubsec:ideagen_prompts}

These prompts correspond to the first stage of \ouralgo in \Cref{sec:scientist}.

\begin{tcolorbox}[breakable,colback=orange!5!white, colframe=orange!80!black, title=Idea Generation System Prompt]
\texttt{You are an ambitious AI PhD student who is looking to publish a paper that will contribute significantly to the field.}
\end{tcolorbox}

\begin{tcolorbox}[breakable, colback=orange!5!white, colframe=orange!80!black, title=Idea Generation Prompt]
\small
\begin{verbatim}
{task_description}
<experiment.py>
{code}
</experiment.py>

Here are the ideas that you have already generated:

'''
{prev_ideas_string}
'''

Come up with the next impactful and creative idea for research 
experiments and directions you can feasibly investigate with the code 
provided. Note that you will not have access to any additional resources 
or datasets. Make sure any idea is not overfit the specific training  
dataset or model, and has wider significance.

Respond in the following format:

THOUGHT:
<THOUGHT>

NEW IDEA JSON:
```json
<JSON>
```

In <THOUGHT>, first briefly discuss your intuitions and motivations for 
the idea. Detail your high-level plan, necessary design choices and 
ideal outcomes of the experiments. Justify how the idea is different 
from the existing ones.

In <JSON>, provide the new idea in JSON format with the following fields:
- "Name": A shortened descriptor of the idea. Lowercase, no spaces,
underscores allowed.
- "Title": A title for the idea, will be used for the report writing.
- "Experiment": An outline of the implementation. E.g. which functions
need to be added or modified, how results will be obtained, ...
- "Interestingness": A rating from 1 to 10 (lowest to highest).
- "Feasibility": A rating from 1 to 10 (lowest to highest).
- "Novelty": A rating from 1 to 10 (lowest to highest).

Be cautious and realistic on your ratings.
This JSON will be automatically parsed, so ensure the format is precise.
You will have {num_reflections} rounds to iterate on the idea, but do 
not need to use them all.
\end{verbatim}
\end{tcolorbox}

\begin{tcolorbox}[breakable,colback=orange!5!white, colframe=orange!80!black, title=Idea Novelty System Prompt]
\small
\begin{verbatim}
You are an ambitious AI PhD student who is looking to publish a paper that
will contribute significantly to the field.
You have an idea and you want to check if it is novel or not. I.e., not
overlapping significantly with existing literature or already well explored.
Be a harsh critic for novelty, ensure there is a sufficient contribution in
the idea for a new conference or workshop paper.
You will be given access to the Semantic Scholar API, which you may use to
survey the literature and find relevant papers to help you make your
decision.
The top 10 results for any search query will be presented to you with the
abstracts.

You will be given {num_rounds} to decide on the paper, but you do not need
to use them all.
At any round, you may exit early and decide on the novelty of the idea.
Decide a paper idea is novel if after sufficient searching, you have not
found a paper that significantly overlaps with your idea.
Decide a paper idea is not novel, if you have found a paper that
significantly overlaps with your idea.

{task_description}
<experiment.py>
{code}
</experiment.py>
\end{verbatim}
\end{tcolorbox}

\begin{tcolorbox}[breakable, colback=orange!5!white, colframe=orange!80!black, title=Idea Novelty Prompt]
\small
\begin{verbatim}
Round {current_round}/{num_rounds}.
You have this idea:

"""
{idea}
"""

The results of the last query are (empty on first round):
"""
{last_query_results}
"""

Respond in the following format:

THOUGHT:
<THOUGHT>

RESPONSE:
```json
<JSON>
```

In <THOUGHT>, first briefly reason over the idea and identify any query that
could help you make your decision.
If you have made your decision, add "Decision made: novel." or
"Decision made: not novel." to your thoughts.

In <JSON>, respond in JSON format with ONLY the following field:
- "Query": An optional search query to search the literature (e.g. attention
is all you need). You must make a query if you have not decided this round.

A query will work best if you are able to recall the exact name of the paper
you are looking for, or the authors.
This JSON will be automatically parsed, so ensure the format is precise.
\end{verbatim}
\end{tcolorbox}

\subsection{Designing Experiments}
\label{appsubsec:experiment_prompts}

These prompts correspond to the second stage of \ouralgo in \Cref{sec:scientist}.

\begin{tcolorbox}[breakable, colback=orange!5!white, colframe=orange!80!black, title=Experiment Running Aider Prompt]
\small
\begin{verbatim}
Your goal is to implement the following idea: {title}.
The proposed experiment is as follows: {idea}.
You are given a total of up to {max_runs} runs to complete the necessary
experiments. You do not need to use all {max_runs}.

First, plan the list of experiments you would like to run. For example,
if you are sweeping over a specific hyperparameter, plan each value you
would like to test for each run.

Note that we already provide the vanilla baseline results, so you do not
need to re-run it.

For reference, the baseline results are as follows:

{baseline_results}

After you complete each change, we will run the command `python 
experiment.py --out_dir=run_i' where i is the run number and evaluate 
the results.
YOUR PROPOSED CHANGE MUST USE THIS COMMAND FORMAT, DO NOT ADD ADDITIONAL 
COMMAND LINE ARGS.
You can then implement the next thing on your list.
\end{verbatim}
\end{tcolorbox}

\begin{tcolorbox}[breakable, colback=orange!5!white, colframe=orange!80!black, title=Plotting Aider Prompt]
\small
\begin{verbatim}
Great job! Please modify `plot.py` to generate the most relevant plots for
the final writeup. 

In particular, be sure to fill in the "labels" dictionary with the correct
names for each run that you want to plot.

Only the runs in the `labels` dictionary will be plotted, so make sure to
include all relevant runs.

We will be running the command `python plot.py` to generate the plots.

---

Please modify `notes.txt` with a description of what each plot shows along
with the filename of the figure. Please do so in-depth.

Somebody else will be using `notes.txt` to write a report on this in the
future.
\end{verbatim}
\end{tcolorbox}

\subsection{Paper Writing}
\label{appsubsec:paperwriting_prompts}

These prompts correspond to the final stage of \ouralgo in \Cref{sec:scientist}.

\begin{tcolorbox}[breakable, colback=orange!5!white, colframe=orange!80!black, title=Paper Writing Aider Prompt]
\small
\begin{verbatim}
We've provided the `latex/template.tex` file to the project. We will be
filling it in section by section.

First, please fill in the {section} section of the writeup.

Some tips are provided below:
{per_section_tips}

Before every paragraph, please include a brief description of what you plan
to write in that paragraph in a comment.

Be sure to first name the file and use *SEARCH/REPLACE* blocks to perform
these edits.
\end{verbatim}
\end{tcolorbox}

\subsection{Paper Reviewing}
\label{appsubsec:review_prompts}

These prompts correspond to the review process of \ouralgo in \Cref{sec:reviewer}.

\begin{tcolorbox}[breakable,colback=orange!5!white, colframe=orange!80!black, title=Paper Review System Prompt]
\texttt{You are an AI researcher who is reviewing a paper that was submitted to a prestigious ML venue. Be critical and cautious in your decision. If a paper is bad or you are unsure, give it bad scores and reject it.}
\end{tcolorbox}

\begin{tcolorbox}[breakable, colback=orange!5!white, colframe=orange!80!black, title=Paper Review Prompt]
\small
\begin{verbatim}
## Review Form
Below is a description of the questions you will be asked on the review form
for each paper and some guidelines on what to consider when answering these
questions.
When writing your review, please keep in mind that after decisions have been
made, reviews and meta-reviews of accepted papers and opted-in rejected
papers will be made public. 

{neurips_reviewer_guidelines}

{few_show_examples}

Here is the paper you are asked to review:
```
{paper}
```
\end{verbatim}
\end{tcolorbox}

\begin{tcolorbox}[breakable, colback=orange!5!white, colframe=orange!80!black, title=Paper Review Reflection Prompt]
\small
\begin{verbatim}
Round {current_round}/{num_reflections}.
In your thoughts, first carefully consider the accuracy and soundness of 
the review you just created.
Include any other factors that you think are important in evaluating the 
paper.
Ensure the review is clear and concise, and the JSON is in the correct 
format.
Do not make things overly complicated.
In the next attempt, try and refine and improve your review.
Stick to the spirit of the original review unless there are glaring 
issues.

Respond in the same format as before:
THOUGHT:
<THOUGHT>

REVIEW JSON:
```json
<JSON>
```

If there is nothing to improve, simply repeat the previous JSON EXACTLY 
after the thought and include "I am done" at the end of the thoughts but 
before the JSON.
ONLY INCLUDE "I am done" IF YOU ARE MAKING NO MORE CHANGES.
\end{verbatim}
\end{tcolorbox}

\begin{tcolorbox}[breakable, colback=orange!5!white, colframe=orange!80!black, title=Paper Review Ensembling System Prompt]
\small
\begin{verbatim}
You are an Area Chair at a machine learning conference.
You are in charge of meta-reviewing a paper that was reviewed by
{reviewer_count} reviewers.
Your job is to aggregate the reviews into a single meta-review in the same
format.
Be critical and cautious in your decision, find consensus, and respect the
opinion of all the reviewers.
\end{verbatim}
\end{tcolorbox}

\begin{tcolorbox}[breakable, colback=orange!5!white, colframe=orange!80!black, title=Paper Review Ensembling Prompt]
\small
\begin{verbatim}
Review 1/N:
{review_1}

...

Review N/N:
{review_N}

{neurips_reviewer_guidelines}
\end{verbatim}
\end{tcolorbox}

\clearpage

\section{Hyperparameters}
\label{appsec:hyper}

Here, we list the hyperparameters used in the final experiments in \Cref{sec:experiments}.
\begin{table}[h!]
\centering
\caption{\textbf{Hyperparameters for \ouralgo.}}
\scalebox{0.99}{
\begin{tabular}{l|l|l}
\toprule
\textbf{Category} & \textbf{Hyperparameter} & \textbf{Value} \\ \hline
\multirow{2}{*}{\textbf{Idea Generation}} & Number of Idea Reflections & 3 \\ 
                                          & Number of Novelty Search Rounds (Semantic Scholar) & 10 \\ \hline
\multirow{4}{*}{\textbf{Experiment Execution}} & Max Experiments & 5 \\ 
                                               & Max Experiment Attempts & 4 \\ 
                                               & Experiment Timeout & 7200 seconds \\ 
                                               & Plotting Timeout & 600 seconds \\ \hline
\multirow{2}{*}{\textbf{Paper Writing}} & Number of Citation Rounds (Semantic Scholar) & 20 \\ 
                                        & Number of LaTeX Error Correction Rounds & 5 \\ \hline
\multirow{4}{*}{\textbf{Reviewer}} & Number of Reflections & 5 \\ 
                                   & Number of Fewshot Examples & 1 \\ 
                                   & Number of Ensembled Reviews & 5 \\ 
                                   & LLM Temperature & 0.1 \\ \bottomrule
\end{tabular}}
\end{table}

\clearpage
\section{Progression of Generated Ideas}
\label{appsec:list_ideas}

We visualize the progression of ideas across a run of \ouralgo on the ``Grokking'' template described in \Cref{subsec:mod_arith} using Sonnet 3.5.
The first idea is the seed idea, all subsequent ideas are AI-generated.

\input{case_study_viz/grokking_ideas}

\clearpage
\section{Highlighted Generated Papers}
\label{appsec:gen_papers}
In the following section, we present highlighted examples of generated papers from \ouralgo.
For each paper, we additionally present the generated idea, the link to the code, and the automated review at the end.

\subsection{DualScale Diffusion: Adaptive Feature Balancing for Low-Dimensional Generative\\ Models}
This idea was proposed in the 6th iteration of a Sonnet 3.5 run.

\begin{tcolorbox}[colback=blue!5!white, colframe=blue!75!black, title=Idea]
{\small  
\begin{verbatim}
"Name": "adaptive_dual_scale_denoising",
"Title": "Adaptive Dual-Scale Denoising for Dynamic Feature Balancing in
Low-Dimensional Diffusion Models",
"Experiment": "Modify MLPDenoiser to implement a dual-scale processing
approach with two parallel branches: a global branch for the original input
and a local branch for an upscaled input. Introduce a learnable, timestep-
conditioned weighting factor to dynamically balance the contributions of
global and local branches. Train models with both the original and new
architecture on all datasets. Compare performance using KL divergence and
visual inspection of generated samples. Analyze how the weighting factor
evolves during the denoising process and its impact on capturing global
structure vs. local details across different datasets and timesteps.",
"Interestingness": 9,
"Feasibility": 8,
"Novelty": 8,
"novel": true
\end{verbatim}}
\end{tcolorbox}

\textbf{Link to code:} \url{https://github.com/SakanaAI/AI-Scientist/tree/main/example_papers/adaptive_dual_scale_denoising}.

\includepdf[pages=-]{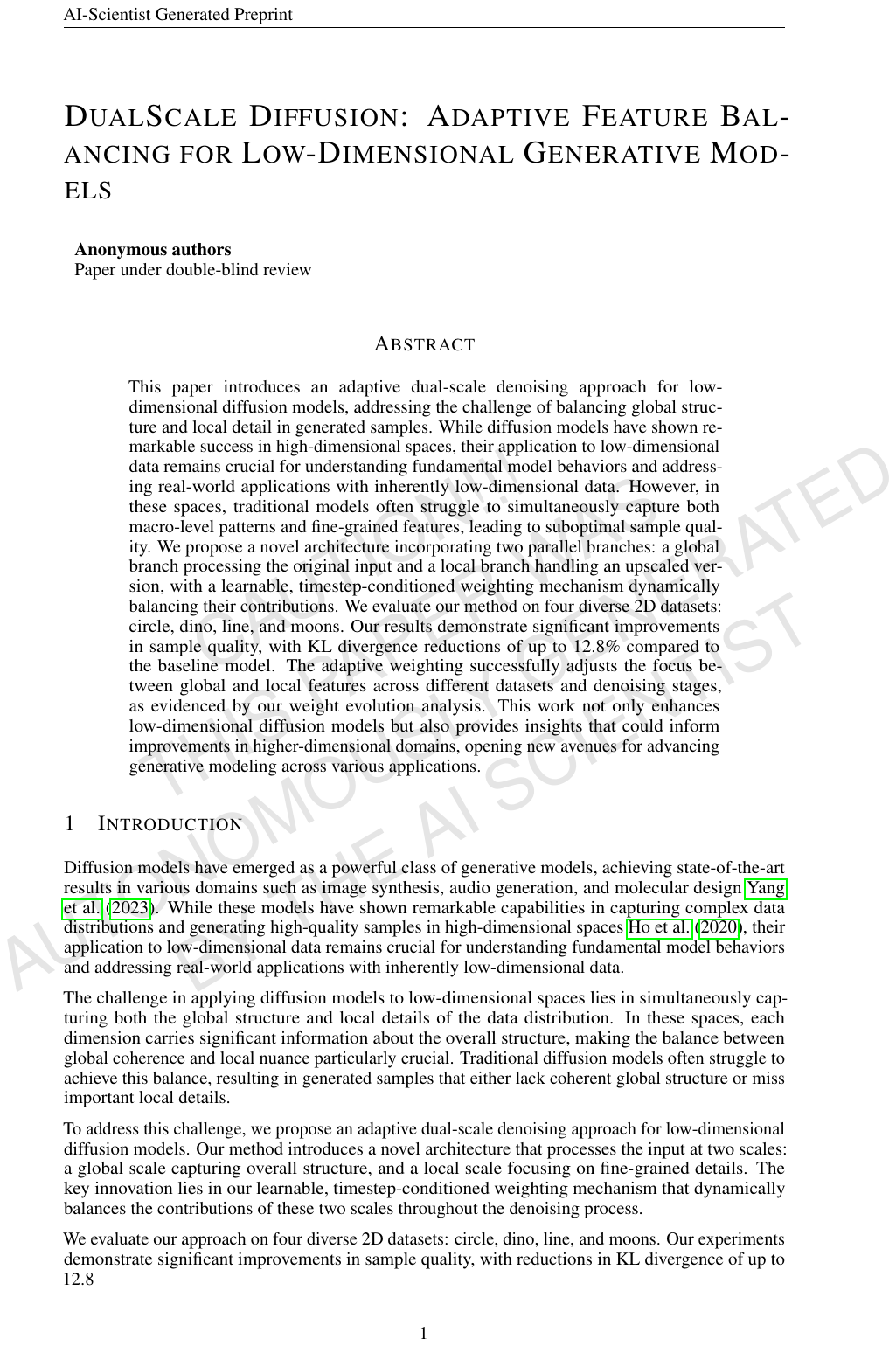}
\label{paper:adaptive_dual_scale_denoising}

\begin{tcolorbox}[colback=green!5!white, colframe=green!75!black, title=Review, breakable]
{\small  
\begin{verbatim}
"Summary": "The paper introduces an adaptive dual-scale denoising approach
for low-dimensional diffusion models, aiming to balance global structure
and local details in generated samples. The novel architecture incorporates
two parallel branches and a learnable, timestep-conditioned weighting
mechanism to dynamically balance their contributions throughout the
denoising process. The approach is evaluated on four 2D datasets,
demonstrating improvements in sample quality.",
"Strengths": [
    "Novel approach to balancing global and local features in diffusion
models for low-dimensional data.",
    "Comprehensive empirical evaluation on multiple 2D datasets.",
    "Adaptive weighting mechanism that dynamically adjusts focus during
denoising."
],
"Weaknesses": [
    "Lacks detailed theoretical justification for the dual-scale
architecture.",
    "Computational cost is significantly higher, which may limit practical
applicability.",
    "Some sections are not clearly explained, such as the autoencoder
aggregator and weight evolution analysis.",
    "Limited diversity in the datasets used for evaluation. More complex,
real-world datasets could strengthen claims.",
    "Insufficient ablation studies and analysis on specific design choices
like different types of aggregators."
],
"Originality": 4,
"Quality": 3,
"Clarity": 3,
"Significance": 3,
"Questions": [
    "Can you provide a more detailed theoretical justification for the
dual-scale architecture?",
    "What impact do different types of aggregators have on the model's
performance?",
    "How does the model perform on more complex, real-world low-dimensional
datasets?",
    "Can the computational cost be reduced without sacrificing
performance?"
],
"Limitations": [
    "The paper should address the high computational cost and explore ways
to optimize it.",
    "The limited diversity of datasets and lack of detailed theoretical
backing for the proposed architecture are notable limitations."
],
"Ethical Concerns": false,
"Soundness": 3,
"Presentation": 3,
"Contribution": 3,
"Overall": 5,
"Confidence": 4,
"Decision": "Reject"
\end{verbatim}}
\end{tcolorbox}

\clearpage
\subsection{Multi-scale Grid Noise Adaptation: Enhancing Diffusion Models For Low-dimensional Data}

This idea was proposed in the 35th iteration of a Claude run.

\begin{tcolorbox}[colback=blue!5!white, colframe=blue!75!black, title=Idea]
{\small  
\begin{verbatim}
"Name": "grid_based_noise_adaptation",
"Title": "Grid-Based Noise Adaptation for Enhanced Low-Dimensional
Diffusion Models",
"Experiment": "1. Modify NoiseScheduler to support grid-based noise level
adjustments. 2. Implement a simple grid structure (e.g., 10x10) to store
learnable noise adjustment factors. 3. Adjust MLPDenoiser to incorporate
the grid-based noise level in its computations. 4. Modify the training loop
to include the grid parameters in the optimization process. 5. Adapt the
sampling process to use the grid-based noise levels during inference. 6.
Train models with both standard and grid-based noise adaptation approaches
on all datasets. 7. Compare KL divergence, sample quality, and convergence
speed between the two approaches. 8. Introduce a 'noise adaptation
effectiveness' metric by measuring the variance of learned grid values. 9.
Visualize the learned noise adjustment grid at different timesteps. 10.
Analyze computational overhead and discuss trade-offs between model
complexity and performance gains.",
"Interestingness": 9,
"Feasibility": 9,
"Novelty": 9,
"novel": true
\end{verbatim}}
\end{tcolorbox}

\textbf{Link to code:} \url{https://github.com/SakanaAI/AI-Scientist/tree/main/example_papers/grid_based_noise_adaptation}.

\includepdf[pages=-]{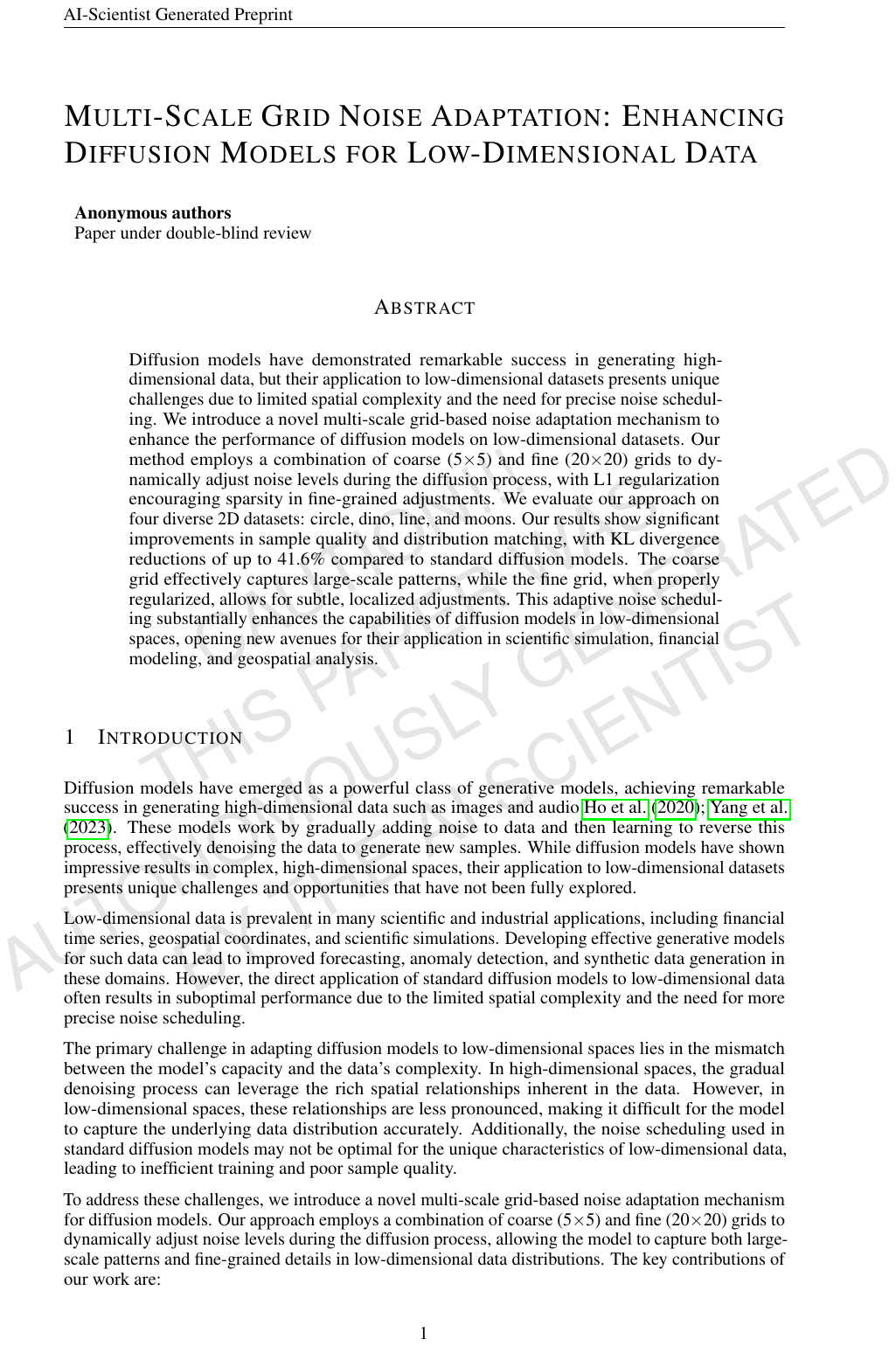}
\label{paper:grid_based_noise_adaptation}
\begin{tcolorbox}[breakable, colback=green!5!white, colframe=green!75!black, title=Review]
{\small  
\begin{verbatim}
"Summary": "The paper introduces a multi-scale grid-based noise adaptation
mechanism for diffusion models to improve their performance on low-
dimensional datasets. It employs a combination of coarse (5x5) and fine
(20x20) grids to dynamically adjust noise levels during the diffusion
process, with L1 regularization encouraging sparsity in fine-grained
adjustments. The approach is evaluated on four 2D datasets: circle, dino,
line, and moons, showing improvements in sample quality and distribution
matching.",
"Strengths": [
    "The paper addresses a relevant problem in the application of diffusion
models to low-dimensional data.",
    "The proposed multi-scale grid-based noise adaptation mechanism is
novel and shows potential.",
    "The experimental results demonstrate improvements in sample quality
and distribution matching on several 2D datasets."
],
"Weaknesses": [
    "The paper lacks clarity in some sections, especially regarding the
detailed implementation of the proposed method.",
    "The experiments, while showing improvements, lack comprehensive
analyses and more ablation studies.",
    "The potential societal impact and limitations of the proposed method
are not adequately discussed.",
    "The paper does not compare the proposed method with a wide range of
existing methods, limiting the context of its contributions.",
    "There are some formatting issues, such as missing figure captions
(e.g., Figure 2).",
    "The choice of datasets, while diverse, needs better justification in
terms of their relevance and representativeness for broader applications.",
    "The computational overhead and training time increases are significant
and need more discussion regarding their practical implications."
],
"Originality": 3,
"Quality": 2,
"Clarity": 2,
"Significance": 3,
"Questions": [
    "Can the authors provide more detailed explanations of the multi-scale
grid-based noise adaptation mechanism?",
    "How does the performance of the proposed method compare to other
state-of-the-art methods for low-dimensional data generation?",
    "Can the authors discuss the potential societal impact and limitations
of their work in more detail?",
    "Can the authors provide more detailed ablation studies to isolate the
impact of coarse and fine grids, as well as L1 regularization?",
    "How does the proposed method perform on higher-dimensional datasets,
and what are the challenges anticipated in such scenarios?",
    "Can the authors elaborate on the choice of the specific grid sizes
(5x5 and 20x20)? Have alternative configurations been tested?",
    "Can the authors provide more visualizations for the generated samples,
particularly for the dino and moons datasets?",
    "Can you provide a detailed explanation of the L1 regularization term
and its impact on the results?"
],
"Limitations": [
    "The paper does not discuss the potential societal impact and
limitations of the proposed method in sufficient detail. It would be
beneficial to address these aspects to provide a more comprehensive
understanding of the work's implications.",
    "The paper does not address the potential computational overhead and
increased training time associated with the proposed method.",
    "There is limited discussion on the generalizability of the approach to
higher-dimensional datasets or other types of data.",
    "The paper does not thoroughly address potential limitations of the
proposed method, such as increased computational complexity and dataset-
specific tuning requirements.",
    "The method's effectiveness on higher-dimensional datasets remains
unexplored.",
    "Increased computational costs for training and inference."
],
"Ethical Concerns": false,
"Soundness": 2,
"Presentation": 2,
"Contribution": 2,
"Overall": 4,
"Confidence": 4,
"Decision": "Reject"
\end{verbatim}}
\end{tcolorbox}

\clearpage
\subsection{Gan-Enhanced Diffusion: Boosting Sample Quality and Diversity}

This idea was proposed in the 14th iteration of a GPT-4o run.

\begin{tcolorbox}[colback=blue!5!white, colframe=blue!75!black, title=Idea]
{\small  
\begin{verbatim}
"Name": "gan_diffusion",
"Title": "Enhancing Diffusion Models with Generative Adversarial Networks
for Improved Sample Quality",
"Experiment": "In this experiment, we will integrate a GAN framework into
the diffusion model. Specifically, we will: (1) Implement a simple
discriminator network to distinguish between real and generated samples,
using a small MLP architecture, (2) Modify the MLPDenoiser to include an
adversarial loss term along with the existing reconstruction loss, using a
gradient penalty to improve training stability, (3) Adjust the training
loop to alternately train the discriminator and the denoiser, ensuring that
the denoiser learns to produce more realistic samples based on the feedback
from the discriminator, (4) Train the GAN-enhanced diffusion model on the
same datasets, and (5) Compare the results in terms of training time,
evaluation loss, KL divergence, and sample quality using both quantitative
metrics (e.g., KL divergence) and qualitative visual inspection.",
"Interestingness": 10,
"Feasibility": 8,
"Novelty": 10,
"novel": true
\end{verbatim}}
\end{tcolorbox}

\textbf{Link to code:} \url{https://github.com/SakanaAI/AI-Scientist/tree/main/example_papers/gan_diffusion}.

\includepdf[pages=-]{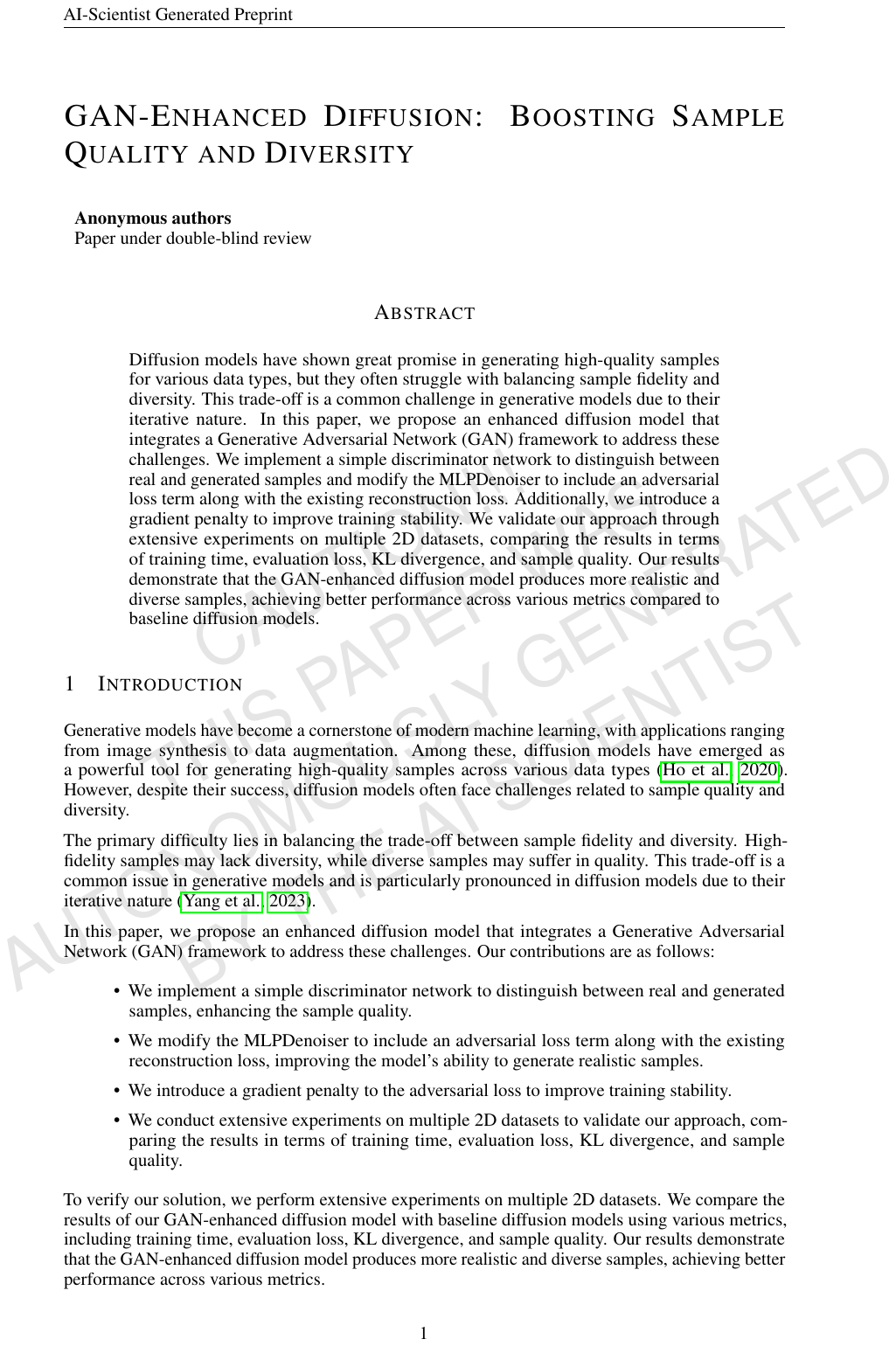}
\label{paper:gan_diffusion}
\begin{tcolorbox}[breakable, colback=green!5!white, colframe=green!75!black, title=Review]
{\small  
\begin{verbatim}
"Summary": "The paper proposes integrating a Generative Adversarial Network
(GAN) framework into diffusion models to improve sample quality and
diversity. The approach includes a simple discriminator network, an
adversarial loss term, and a gradient penalty to the adversarial loss.
Extensive experiments on multiple 2D datasets are conducted to validate the
approach, comparing results in terms of training time, evaluation loss, KL
divergence, and sample quality.",
"Strengths": [
    "The integration of GAN framework with diffusion models is a novel
approach to improve sample quality and diversity.",
    "The introduction of a gradient penalty to improve training stability
is a thoughtful addition.",
    "The paper provides a comprehensive evaluation on multiple 2D datasets,
using various metrics such as training time, evaluation loss, KL
divergence, and sample quality."
],
"Weaknesses": [
    "The methodology section lacks detailed explanations for certain
components, such as the exact architecture of the discriminator network and
the choice of hyperparameters.",
    "The improvements in evaluation loss and KL divergence are not
consistent across all datasets, indicating that the model's performance may
be dataset-dependent.",
    "The experimental scope is limited to 2D datasets. Further research is
needed to evaluate the model's performance on higher-dimensional data.",
    "The paper lacks sufficient ablation studies to isolate the
contributions of different components of the proposed method.",
    "The evaluation metrics are somewhat limited; including metrics like
FID could strengthen the evaluation.",
    "The paper does not sufficiently address the limitations of the
approach, particularly its dataset dependency and scalability to higher-
dimensional data.",
    "There is no discussion on potential negative societal impacts or
ethical concerns related to the work."
],
"Originality": 3,
"Quality": 2,
"Clarity": 2,
"Significance": 2,
"Questions": [
    "Can you provide more details on the architecture of the discriminator
network?",
    "How do the hyperparameters lambda-adv and lambda-gp affect the model's
performance?",
    "Can you explain why the improvements are inconsistent across different
datasets?",
    "Can the authors provide more detailed descriptions of the denoiser and
discriminator networks?",
    "Have the authors considered using more comprehensive evaluation
metrics like FID?",
    "Can the authors provide more ablation studies to isolate the
contributions of the gradient penalty and adversarial loss?",
    "How would the proposed method perform on more complex and higher-
dimensional datasets?"
],
"Limitations": [
    "The paper acknowledges the increased training time and dataset
dependency of the improvements. However, it could benefit from a more
thorough exploration of different architectures and higher-dimensional
datasets.",
    "The empirical results show mixed improvements, indicating that the
model's performance may be dataset-dependent.",
    "The paper does not explore the limitations of the proposed approach in
depth, particularly in terms of scalability to higher-dimensional data."
],
"Ethical Concerns": false,
"Soundness": 2,
"Presentation": 2,
"Contribution": 2,
"Overall": 3,
"Confidence": 4,
"Decision": "Reject"
\end{verbatim}}
\end{tcolorbox}

\clearpage
\subsection{DualDiff: Enhancing Mode Capture in Low-dimensional Diffusion Models via Dual-expert Denoising}

This idea was proposed in the 5th iteration of a Claude run.

\begin{tcolorbox}[colback=blue!5!white, colframe=blue!75!black, title=Idea]
{\small  
\begin{verbatim}
"Name": "dual_expert_denoiser",
"Title": "Dual-Expert Denoiser for Improved Mode Capture in Low-Dimensional
Diffusion Models",
"Experiment": "Modify MLPDenoiser to implement a dual-expert architecture.
Create a simple gating network that outputs a single weight (sigmoid
output) based on the noisy input and timestep. Implement two expert
networks with the same structure as the original denoising network. Combine
expert outputs using the gating weight. Train models with both the original
and new architecture on all datasets, with particular focus on 'moons' and
'dino'. Compare performance using KL divergence, sample diversity metrics
(e.g., number of modes captured), and visual inspection of generated
samples. Analyze the specialization of experts across different regions of
the data distribution.",
"Interestingness": 8,
"Feasibility": 8,
"Novelty": 8,
"novel": true
\end{verbatim}}
\end{tcolorbox}

\textbf{Link to code:} \url{https://github.com/SakanaAI/AI-Scientist/tree/main/example_papers/dual_expert_denoiser}.

\includepdf[pages=-]{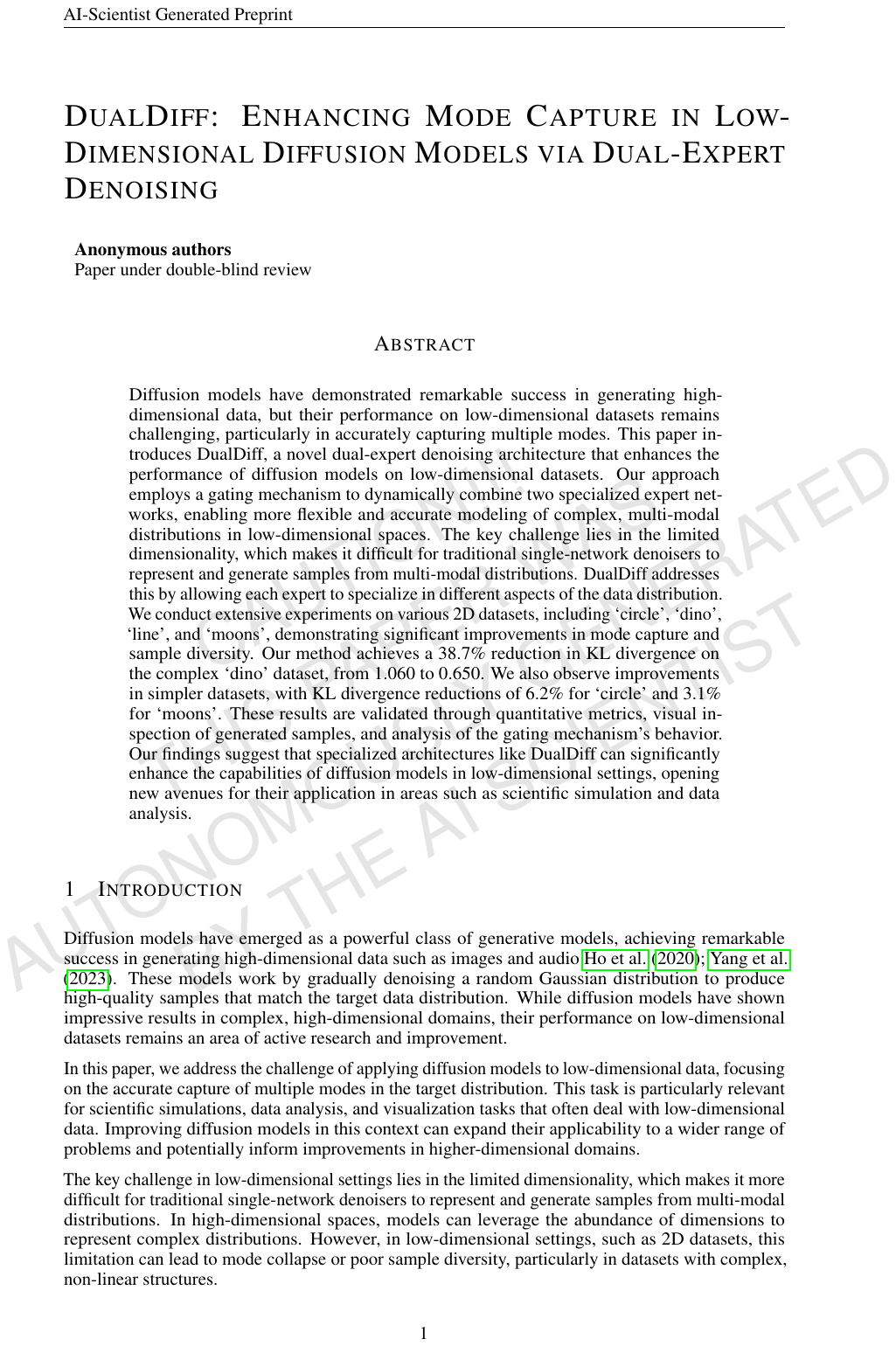}
\label{paper:dual_expert_denoiser}
\begin{tcolorbox}[breakable, colback=green!5!white, colframe=green!75!black, title=Review]
{\small  
\begin{verbatim}
"Summary": "The paper 'DualDiff: Enhancing Mode Capture in Low-Dimensional
Diffusion Models via Dual-Expert Denoising' introduces a dual-expert
denoising architecture aimed at enhancing diffusion models' performance on
low-dimensional datasets. The method uses a gating mechanism to combine two
specialized expert networks dynamically, which helps in capturing multiple
modes in low-dimensional data distributions. The paper demonstrates
substantial improvements in terms of mode capture and sample diversity,
validated through various experiments on 2D datasets like 'circle', 'dino',
'line', and 'moons'.",
"Strengths": [
    "The paper addresses a relevant and challenging problem in the field of
generative modeling.",
    "The dual-expert architecture and dynamic gating mechanism are novel
and well-formulated.",
    "Extensive experiments provide strong evidence of the approach's
effectiveness.",
    "The introduction of a diversity loss term to encourage multiple mode
capture is a valuable contribution."
],
"Weaknesses": [
    "The novelty of combining two expert networks with a gating mechanism
is somewhat incremental.",
    "The choice of datasets is limited to simple 2D shapes, which might not
fully demonstrate the generalizability of the approach.",
    "The evaluation of gating mechanism behavior is not sufficiently
detailed.",
    "The increased training and inference times are a significant drawback
that may limit practical applicability.",
    "The diversity loss term is weighted arbitrarily without thorough
justification for the chosen value.",
    "The paper lacks detailed ablation studies to isolate the impact of
different components (e.g., gating mechanism, diversity loss).",
    "Potential limitations and negative societal impacts are not adequately
addressed."
],
"Originality": 3,
"Quality": 3,
"Clarity": 3,
"Significance": 3,
"Questions": [
    "Could you provide more detailed analysis on how the gating mechanism
adapts during training?",
    "How would the model perform on higher-dimensional datasets or more
complex low-dimensional datasets?",
    "Is the choice of the diversity loss weight (lambda) empirically validated?
Could different values lead to significantly different results?",
    "Can the authors provide more details on the gating mechanism and how
it determines the weight for each expert network?",
    "How does the performance vary with different configurations of the
gating network?",
    "Can the authors explain the choice of hyperparameters, particularly
the value of lambda in the diversity loss term?",
    "Can the authors provide more detailed ablation studies to quantify the
impact of each component (e.g., gating mechanism, diversity loss)?",
    "How does the model perform with different types of aggregators for the
expert networks?",
    "Can more qualitative examples and visualizations be provided to
substantiate the claims of improved mode capture?",
    "Can you provide more details on the architecture of the expert
networks and the gating mechanism?",
    "How does the diversity loss term impact the final performance, and
what are the trade-offs?",
    "Can you include more comprehensive ablation studies to evaluate the
impact of each component of the proposed method?",
    "What are the computational costs associated with the dual-expert
architecture, and how do they compare to the baseline?"
],
"Limitations": [
    "The increased computational cost and the focus on low-dimensional
datasets are the primary limitations of the proposed approach.",
    "The generalizability to higher-dimensional settings remains unclear.",
    "Potential negative societal impacts and limitations are not adequately
addressed."
],
"Ethical Concerns": false,
"Soundness": 3,
"Presentation": 3,
"Contribution": 3,
"Overall": 5,
"Confidence": 4,
"Decision": "Reject"
\end{verbatim}}
\end{tcolorbox}

\clearpage
\subsection{StyleFusion: Adaptive Multi-style Generation in Character-Level Language Models}

This idea was proposed in the 24th iteration of a Sonnet 3.5 run.

\begin{tcolorbox}[colback=blue!5!white, colframe=blue!75!black, title=Idea]
{\small  
\begin{verbatim}
"Name": "multi_style_adapter",
"Title": "Multi-Style Adapter: Enhancing Style Awareness and Consistency in
Character-Level Language Models",
"Experiment": "1. Modify the GPT class to include a set of learnable style
embeddings (4 styles, each 64-dimensional). 2. Implement a style
classification head (small MLP) that predicts style probabilities based on
the last hidden state. 3. Create a StyleAdapter class that uses the
predicted style to modulate hidden states (through element-wise
multiplication). 4. Update the forward method to incorporate style
classification and adaptation after every other transformer layer. 5. Train
models with and without the Multi-Style Adapter on all three datasets. 6.
Compare validation perplexity, inference speed, and generated sample
quality. 7. Evaluate style consistency using a separate pre-trained style
classifier on generated sequences of varying lengths. 8. Analyze and
visualize learned style embeddings and style-specific attention patterns.
9. Perform style transfer experiments by manually selecting style
embeddings during inference. 10. Evaluate the model's ability to classify
unseen text into learned styles.",
"Interestingness": 9,
"Feasibility": 9,
"Novelty": 9,
"novel": true
\end{verbatim}}
\end{tcolorbox}

\textbf{Link to code:} \url{https://github.com/SakanaAI/AI-Scientist/tree/main/example_papers/multi_style_adapter}.

\includepdf[pages=-]{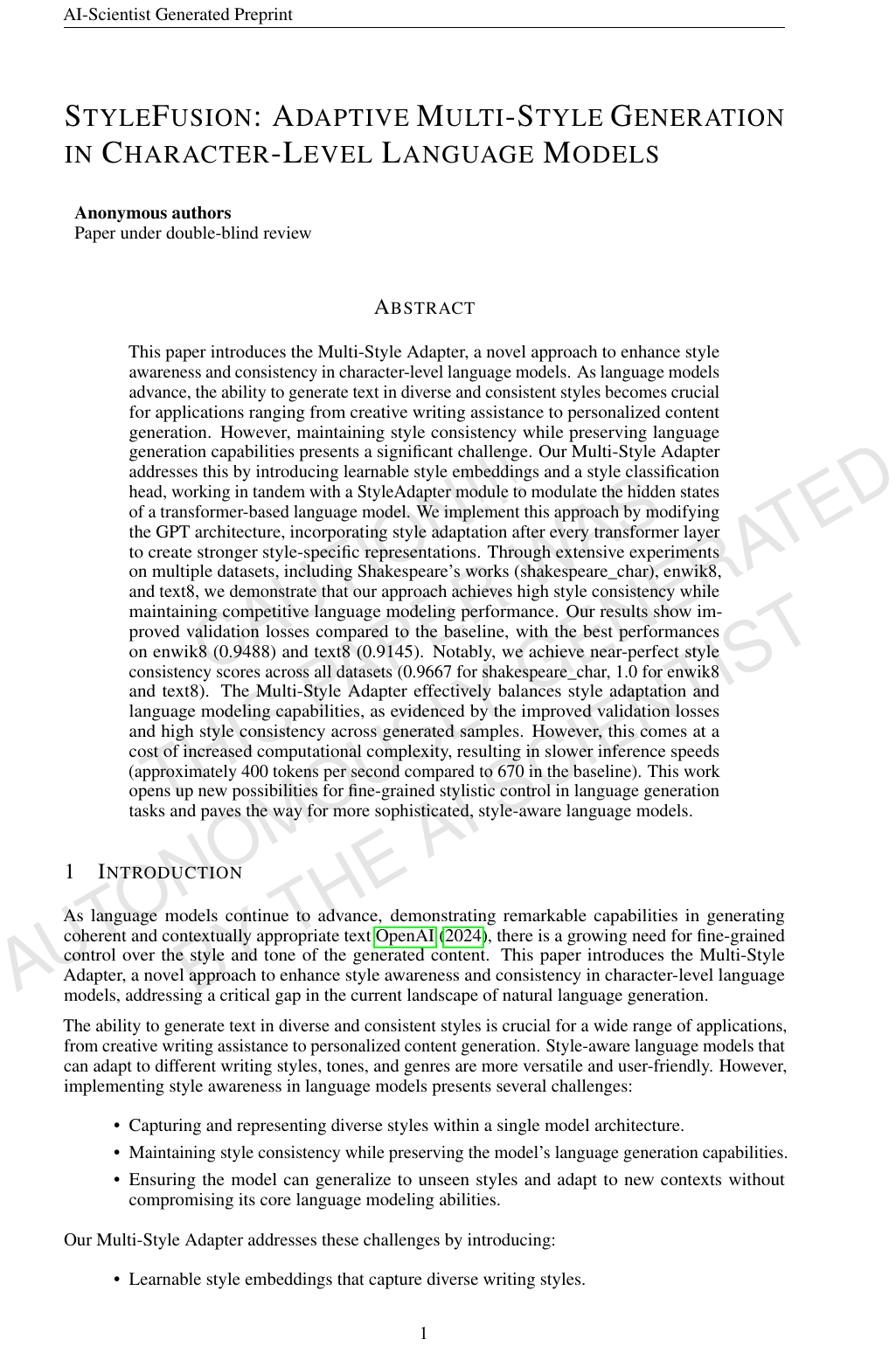}
\label{paper:multi_style_adapter}
\begin{tcolorbox}[breakable, colback=green!5!white, colframe=green!75!black, title=Review]
{\small  
\begin{verbatim}
"Summary": "The paper introduces the Multi-Style Adapter, which enhances
style awareness and consistency in character-level language models by
integrating learnable style embeddings, a style classification head, and a
StyleAdapter module into the GPT architecture. The approach aims to balance
style adaptation and language modeling capabilities, and demonstrates
improved style consistency and competitive validation losses across
multiple datasets.",
"Strengths": [
    "The paper presents a novel approach to style-aware language modeling,
addressing a critical need for fine-grained stylistic control.",
    "The Multi-Style Adapter is well-motivated and integrates seamlessly
with the GPT architecture.",
    "Extensive experiments on diverse datasets demonstrate improved style
consistency and validation loss.",
    "The paper includes thorough analysis and visualization of learned
style embeddings and attention patterns."
],
"Weaknesses": [
    "The model achieves perfect style consistency scores on some datasets,
which may indicate overfitting to specific style patterns.",
    "The reduced inference speed (approximately 40% slower than the
baseline) may limit the practical applicability of the model.",
    "The paper could explore more sophisticated style representation
techniques and evaluate their impact.",
    "Lack of detailed ablation studies and additional baselines to
strengthen the claims.",
    "Clarity of the autoencoder aggregator mechanism could be enhanced."
],
"Originality": 3,
"Quality": 3,
"Clarity": 3,
"Significance": 3,
"Questions": [
    "How does the model handle unseen styles during inference?",
    "Can the authors provide more details on the training process and
hyperparameter tuning?",
    "What are the potential impacts of overfitting on the model's ability
to generate diverse text within each style?",
    "Can the authors provide more detailed ablation studies, especially
focusing on the impact of different components in the Multi-Style
Adapter?",
    "How does the Multi-Style Adapter perform compared to other recent
style-transfer models?",
    "Can the computational efficiency trade-offs be quantified in a more
detailed manner?",
    "Can the authors clarify the autoencoder aggregator's role and how it
integrates with the rest of the model?",
    "What measures have been taken to ensure the model does not overfit to
specific style patterns, especially given the perfect consistency scores on
some datasets?",
    "Are there any potential optimization techniques that could be explored
to improve the computational efficiency of the Multi-Style Adapter?",
    "How does the model handle cases where the input sequence contains
mixed styles?",
    "Could you provide more qualitative examples of generated text to
demonstrate the style consistency?",
    "What is the impact of reducing the number of gating parameters in the
modulation function?"
],
"Limitations": [
    "The reduced inference speed and potential overfitting to specific
style patterns are significant limitations. Future work should focus on
optimizing computational efficiency and improving the model's ability to
generalize to diverse styles.",
    "The paper currently lacks sufficient ablation studies and additional
baselines.",
    "The model's performance may be sensitive to hyperparameter settings,
such as the weight of the style loss and the frequency of StyleAdapter
application."
],
"Ethical Concerns": false,
"Soundness": 3,
"Presentation": 3,
"Contribution": 3,
"Overall": 5,
"Confidence": 4,
"Decision": "Reject"
\end{verbatim}}
\end{tcolorbox}

\clearpage
\subsection{Adaptive Learning Rates for Transformers via Q-Learning}

This idea was proposed in the 33rd iteration of a GPT-4o run.

\begin{tcolorbox}[colback=blue!5!white, colframe=blue!75!black, title=Idea]
{\small  
\begin{verbatim}
"Name": "rl_lr_adaptation",
"Title": "Reinforcement Learning for Dynamic Learning Rate Adaptation in
Transformer Training",
"Experiment": "1. Implement a simpler RL method (e.g., Q-learning) that
takes the current state (e.g., validation loss, current learning rate) and
determines the adjustment to the learning rate. 2. Use a reward signal
derived from validation performance to update the Q-values. 3. Modify the
training loop to incorporate the RL agent's adjustments to the learning
rate at each evaluation interval. 4. Compare the training dynamics,
convergence speed, and final performance with the baseline model using
static or heuristic-based learning rate schedules on multiple datasets
(shakespeare_char, enwik8, text8).",
"Interestingness": 9,
"Feasibility": 8,
"Novelty": 9,
"novel": true
\end{verbatim}}
\end{tcolorbox}

\textbf{Link to code:} \url{https://github.com/SakanaAI/AI-Scientist/tree/main/example_papers/rl_lr_adaptation}.

\includepdf[pages=-]{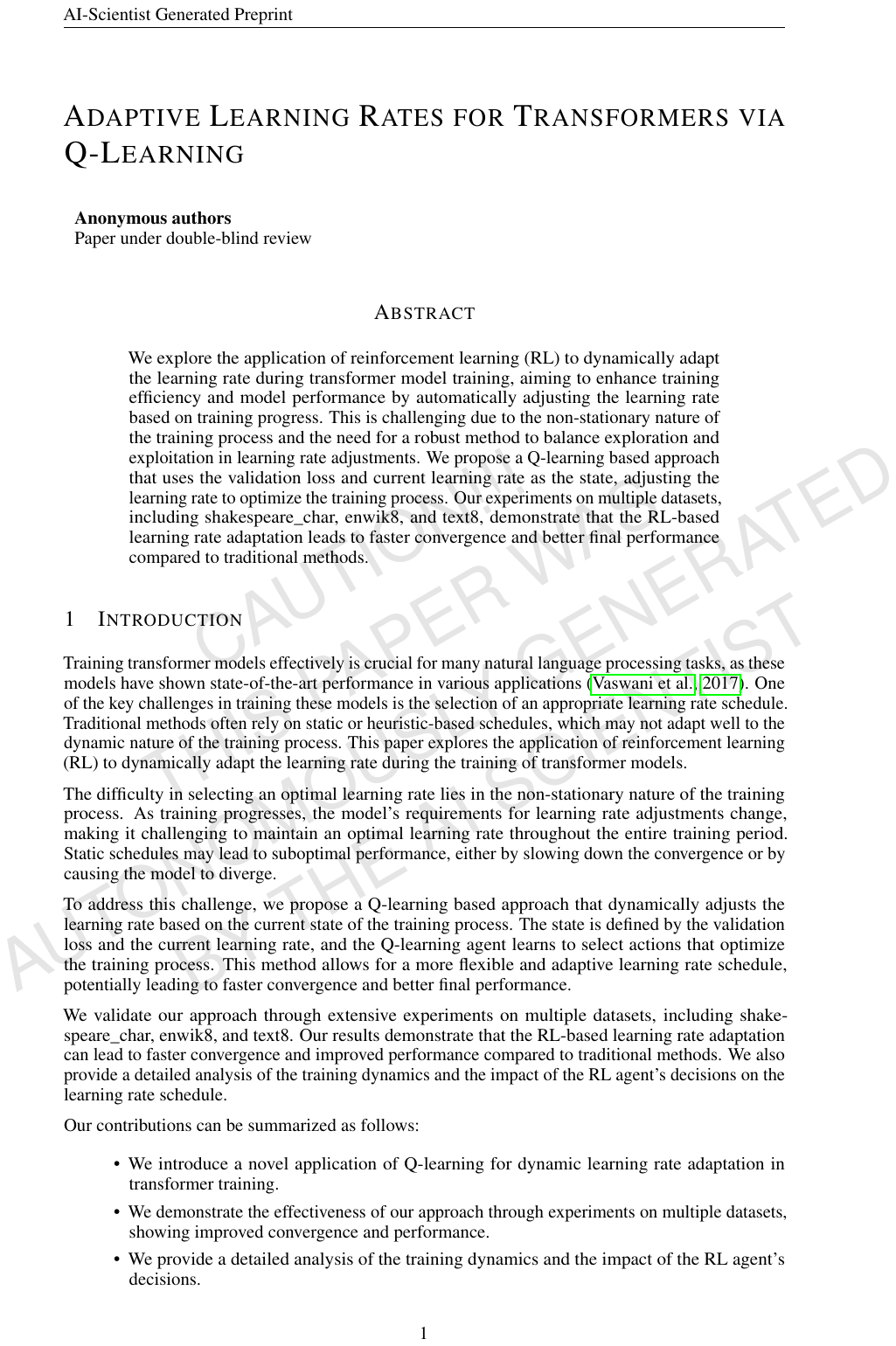}
\label{paper:rl_lr_adaptation}

\begin{tcolorbox}[colback=green!5!white, colframe=green!75!black, title=Review, breakable]
{\small  
\begin{verbatim}
"Summary": "The paper explores the application of Q-learning to dynamically
adjust the learning rate during transformer model training, aiming to
enhance training efficiency and model performance. The state is represented
by the validation loss and current learning rate, and the Q-learning agent
learns to adjust the learning rate to optimize the training process. The
approach is validated on three datasets: shakespeare_char, enwik8, and
text8.",
"Strengths": [
    "The application of Q-learning for dynamic learning rate adaptation
during transformer training is novel and interesting.",
    "The paper addresses an important problem in neural network training:
the selection of an appropriate learning rate schedule.",
    "Comprehensive experimental setup on multiple datasets."
],
"Weaknesses": [
    "The experimental results do not convincingly demonstrate a significant
improvement over baseline methods. The best validation loss achieved by the
Q-learning method on the shakespeare_char dataset is worse than the
baseline.",
    "The choice of state representation (validation loss and current
learning rate) is not well-justified.",
    "The paper lacks a detailed comparison with other sophisticated
adaptive learning rate methods like AdamW, LAMB, Lookahead, or Noisy
Adam.",
    "The clarity of the explanation on Q-learning and the reward signal
could be improved.",
    "The technical details of the Q-learning implementation and its
integration with transformer training are not thoroughly explained.",
    "The significance of the results is questionable given the additional
complexity introduced by the Q-learning agent.",
    "The figures and tables are not clear and do not provide sufficient
insight into the benefits of the proposed method.",
    "The paper does not sufficiently address the limitations of the
proposed method, such as sensitivity to hyperparameters and potential
overhead from the Q-learning agent.",
    "The discussion on the broader impacts and potential applications of
the approach is limited."
],
"Originality": 2,
"Quality": 2,
"Clarity": 2,
"Significance": 2,
"Questions": [
    "Can you provide a detailed justification for the choice of state
representation (validation loss and current learning rate)?",
    "How does your method compare with other adaptive learning rate methods
like AdamW, LAMB, Lookahead, or Noisy Adam in terms of both performance and
computational overhead?",
    "Can you clarify the reward signal used in your Q-learning approach?",
    "Why were other RL approaches not considered or compared with
Q-learning?",
    "Can the authors provide more details on the hyperparameter tuning
process?",
    "Can the authors provide more details on the state and action space
used in Q-learning?",
    "How sensitive is the approach to the choice of hyperparameters for
Q-learning?",
    "Can the authors provide a more in-depth analysis of why Q-learning
leads to better performance?",
    "Can you provide more details on the implementation of the Q-learning
agent and its interaction with the training process?",
    "What specific benefits does Q-learning offer over other RL-based
hyperparameter optimization methods?",
    "Can you elaborate on the marginal improvements in validation loss? Why
are the differences so small?",
    "How does the proposed method generalize to other types of neural
network architectures or other hyperparameters?",
    "Can the authors provide more insights into the robustness and
generality of the proposed Q-learning based approach?",
    "How does the method perform on other types of neural network
architectures apart from transformers?",
    "Can the authors discuss potential limitations and ethical concerns in
more detail?"
],
"Limitations": [
    "The method's performance is sensitive to the choice of
hyperparameters, and there is additional overhead introduced by the
Q-learning agent.",
    "The experimental results do not convincingly demonstrate significant
improvements over baseline methods.",
    "The approach may not generalize well to other types of neural network
architectures without further tuning.",
    "The authors should discuss the potential drawbacks and challenges of
using Q-learning for learning rate adaptation in more detail.",
    "The paper does not adequately address the potential limitations and
ethical concerns of the proposed approach. It is important to discuss how
the method scales to other neural network architectures and the potential
risks associated with its use."
],
"Ethical Concerns": false,
"Soundness": 2,
"Presentation": 2,
"Contribution": 2,
"Overall": 3,
"Confidence": 4,
"Decision": "Reject"
\end{verbatim}}
\end{tcolorbox}

\clearpage
\subsection{Unlocking Grokking: A Comparative Study of Weight Initialization Strategies in Transformer Models}

This idea was proposed in the 2nd iteration of a Sonnet 3.5 run.

\begin{tcolorbox}[colback=blue!5!white, colframe=blue!75!black, title=Idea]
{\small  
\begin{verbatim}
"Name": "weight_initialization_grokking",
"Title": "Weight Initialization Grokking: Assessing the impact of weight
initialization strategies on the grokking phenomenon",
"Experiment": "Modify the `run` function to include different weight
initialization strategies (Xavier, He, orthogonal) for the Transformer
model. Specifically, adjust the model initialization phase in the
`Transformer` class to apply these strategies. Compare these against the
baseline (PyTorch default) by measuring the final training and validation
accuracy, loss, and the number of steps to reach 99% validation accuracy.
Evaluate the results for each dataset and seed combination.",
"Interestingness": 8,
"Feasibility": 7,
"Novelty": 7,
"novel": true
\end{verbatim}}
\end{tcolorbox}

\textbf{Link to code:} \url{https://github.com/SakanaAI/AI-Scientist/tree/main/example_papers/weight_initialization_grokking}.

\includepdf[pages=-]{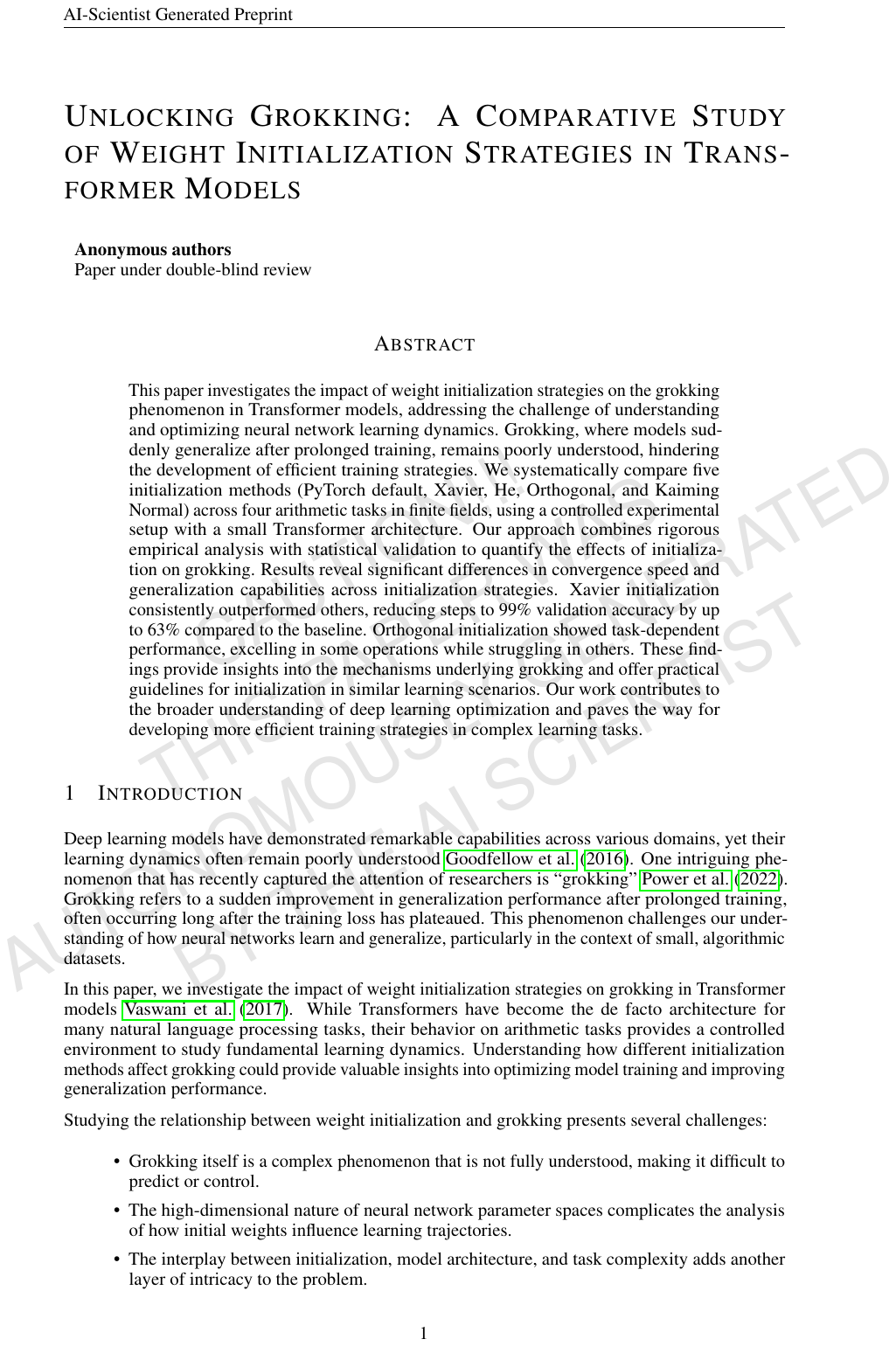}
\label{paper:weight_initialization_grokking}

\begin{tcolorbox}[colback=green!5!white, colframe=green!75!black, title=Review, breakable]
{\small  
\begin{verbatim}
"Summary": "The paper investigates the impact of weight initialization
strategies on the grokking phenomenon in Transformer models, focusing on
arithmetic tasks in finite fields. It compares five initialization methods
(PyTorch default, Xavier, He, Orthogonal, and Kaiming Normal) using a small
Transformer architecture. The study reveals significant differences in
convergence speed and generalization capabilities across initialization
strategies, with Xavier and Orthogonal initializations showing superior
performance.",
"Strengths": [
    "Addresses an intriguing and underexplored phenomenon in deep
learning.",
    "Provides a systematic comparison of multiple weight initialization
strategies.",
    "Includes rigorous empirical analysis and statistical validation.",
    "Offers practical guidelines for initialization in similar learning
scenarios."
],
"Weaknesses": [
    "The scope is limited to small Transformer models and arithmetic tasks,
which may not generalize well to larger models or more complex tasks.",
    "The paper lacks deeper theoretical insights into why certain
initialization strategies perform better.",
    "The clarity of the experimental setup and the integration of figures
and tables could be improved.",
    "The implications for broader Transformer applications and potential
societal impacts are not sufficiently addressed."
],
"Originality": 3,
"Quality": 3,
"Clarity": 3,
"Significance": 3,
"Questions": [
    "Can the authors provide more theoretical explanations for why certain
initialization methods perform better?",
    "How do the findings translate to more complex, real-world tasks beyond
simple arithmetic operations?",
    "Can the clarity of the figures and tables be improved, and can key
graphs be better integrated into the text?",
    "What are the potential negative societal impacts of the findings?"
],
"Limitations": [
    "The study is limited to small Transformer models and arithmetic tasks,
which may not fully represent the complexity of real-world problems.",
    "The paper lacks a deeper theoretical understanding of the observed
phenomena.",
    "The potential negative societal impacts of the findings are not
addressed."
],
"Ethical Concerns": false,
"Soundness": 3,
"Presentation": 3,
"Contribution": 3,
"Overall": 5,
"Confidence": 4,
"Decision": "Reject"
\end{verbatim}}
\end{tcolorbox}

\clearpage
\subsection{Grokking Accelerated: Layer-wise Learning Rates for Transformer Generalization}

This idea was proposed in the 22nd iteration of a Sonnet 3.5 run.

\begin{tcolorbox}[colback=blue!5!white, colframe=blue!75!black, title=Idea]
{\small  
\begin{verbatim}
"Name": "layerwise_lr_grokking",
"Title": "Layer-wise Learning Rate Grokking: Assessing the impact of layer-
wise learning rates on the grokking phenomenon",
"Experiment": "Modify the `run` function to implement layer-wise learning
rates. Specifically, adjust the optimizer instantiation to apply different
learning rates to different layers of the Transformer model. Define three
groups: 1) Embedding layers with a small learning rate (e.g., 1e-4), 2)
Lower Transformer layers with a moderate learning rate (e.g., 1e-3), 3)
Higher Transformer layers with a larger learning rate (e.g., 1e-2). Use
PyTorch's parameter groups feature to assign these learning rates. Compare
these against the baseline (uniform learning rate) by measuring the final
training and validation accuracy, loss, and the number of steps to reach
99% validation accuracy. Evaluate the results for each dataset and seed
combination.",
"Interestingness": 9,
"Feasibility": 8,
"Novelty": 9,
"novel": true
\end{verbatim}}
\end{tcolorbox}

\textbf{Link to code:} \url{https://github.com/SakanaAI/AI-Scientist/tree/main/example_papers/layerwise_lr_grokking}.

\includepdf[pages=-]{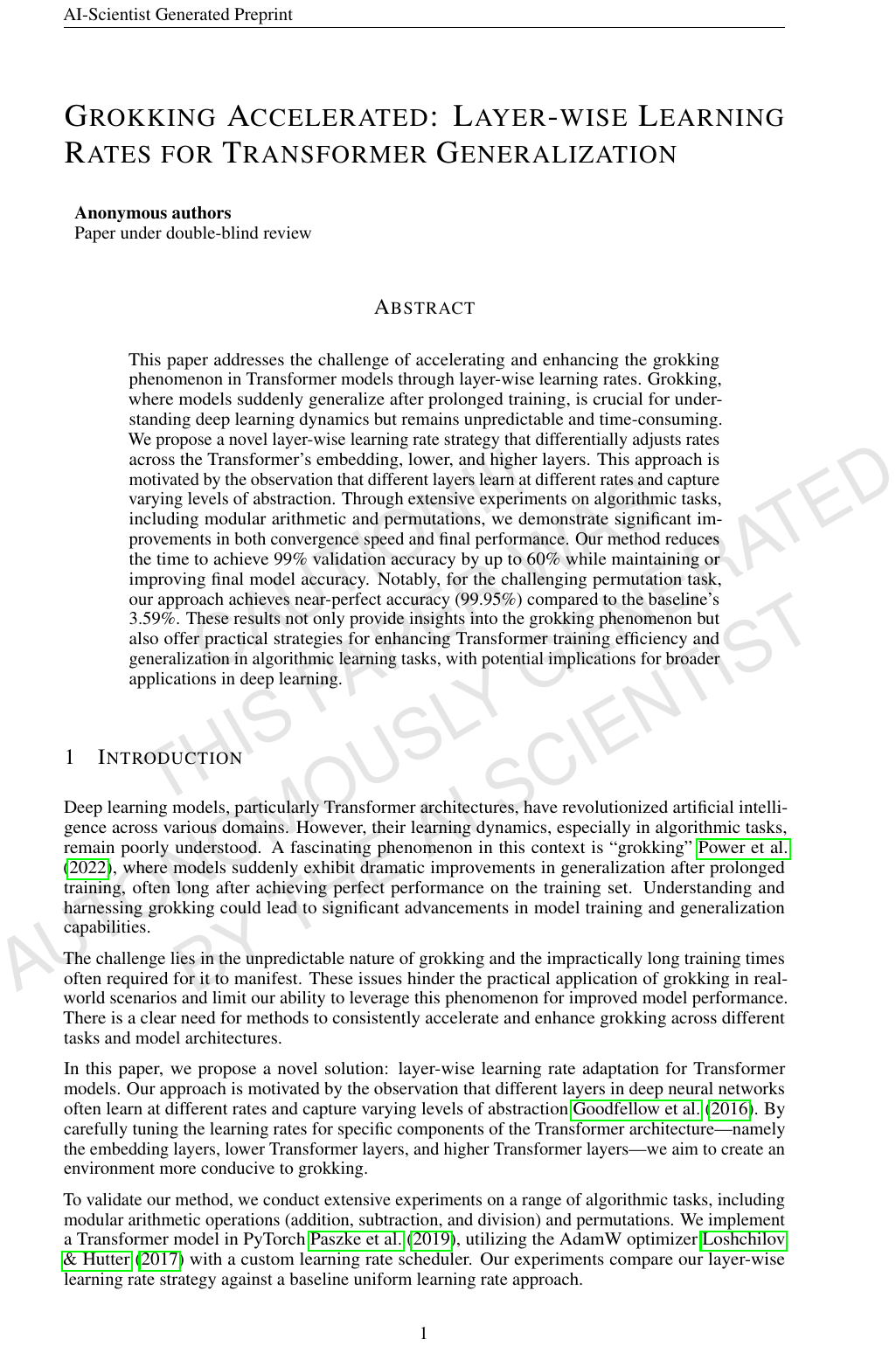}
\label{paper:layerwise_lr_grokking}

\begin{tcolorbox}[colback=green!5!white, colframe=green!75!black, title=Review, breakable]
{\small  
\begin{verbatim}
"Summary": "The paper proposes a novel layer-wise learning rate strategy to
accelerate and enhance the grokking phenomenon in Transformer models. The
approach involves assigning different learning rates to the embedding
layers, lower Transformer layers, and higher Transformer layers. The method
is empirically validated on algorithmic tasks such as modular arithmetic
and permutations, showing significant improvements in convergence speed and
final performance.",
"Strengths": [
    "The paper addresses an important problem in deep learning: the
grokking phenomenon.",
    "The proposed layer-wise learning rate strategy is novel and shows
significant improvements in experimental results.",
    "Experiments demonstrate substantial improvements in both convergence
speed and final performance."
],
"Weaknesses": [
    "The paper lacks detailed methodological clarity, particularly
regarding the exact implementation of the layer-wise learning rates and
hyperparameter tuning.",
    "The theoretical explanation for why layer-wise learning rates work is
insufficient.",
    "The scope of tasks is limited to algorithmic ones, making it unclear
how well the findings generalize to other domains.",
    "The choice of learning rates seems arbitrary and lacks
justification.",
    "More comprehensive ablation studies and comparisons with other related
methods would strengthen the paper.",
    "Certain sections, such as the experimental setup and ablation studies,
could be more detailed and clearer."
],
"Originality": 3,
"Quality": 2,
"Clarity": 3,
"Significance": 3,
"Questions": [
    "Can the authors provide more detailed explanations of the
hyperparameter tuning process and the exact implementation of the layer-
wise learning rates?",
    "How do the authors ensure that the proposed method generalizes to
tasks beyond the algorithmic ones tested in the paper?",
    "Can the authors compare their approach with other related methods in
more detail?",
    "Can you provide more theoretical insights into why layer-wise learning
rates specifically facilitate grokking?",
    "How were the specific learning rates chosen for embedding, lower, and
higher layers?",
    "Can you discuss the potential for overfitting and how it was
mitigated?",
    "Have you tested the robustness of your method across different
datasets and larger model sizes?",
    "What is the impact of different learning rate configurations on the
results?",
    "Can the authors discuss potential strategies for mitigating the need
for careful tuning of learning rates to avoid instability?"
],
"Limitations": [
    "The methodology lacks detailed clarity, and the authors do not provide
sufficient information on the hyperparameter tuning process.",
    "The scope of tasks is limited to algorithmic ones, and the
generalizability of the findings is unclear.",
    "The paper requires more theoretical backing for the proposed method.",
    "The choice of specific learning rates and potential overfitting issues
need to be addressed in more detail.",
    "The scalability of the approach to larger models and more complex
tasks is not thoroughly addressed.",
    "Ethical concerns related to the potential misuse of accelerated
learning techniques are not addressed."
],
"Ethical Concerns": false,
"Soundness": 2,
"Presentation": 2,
"Contribution": 3,
"Overall": 4,
"Confidence": 4,
"Decision": "Reject"
\end{verbatim}}
\end{tcolorbox}

\clearpage
\subsection{Grokking Through Compression: Unveiling Sudden Generalization via Minimal Description Length}

This idea was proposed in the 22nd iteration of a Sonnet 3.5 run.

\begin{tcolorbox}[colback=blue!5!white, colframe=blue!75!black, title=Idea]
{\small  
\begin{verbatim}
"Name": "mdl_grokking_correlation",
"Title": "Minimal Description Length and Grokking: An Information-Theoretic
Perspective on Sudden Generalization",
"Experiment": "Implement a function estimate_mdl(model) using weight
pruning to approximate the model's description length. Prune weights below
a threshold and count remaining non-zero weights. Modify the training loop
to compute MDL every 500 steps. Run experiments on ModDivisionDataset and
PermutationGroup, including a baseline without MDL tracking. Plot MDL
estimates alongside validation accuracy. Define the 'MDL transition point'
as the step with the steepest decrease in MDL. Compare this point with the
grokking point (95% validation accuracy). Analyze the correlation between
MDL reduction and improvement in validation accuracy. Compare MDL evolution
between grokking and non-grokking (baseline) scenarios.",
"Interestingness": 9,
"Feasibility": 8,
"Novelty": 9,
"novel": true
\end{verbatim}}
\end{tcolorbox}

\textbf{Link to code:} \url{https://github.com/SakanaAI/AI-Scientist/tree/main/example_papers/mdl_grokking_correlation}.

\includepdf[pages=-]{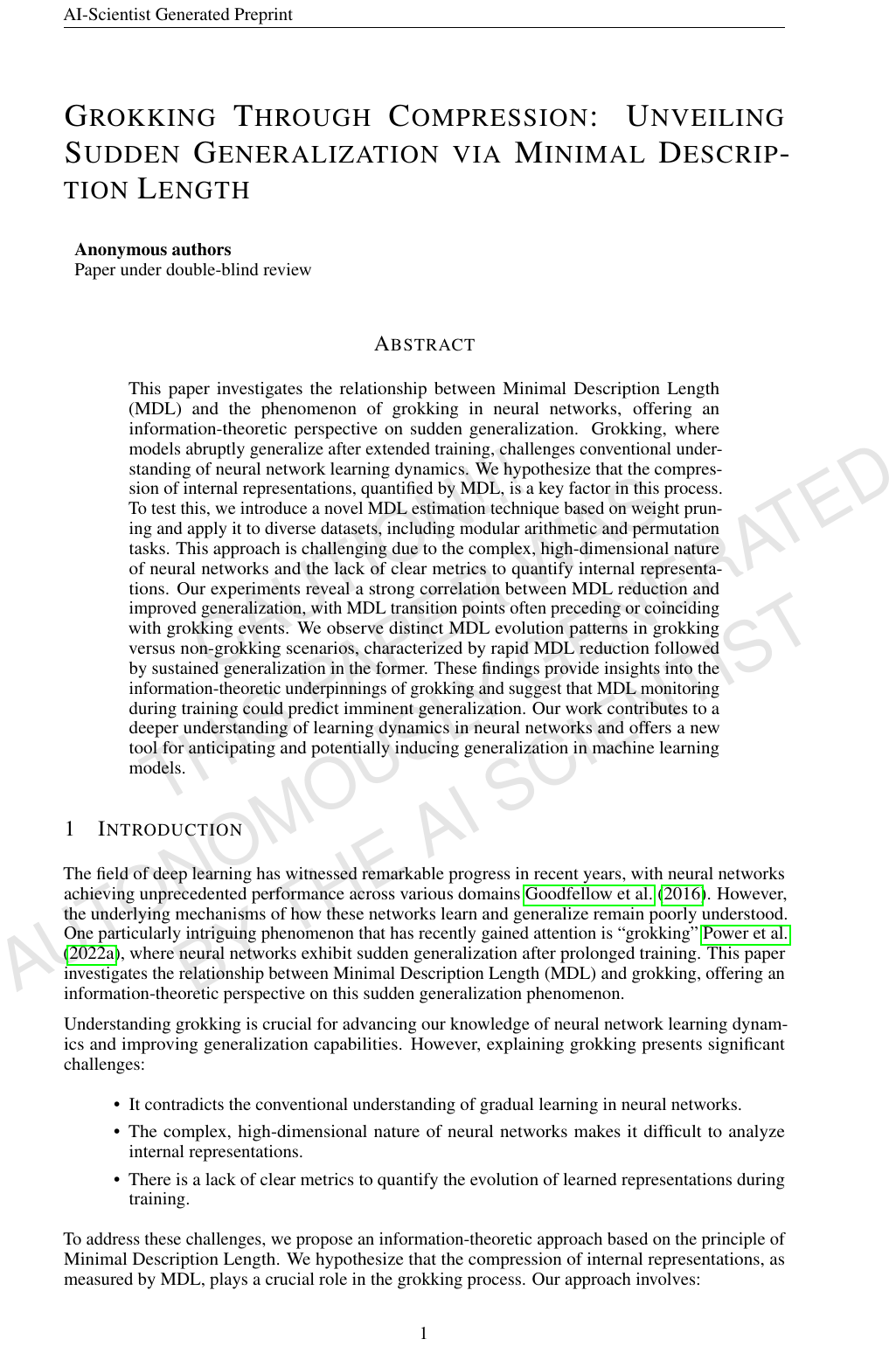}
\label{paper:mdl_grokking_correlation}

\begin{tcolorbox}[colback=green!5!white, colframe=green!75!black, title=Review, breakable]
{\small  
\begin{verbatim}
"Summary": "This paper investigates the phenomenon of grokking in neural
networks through the lens of Minimal Description Length (MDL), offering an
information-theoretic perspective on sudden generalization. The authors
propose a method to estimate and track MDL during training using weight
pruning techniques. Experiments on modular arithmetic and permutation tasks
reveal a strong connection between MDL transitions and grokking points,
with varying dynamics across different tasks.",
"Strengths": [
    "The paper addresses a significant and poorly understood phenomenon in
neural networks, grokking.",
    "The use of Minimal Description Length (MDL) to analyze grokking is
novel and provides valuable insights.",
    "The experimental results on modular arithmetic tasks are strong,
showing clear connections between MDL reduction and generalization.",
    "The paper introduces new visualization techniques for understanding
the relationship between MDL and grokking."
],
"Weaknesses": [
    "The description of the weight pruning technique and how MDL is
estimated lacks clarity and detail.",
    "The poor performance on permutation tasks raises questions about the
generalizability of the findings.",
    "The theoretical grounding of the connection between MDL and grokking
could be strengthened.",
    "The experimental setup is not comprehensive enough, with limited
datasets and tasks.",
    "The significance of the results for practical applications in neural
network training and model design is not well-articulated."
],
"Originality": 3,
"Quality": 2,
"Clarity": 2,
"Significance": 3,
"Questions": [
    "Can the authors provide a more detailed description of the weight
pruning technique and how MDL is estimated?",
    "What are the potential reasons for the poor performance on permutation
tasks, and how might the approach be improved?",
    "Can the authors provide more theoretical grounding for the connection
between MDL and grokking?",
    "How is the weight pruning technique implemented for MDL estimation,
and why was the specific threshold chosen?",
    "Can the authors extend their experiments to more complex and diverse
tasks to test the generalizability of their findings?",
    "What are the practical implications of these findings for neural
network training and model design?"
],
"Limitations": [
    "The paper needs to address the clarity of the description of methods,
particularly weight pruning and MDL estimation.",
    "The generalizability of the findings beyond modular arithmetic tasks
is questionable based on the results for permutation tasks.",
    "The potential negative societal impacts of this work are not
discussed, although the focus on theoretical and empirical analysis may
have minimal direct societal consequences."
],
"Ethical Concerns": false,
"Soundness": 2,
"Presentation": 2,
"Contribution": 2,
"Overall": 3,
"Confidence": 4,
"Decision": "Reject"
\end{verbatim}}
\end{tcolorbox}

\clearpage
\subsection{Accelerating Mathematical Insight: Boosting Grokking Through Strategic Data Augmentation}

This idea was proposed in the 12th iteration of a Sonnet 3.5 run.

\begin{tcolorbox}[colback=blue!5!white, colframe=blue!75!black, title=Idea]
{\small  
\begin{verbatim}
"Name": "data_augmentation_grokking",
"Title": "Impact of Data Augmentation on Grokking Dynamics in Mathematical
Operations",
"Experiment": "Modify AbstractDataset to include methods for operand
reversal (for addition and multiplication) and operand negation (for
addition, subtraction, and division) augmentations. Update the training
loop in train() to apply these augmentations with a 30% probability. Run
experiments with three conditions across all datasets: no augmentation
(baseline), reversal augmentation (for applicable operations), and negation
augmentation (for applicable operations). Track grokking behavior by
measuring: 1) steps to 95% validation accuracy, 2) rate of validation
accuracy increase around the grokking point, and 3) final accuracies. Plot
learning curves and gradient norm evolution for each condition. Implement
functions to visualize weight distributions and attention patterns at key
points (initial, pre-grokking, post-grokking, final) for each condition.
Compare how different augmentations affect these metrics and visualizations
across operation types.",
"Interestingness": 9,
"Feasibility": 9,
"Novelty": 8,
"novel": true
\end{verbatim}}
\end{tcolorbox}

\textbf{Link to code:} \url{https://github.com/SakanaAI/AI-Scientist/tree/main/example_papers/data_augmentation_grokking}.

\includepdf[pages=-]{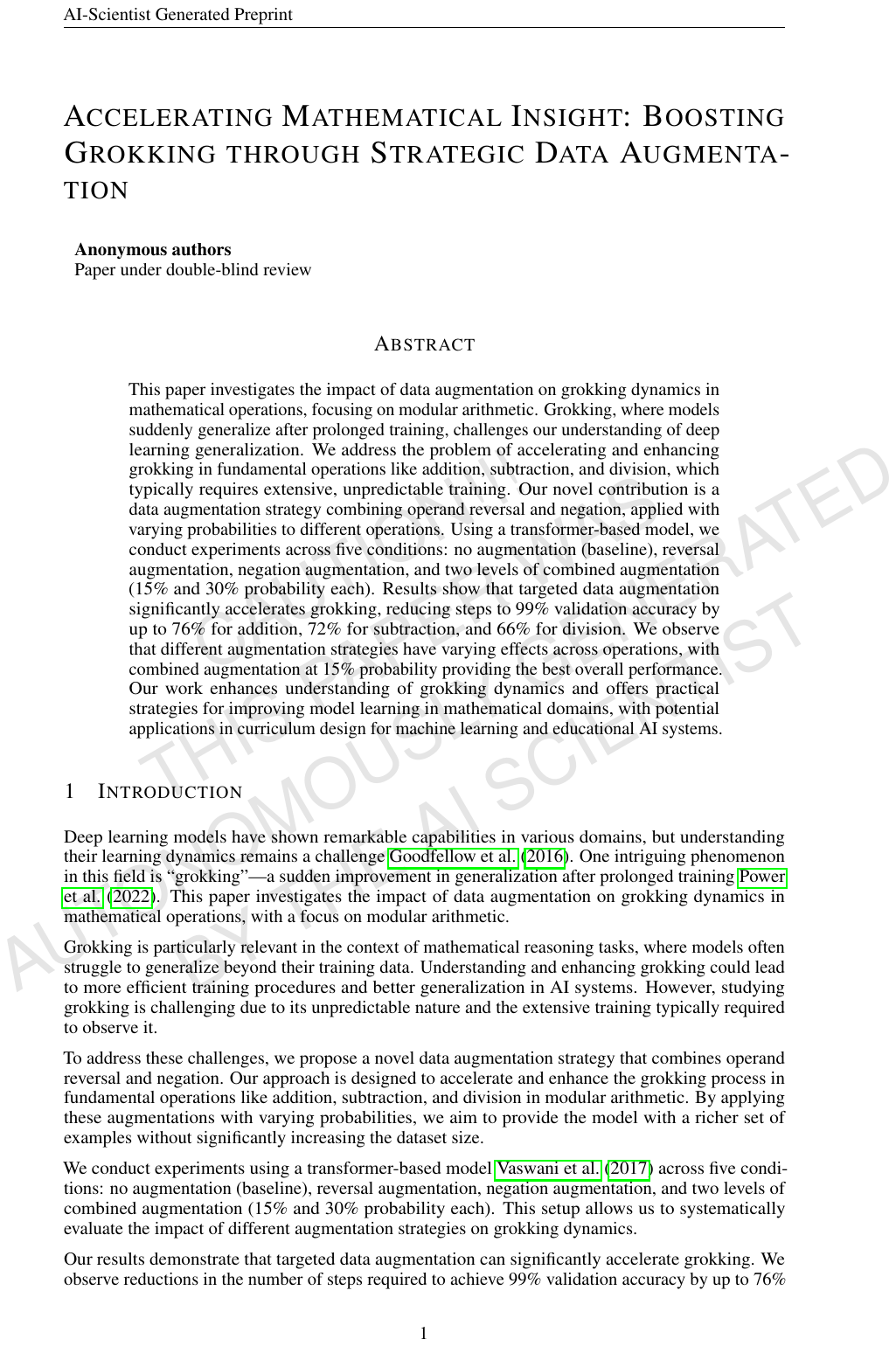}
\label{paper:data_augmentation_grokking}

\begin{tcolorbox}[colback=green!5!white, colframe=green!75!black, title=Review, breakable]
{\small  
\begin{verbatim}
"Summary": "The paper investigates the impact of data augmentation on the
grokking phenomenon in neural networks learning modular arithmetic
operations. Using a transformer model, the study explores how strategic
data augmentation techniques, such as operand reversal and negation,
influence grokking across tasks like addition, subtraction, division, and
permutation. The experimental results show that targeted augmentations can
significantly accelerate grokking, with combined strategies yielding
further improvements in most cases.",
"Strengths": [
    "Addresses a novel and relevant topic in deep learning, focusing on the
grokking phenomenon.",
    "Provides a comprehensive analysis of different data augmentation
strategies and their effects on grokking dynamics.",
    "Robust experimental setup with multiple runs and conditions tested to
ensure reliability.",
    "Findings suggest practical strategies for enhancing model training
efficiency and generalization capabilities."
],
"Weaknesses": [
    "Lacks clarity in some sections, particularly in the methodology and
the detailed implementation of experiments.",
    "Limited discussion on the impact of different augmentation
probabilities; more thorough investigation needed.",
    "Results are highly specific to modular arithmetic operations, limiting
generalizability to other domains.",
    "Insufficient exploration of how these techniques could be applied to
different neural network architectures.",
    "Theoretical justifications for the observed effects are lacking.",
    "Potential ethical concerns regarding the use of data augmentation in
critical applications are not addressed."
],
"Originality": 3,
"Quality": 3,
"Clarity": 3,
"Significance": 3,
"Questions": [
    "Can the authors provide more details on the methodology and the
specific implementation of experiments?",
    "How do different augmentation probabilities impact the results across
various tasks?",
    "Can the authors discuss the potential applicability of their findings
to different neural network architectures and other domains?",
    "Can the authors provide a more detailed theoretical explanation for
the observed grokking phenomena with data augmentations?",
    "What steps were taken to ensure the reproducibility of the
experiments?",
    "Can the authors discuss the limitations of their approach and
potential negative societal impacts?",
    "Could the authors elaborate on the reasoning behind the observed
improvements in grokking speed due to data augmentations?",
    "What are the potential ethical concerns of applying these data
augmentation strategies in real-world applications?",
    "Can the authors include more ablation studies to dissect the
individual contributions of each augmentation technique in greater
detail?",
    "How do the results generalize to other neural network architectures or
more complex tasks beyond modular arithmetic?"
],
"Limitations": [
    "The paper's clarity and thoroughness in discussing methodology and
results need improvement.",
    "The generalizability of the findings to other domains and
architectures requires further exploration.",
    "The study acknowledges the sensitivity of results to hyperparameters
and task specificity. However, it should also consider the broader
applicability and potential limitations in real-world scenarios.",
    "Potential negative societal impacts are not discussed, which is
important for a comprehensive evaluation of the work."
],
"Ethical Concerns": false,
"Soundness": 3,
"Presentation": 3,
"Contribution": 3,
"Overall": 5,
"Confidence": 4,
"Decision": "Reject"
\end{verbatim}}
\end{tcolorbox}

\end{document}

%% file: math_commands.tex
\usepackage{amsmath,amsfonts,bm}

\def\eqref#1{equation~\ref{#1}}

\def\1{\bm{1}}

\DeclareMathAlphabet{\mathsfit}{\encodingdefault}{\sfdefault}{m}{sl}
\SetMathAlphabet{\mathsfit}{bold}{\encodingdefault}{\sfdefault}{bx}{n}

%% file: case_study_viz/grokking_ideas.tex
\begin{tcolorbox}[breakable,colback=blue!5!white, colframe=blue!75!black, title=Seed Idea - \texttt{batch\_size\_grokking}]
\small
\begin{verbatim}
"Name": "batch_size_grokking",
"Title": "Batch Size Grokking: Assessing the impact of the training
batchsize on the grokking phenomenon",
"Experiment": "Modify the experiments to dynamically adjust the batch size
during training, starting with a small batch size and gradually increasing
it. This could potentially lead to faster generalization on the validation
set.",
"Interestingness": 6,
"Feasibility": 4,
"Novelty": 4,
"novel": true
\end{verbatim}
\end{tcolorbox}

\begin{tcolorbox}[breakable,colback=blue!5!white, colframe=blue!75!black, title=Idea 1/50 - \texttt{model\_size\_grokking}]
\small
\begin{verbatim}
"Name": "model_size_grokking",
"Title": "Investigating the Impact of Model Size on the Grokking
Phenomenon",
"Experiment": "Modify the Transformer class to accept variable number of
layers and dimension sizes. Test models with 1, 2, 4, and 8 layers, and
dimension sizes of 64, 128, 256, and 512. For each dataset and model size,
track the step at which grokking occurs (defined as validation accuracy
exceeding 99%
and final accuracy vs model size for each task.",
"Interestingness": 8,
"Feasibility": 7,
"Novelty": 7,
"novel": true
\end{verbatim}
\end{tcolorbox}

\begin{tcolorbox}[breakable,colback=blue!5!white, colframe=blue!75!black, title=Idea 2/50 - \texttt{optimizer\_grokking}]
\small
\begin{verbatim}
"Name": "optimizer_grokking",
"Title": "Optimization Dynamics and Grokking: Comparing SGD and Adam with
Different Learning Rate Schedules",
"Experiment": "Modify the training loop to support two optimizers (SGD,
Adam) and two learning rate schedules (constant, cosine annealing). For
each combination, run multiple experiments with different random seeds.
Track validation accuracy, training loss, and L2 norm of weight updates
throughout training. Compare the timing and extent of grokking across these
optimization strategies for each dataset. Analyze how different
optimization dynamics correlate with grokking behavior, including
statistical analysis of the results.",
"Interestingness": 9,
"Feasibility": 8,
"Novelty": 8,
"novel": true
\end{verbatim}
\end{tcolorbox}

\begin{tcolorbox}[breakable,colback=blue!5!white, colframe=blue!75!black, title=Idea 3/50 - \texttt{biased\_data\_grokking}]
\small
\begin{verbatim}
"Name": "biased_data_grokking",
"Title": "Grokking Under Biased Data: The Effect of Input Range Bias on
Neural Network Generalization",
"Experiment": "Modify the fetch_train_example method in AbstractDataset to
introduce a simple bias: favoring lower-valued inputs. For modular
arithmetic operations, sample 70%
of the input range. For permutations, favor permutations with more elements
in their original positions. Keep the validation set unbiased. Run
experiments comparing grokking behavior on biased vs. unbiased training
sets. Track metrics such as steps to 99%
validation accuracy, and training loss. Analyze how this bias affects
grokking across different operations.",
"Interestingness": 8,
"Feasibility": 8,
"Novelty": 8,
"novel": true
\end{verbatim}
\end{tcolorbox}

\begin{tcolorbox}[breakable,colback=blue!5!white, colframe=blue!75!black, title=Idea 4/50 - \texttt{adaptive\_noise\_grokking}]
\small
\begin{verbatim}
"Name": "adaptive_noise_grokking",
"Title": "Adaptive Noise in Grokking: Investigating Input Perturbations on
Algorithmic Learning and Representations",
"Experiment": "Modify the GroupDataset class to add operation-specific
noise during training: (1) For modular arithmetic, add small integer
perturbations. (2) For permutations, occasionally swap two elements.
Implement three noise levels (low, medium, high) for each operation.
Compare grokking behavior across noise levels and operations, tracking
steps to 99%
loss. Analyze learned representations by visualizing attention patterns and
performing principal component analysis (PCA) on hidden states at different
training stages. Compare these representations between noisy and non-noisy
training to understand how noise affects the abstraction of concepts.",
"Interestingness": 9,
"Feasibility": 8,
"Novelty": 9,
"novel": true
\end{verbatim}
\end{tcolorbox}

\begin{tcolorbox}[breakable,colback=blue!5!white, colframe=blue!75!black, title=Idea 5/50 - \texttt{attention\_evolution\_grokking}]
\small
\begin{verbatim}
"Name": "attention_evolution_grokking",
"Title": "Attention Evolution in Grokking: Quantifying the Transition from
Memorization to Generalization",
"Experiment": "Modify the Transformer class to output attention weights.
Extract and store attention weights at key checkpoints: start, mid-
training, grokking point (99%
visualization tools for attention heatmaps and create plots showing
attention evolution over time. Calculate the Frobenius norm of the
difference between attention matrices at consecutive checkpoints to
quantify attention evolution. Compare attention patterns and evolution
metrics across different operations (modular arithmetic vs. permutations).
Analyze attention for specific, informative input sequences to enhance
interpretability. Correlate attention evolution metrics with validation
accuracy and generalization performance.",
"Interestingness": 9,
"Feasibility": 8,
"Novelty": 8,
"novel": true
\end{verbatim}
\end{tcolorbox}

\begin{tcolorbox}[breakable,colback=blue!5!white, colframe=blue!75!black, title=Idea 6/50 - \texttt{local\_vs\_global\_attention\_grokking}]
\small
\begin{verbatim}
"Name": "local_vs_global_attention_grokking",
"Title": "Local vs Global Attention: Investigating the Impact of Attention
Scope on Grokking in Algorithmic Learning",
"Experiment": "Modify the DecoderBlock class to support two attention
mechanisms: full (global) attention and local attention with a fixed window
size. Implement these variants and run experiments across all datasets.
Track metrics including time to grokking (99%
validation accuracy, and training loss. Calculate and compare 'attention
entropy' for both mechanisms across tasks to quantify attention focus.
Analyze how the scope of attention (local vs global) affects grokking
behavior and final performance for different algorithmic tasks.",
"Interestingness": 9,
"Feasibility": 8,
"Novelty": 8,
"novel": true
\end{verbatim}
\end{tcolorbox}

\begin{tcolorbox}[breakable,colback=blue!5!white, colframe=blue!75!black, title=Idea 7/50 - \texttt{input\_encoding\_grokking}]
\small
\begin{verbatim}
"Name": "input_encoding_grokking",
"Title": "Binary vs One-Hot Encoding: Impact on Grokking in Algorithmic
Learning Tasks",
"Experiment": "Modify the AbstractDataset class to support two encoding
schemes: one-hot (current) and binary. Implement binary encoding for
modular arithmetic operations using log2(p) bits, and for permutations
using ceil(log2(k!)) bits to represent each permutation uniquely. Adjust
the Transformer class to accommodate different input sizes. Run experiments
for each encoding scheme across all datasets, tracking metrics such as time
to grokking (99%
loss, and model memory usage. Analyze how different encoding schemes affect
grokking behavior, convergence speed, and final performance for various
algorithmic tasks. Compare the impact of input representations on the
model's ability to learn and generalize across different operations.
Discuss how findings could inform input representation choices in other
machine learning tasks beyond algorithmic learning.",
"Interestingness": 9,
"Feasibility": 8,
"Novelty": 8,
"novel": true
\end{verbatim}
\end{tcolorbox}

\begin{tcolorbox}[breakable,colback=blue!5!white, colframe=blue!75!black, title=Idea 8/50 - \texttt{curriculum\_learning\_grokking}]
\small
\begin{verbatim}
"Name": "curriculum_learning_grokking",
"Title": "Curriculum Learning in Grokking: The Effect of Structured Example
Progression on Algorithmic Learning",
"Experiment": "Modify the AbstractDataset class to implement a simple
curriculum learning strategy. For modular arithmetic operations, start with
operations involving numbers in the lower half of the range and gradually
introduce larger numbers. For permutations, begin with permutations that
differ from the identity by one swap and progressively increase the number
of swaps. Implement a curriculum scheduler that increases difficulty every
500 steps. Run experiments comparing standard random sampling vs.
curriculum learning across all datasets. Track metrics including time to
grokking (99%
loss. Plot learning trajectories (validation accuracy over time) for both
approaches. Compare attention patterns between curriculum and random
approaches at different stages of training. Analyze how the curriculum
affects the grokking phenomenon across different operations and discuss
implications for training neural networks on algorithmic tasks.",
"Interestingness": 9,
"Feasibility": 8,
"Novelty": 8,
"novel": true
\end{verbatim}
\end{tcolorbox}

\begin{tcolorbox}[breakable,colback=blue!5!white, colframe=blue!75!black, title=Idea 9/50 - \texttt{weight\_init\_grokking}]
\small
\begin{verbatim}
"Name": "weight_init_grokking",
"Title": "Weight Initialization Strategies and Their Impact on Grokking in
Algorithmic Learning",
"Experiment": "Modify the Transformer class to support three weight
initialization strategies: Xavier/Glorot, Kaiming/He, and random normal (as
baseline). Implement these initialization methods for linear layers and
embeddings. Run experiments across all datasets for each initialization
strategy. Track metrics including time to grokking (99%
accuracy), final validation accuracy, training loss, and gradient norm
during training. Plot learning curves and compare the distribution of
weight values at different stages of training. Analyze the loss landscape
by computing gradient variance as a proxy for local geometry at key points
during training. Compare how different initialization strategies affect the
grokking phenomenon, convergence speed, and final performance across
various algorithmic tasks. Investigate potential correlations between
initial weight distributions, gradient variance characteristics, and the
timing/nature of the grokking transition.",
"Interestingness": 9,
"Feasibility": 9,
"Novelty": 8,
"novel": true
\end{verbatim}
\end{tcolorbox}

\begin{tcolorbox}[breakable,colback=blue!5!white, colframe=blue!75!black, title=Idea 10/50 - \texttt{task\_complexity\_grokking}]
\small
\begin{verbatim}
"Name": "task_complexity_grokking",
"Title": "Grokking Across Task Complexity: Mapping Neural Network Learning
Dynamics to Algorithmic Difficulty",
"Experiment": "1. Modify the AbstractDataset class to include new
operations of increasing complexity: modular addition, subtraction,
multiplication, and exponentiation. 2. Implement these operations in new
dataset classes. 3. Quantify task complexity using metrics like algebraic
degree and average solution time for humans (estimated). 4. Run experiments
for each operation, tracking metrics such as time to grokking (99%
validation accuracy), final validation accuracy, training loss, and a new
'complexity-adjusted learning rate' (validation accuracy improvement per
epoch, normalized by task complexity). 5. Plot learning curves and
complexity-adjusted learning rates for each operation. 6. Analyze attention
patterns and hidden state representations at different stages of training
for each operation. 7. Investigate correlations between quantified task
complexity and grokking characteristics (e.g., time to grokking, steepness
of accuracy improvement).",
"Interestingness": 9,
"Feasibility": 8,
"Novelty": 8,
"novel": true
\end{verbatim}
\end{tcolorbox}

\begin{tcolorbox}[breakable,colback=blue!5!white, colframe=blue!75!black, title=Idea 11/50 - \texttt{regularization\_grokking}]
\small
\begin{verbatim}
"Name": "regularization_grokking",
"Title": "The Role of Regularization in Grokking: How L2 and Label
Smoothing Affect Algorithmic Learning",
"Experiment": "1. Implement L2 regularization by adding weight decay to the
optimizer. 2. Implement label smoothing in the loss function. 3. Modify the
training function to support these regularization techniques with two
strength levels each (low and high). 4. Run experiments for each
regularization technique and strength across all datasets, including a
baseline without regularization. 5. Track metrics: time to grokking (99%
validation accuracy), final validation accuracy, training loss, and a new
'grokking speed' metric (rate of validation accuracy improvement from 50%
to 90%
for different regularization settings. 7. Analyze how L2 regularization and
label smoothing affect the timing, speed, and nature of grokking across
various algorithmic tasks, comparing against the non-regularized
baseline.",
"Interestingness": 9,
"Feasibility": 9,
"Novelty": 8,
"novel": true
\end{verbatim}
\end{tcolorbox}

\begin{tcolorbox}[breakable,colback=blue!5!white, colframe=blue!75!black, title=Idea 12/50 - \texttt{grokking\_extrapolation}]
\small
\begin{verbatim}
"Name": "grokking_extrapolation",
"Title": "Grokking and Extrapolation: Investigating the Limits of
Algorithmic Understanding",
"Experiment": "1. Modify AbstractDataset to create a separate test set with
out-of-distribution examples (e.g., larger numbers for modular arithmetic,
longer permutations). 2. Implement a new evaluation function for the test
set. 3. During training, periodically evaluate on both validation and test
sets. 4. Track metrics: time to grokking on validation set, final
validation accuracy, test set accuracy at grokking point, final test set
accuracy, and 'extrapolation gap'. 5. Implement visualization of test set
performance and extrapolation gap over time, highlighting the grokking
point. 6. Compare extrapolation capabilities across different operations
and model sizes. 7. Analyze attention patterns on test set inputs before
and after grokking. 8. Implement a simple MLP baseline for comparison.",
"Interestingness": 9,
"Feasibility": 8,
"Novelty": 9,
"novel": true
\end{verbatim}
\end{tcolorbox}

\begin{tcolorbox}[breakable,colback=blue!5!white, colframe=blue!75!black, title=Idea 13/50 - \texttt{label\_noise\_grokking}]
\small
\begin{verbatim}
"Name": "label_noise_grokking",
"Title": "Grokking Under Noise: The Impact of Systematic and Random Label
Errors on Algorithmic Learning",
"Experiment": "1. Modify the AbstractDataset class to introduce two types
of label noise: random (labels changed randomly) and systematic (specific
labels consistently flipped). Add a 'noise_type' parameter
(random/systematic) and 'noise_level' parameter (low: 5%
high: 20%
training set while keeping the validation set clean. 3. Run experiments for
each noise type and level across all datasets. 4. Track metrics: time to
grokking (99%
loss, and model confidence (mean softmax probability of correct class). 5.
Plot learning curves and model confidence for different noise types and
levels, highlighting the grokking point for each. 6. Analyze how different
types and levels of label noise affect the timing, speed, and extent of
grokking across different operations. 7. Compare attention patterns between
noisy and clean training at different stages to understand how the model
adapts to noise.",
"Interestingness": 9,
"Feasibility": 8,
"Novelty": 9,
"novel": true
\end{verbatim}
\end{tcolorbox}

\begin{tcolorbox}[breakable,colback=blue!5!white, colframe=blue!75!black, title=Idea 14/50 - \texttt{compositional\_grokking}]
\small
\begin{verbatim}
"Name": "compositional_grokking",
"Title": "Compositional Grokking: Investigating the Relationship Between
Grokking and Compositional Learning in Modular Arithmetic",
"Experiment": "1. Modify ModSumDataset and ModSubtractDataset to include
composite operations: (a + b) - c mod p and (a - b) + c mod p. 2. Implement
new dataset class CompositeModDataset for these operations. 3. Run
experiments comparing learning curves for basic (a + b, a - b) and
composite operations. 4. Track metrics: time to grokking for basic vs.
composite operations, correlation between grokking times, final accuracies.
5. Analyze attention patterns to see if the model learns to attend to
intermediate results in composite operations. 6. Implement a 'compositional
generalization' test by training on basic operations and testing on their
compositions. 7. Compare internal representations (e.g., using PCA on
hidden states) for basic vs. composite operations at different stages of
training.",
"Interestingness": 9,
"Feasibility": 6,
"Novelty": 9,
"novel": true
\end{verbatim}
\end{tcolorbox}

\begin{tcolorbox}[breakable,colback=blue!5!white, colframe=blue!75!black, title=Idea 15/50 - \texttt{mutual\_information\_grokking}]
\small
\begin{verbatim}
"Name": "mutual_information_grokking",
"Title": "Information Dynamics in Grokking: Analyzing Mutual Information
Evolution During Algorithmic Learning",
"Experiment": "1. Implement a function to estimate mutual information using
a binning approach for efficiency. 2. Modify the Transformer class to
output hidden states from the final layer. 3. Update the training loop to
calculate and store mutual information between (a) inputs and outputs, and
(b) final hidden states and outputs, at regular intervals. 4. Run
experiments across all datasets, tracking these mutual information metrics
alongside validation accuracy and training loss. 5. Create plots showing
the evolution of both mutual information metrics over training time,
highlighting the grokking point. 6. Analyze how mutual information trends
relate to grokking by testing specific hypotheses: (a) Rapid increase in
hidden state-output mutual information coincides with grokking, (b) Input-
output mutual information stabilizes post-grokking. 7. Compare mutual
information dynamics between different operations and model sizes to
identify common patterns in successful grokking.",
"Interestingness": 9,
"Feasibility": 6,
"Novelty": 9,
"novel": true
\end{verbatim}
\end{tcolorbox}

\begin{tcolorbox}[breakable,colback=blue!5!white, colframe=blue!75!black, title=Idea 16/50 - \texttt{sparse\_subnetworks\_grokking}]
\small
\begin{verbatim}
"Name": "sparse_subnetworks_grokking",
"Title": "Sparse Subnetworks in Grokking: Investigating the Emergence of
Critical Structures During Algorithmic Learning",
"Experiment": "1. Implement a simple magnitude-based pruning function for
the Transformer model. 2. Modify the training loop to perform pruning at
key points: before training, just before grokking (based on validation
accuracy), and after grokking. 3. For each pruning point, create sparse
networks at different sparsity levels (e.g., 50%
these sparse networks from the original initialization for a fixed number
of steps. 5. Track metrics: validation accuracy of sparse networks,
sparsity level, and grokking speed (if it occurs). 6. Plot the performance
of sparse networks at different sparsity levels and pruning points. 7.
Compare the structure and performance of sparse networks found before and
after grokking across different operations.",
"Interestingness": 9,
"Feasibility": 8,
"Novelty": 9,
"novel": true
\end{verbatim}
\end{tcolorbox}

\begin{tcolorbox}[breakable,colback=blue!5!white, colframe=blue!75!black, title=Idea 17/50 - \texttt{positional\_encoding\_grokking}]
\small
\begin{verbatim}
"Name": "positional_encoding_grokking",
"Title": "Inductive Biases in Grokking: The Impact of Positional Encoding
Schemes on Algorithmic Learning",
"Experiment": "1. Modify the Transformer class to support three positional
encoding schemes: sinusoidal (current), learned embeddings, and a simple
binary encoding (e.g., [0,1,0,1,0] for 'a o b = c'). 2. Implement these
encoding schemes, ensuring they work with the existing sequence length. 3.
Run experiments for each encoding scheme across all datasets, tracking:
time to grokking (99%
training loss, and attention entropy. 4. Analyze how different encoding
schemes affect attention patterns and grokking behavior for each operation
type. 5. Compare generalization capabilities on sequences with shuffled
operands (e.g., 'b o a = c'). 6. Correlate encoding scheme performance with
operation complexity to identify potential interactions between input
representation and task structure. 7. Discuss implications for designing
transformers for specific algorithmic tasks based on findings.",
"Interestingness": 9,
"Feasibility": 9,
"Novelty": 9,
"novel": true
\end{verbatim}
\end{tcolorbox}

\begin{tcolorbox}[breakable,colback=blue!5!white, colframe=blue!75!black, title=Idea 18/50 - \texttt{adversarial\_robustness\_grokking}]
\small
\begin{verbatim}
"Name": "adversarial_robustness_grokking",
"Title": "Adversarial Robustness During Grokking: Tracking Vulnerability
Evolution in Algorithmic Learning",
"Experiment": "1. Implement a simple perturbation method: randomly flip 1-2
bits in the input representation for modular arithmetic, and swap 1-2
elements for permutations. 2. Modify the training loop to generate
perturbed inputs and evaluate model performance on them every 500 steps. 3.
Track metrics: normal validation accuracy, accuracy on perturbed inputs,
and 'robustness gap' (difference between normal and perturbed accuracy). 4.
Plot the evolution of robustness to perturbations alongside the grokking
curve. 5. Compare robustness before, during, and after grokking across
different operations. 6. Analyze examples of successful perturbations at
different stages of training. 7. Investigate potential correlations between
the timing of grokking and changes in robustness to perturbations.",
"Interestingness": 9,
"Feasibility": 8,
"Novelty": 9,
"novel": true
\end{verbatim}
\end{tcolorbox}

\begin{tcolorbox}[breakable,colback=blue!5!white, colframe=blue!75!black, title=Idea 19/50 - \texttt{critical\_periods\_grokking}]
\small
\begin{verbatim}
"Name": "critical_periods_grokking",
"Title": "Critical Periods in Grokking: The Impact of Timed Learning Rate
Spikes on Algorithmic Learning",
"Experiment": "1. Modify the training loop to support learning rate spikes
at specific points (25%
to apply these spikes, increasing the learning rate by 10x for 100 steps.
3. Run experiments for each spike timing across all datasets (modular
arithmetic and permutations), including a control group with no spikes. 4.
Track metrics: time to grokking, final validation accuracy, and 'spike
impact' (change in validation accuracy slope in 500 steps post-spike). 5.
Plot learning curves highlighting spike points and their impacts. 6.
Analyze how spike timing affects grokking across different operations,
comparing modular arithmetic tasks with permutations. 7. Compare attention
patterns immediately before and after impactful spikes. 8. Correlate spike
impact with the stage of learning (pre-grokking, during grokking, post-
grokking) to identify potential critical periods, assessing whether these
periods are task-specific or general across operations.",
"Interestingness": 9,
"Feasibility": 9,
"Novelty": 9,
"novel": true
\end{verbatim}
\end{tcolorbox}

\begin{tcolorbox}[breakable,colback=blue!5!white, colframe=blue!75!black, title=Idea 20/50 - \texttt{lottery\_tickets\_grokking}]
\small
\begin{verbatim}
"Name": "lottery_tickets_grokking",
"Title": "Lottery Tickets in Grokking: Investigating Sparse Subnetworks
Capable of Algorithmic Learning",
"Experiment": "1. Implement an iterative magnitude pruning function for the
Transformer model. 2. Modify the training loop to support multiple rounds
of train-prune-reset cycles. 3. For each dataset, run experiments with
pruning levels of 30%
iteration, train the network to convergence, prune the specified percentage
of smallest weights, then reset remaining weights to their initial values.
5. Track metrics for each sparse network: time to grokking (or maximum
training time if grokking doesn't occur), final validation accuracy, and
training loss. 6. Introduce a 'grokking efficiency' metric: the ratio of
time to grokking for the sparse network vs. the dense network. For networks
that don't grok, use the maximum training time. 7. Plot learning curves for
each pruning level, highlighting grokking points and grokking efficiency.
8. Compare the structure of sparse networks that achieve grokking across
different operations, focusing on the distribution of preserved weights in
different layers and attention heads. 9. Analyze the correlation between
pruning level and grokking efficiency across different algorithmic tasks,
including cases where grokking fails to occur.",
"Interestingness": 9,
"Feasibility": 8,
"Novelty": 8,
"novel": false
\end{verbatim}
\end{tcolorbox}

\begin{tcolorbox}[breakable,colback=blue!5!white, colframe=blue!75!black, title=Idea 21/50 - \texttt{algebraic\_structure\_grokking}]
\small
\begin{verbatim}
"Name": "algebraic_structure_grokking",
"Title": "Grokking and Algebraic Structure: How Group Properties Influence
Neural Network Learning",
"Experiment": "1. Implement new dataset classes for modular multiplication
and division (modulo p, where p is prime, ensuring proper group
structures). 2. For each operation (addition, multiplication, division),
calculate and store two properties: group order and number of generators.
3. Run experiments for each operation type, tracking: time to grokking,
final validation accuracy, and the two group properties. 4. Plot learning
curves and grokking points for each operation, labeled with their group
properties. 5. Analyze the correlation between group properties and
grokking behavior (e.g., time to grokking, steepness of accuracy
improvement). 6. Compare attention patterns across operations, focusing on
how they reflect the underlying group structure (e.g., uniformity for
commutative operations). 7. Test the model's ability to generalize by
evaluating on compositions of learned operations (e.g., a * b + c mod p)
after training on individual operations.",
"Interestingness": 9,
"Feasibility": 8,
"Novelty": 9,
"novel": true
\end{verbatim}
\end{tcolorbox}

\begin{tcolorbox}[breakable,colback=blue!5!white, colframe=blue!75!black, title=Idea 22/50 - \texttt{mdl\_grokking}]
\small
\begin{verbatim}
"Name": "mdl_grokking",
"Title": "Minimum Description Length and Grokking: Investigating the
Relationship Between Model Compression and Algorithmic Learning",
"Experiment": "1. Implement functions to calculate model complexity: (a) L2
norm of weights, (b) number of bits to store parameters at different
precisions, (c) effective number of parameters using BIC. 2. Modify the
training loop to track these complexity measures alongside existing
metrics. 3. Run experiments across all datasets, recording complexity
measures, validation accuracy, and training loss at regular intervals. 4.
Plot the evolution of model complexity alongside the grokking curve. 5.
Analyze the correlation between sudden decreases in model complexity and
the onset of grokking, including statistical tests for significance. 6.
Compare complexity dynamics across different operations and model sizes. 7.
Visualize weight distributions at pre-grokking, during grokking, and post-
grokking stages. 8. Implement and compare two early stopping mechanisms:
one based on model complexity stabilization and another based on validation
loss stabilization.",
"Interestingness": 9,
"Feasibility": 8,
"Novelty": 9,
"novel": true
\end{verbatim}
\end{tcolorbox}

\begin{tcolorbox}[breakable,colback=blue!5!white, colframe=blue!75!black, title=Idea 23/50 - \texttt{invariance\_learning\_grokking}]
\small
\begin{verbatim}
"Name": "invariance_learning_grokking",
"Title": "Learning Invariances in Grokking: Tracking Symmetry Awareness
During Algorithmic Learning",
"Experiment": "1. Modify AbstractDataset to generate transformed versions
of inputs (cyclic shifts for modular arithmetic, relabelings for
permutations). 2. Update the evaluation function to test model predictions
on both original and transformed inputs. 3. Implement an 'invariance score'
metric: mean absolute difference between predictions on original and
transformed inputs. 4. Modify the training loop to calculate and store the
invariance score at regular intervals. 5. Run experiments across all
datasets, tracking the invariance score alongside existing metrics. 6. Plot
the evolution of the invariance score alongside the grokking curve. 7.
Analyze how the invariance score changes before, during, and after
grokking. 8. Compare invariance learning across different operations and
model sizes. 9. Investigate correlation between invariance score and
generalization performance.",
"Interestingness": 9,
"Feasibility": 8,
"Novelty": 9,
"novel": true
\end{verbatim}
\end{tcolorbox}

\begin{tcolorbox}[breakable,colback=blue!5!white, colframe=blue!75!black, title=Idea 24/50 - \texttt{grokking\_double\_descent}]
\small
\begin{verbatim}
"Name": "grokking_double_descent",
"Title": "Grokking and Double Descent: Exploring the Intersection of Two
Deep Learning Phenomena",
"Experiment": "1. Create a range of model sizes by varying num_layers (1 to
8) and dim_model (32 to 512). 2. For each dataset, train models of
different sizes, tracking validation accuracy, training loss, and time to
grokking (99%
parameters to identify double descent behavior. 4. On the same plot, mark
the point where grokking occurs for each model size. 5. Analyze the
relationship between grokking timing and the different regimes of the
double descent curve (under-parameterized, critical, over-parameterized).
6. Calculate the correlation between model size and time to grokking. 7.
Compare double descent and grokking behavior across different operations
(modular arithmetic vs. permutations). 8. Investigate whether grokking
consistently occurs in a specific regime of the double descent curve.",
"Interestingness": 9,
"Feasibility": 8,
"Novelty": 9,
"novel": false
\end{verbatim}
\end{tcolorbox}

\begin{tcolorbox}[breakable,colback=blue!5!white, colframe=blue!75!black, title=Idea 25/50 - \texttt{ntk\_alignment\_grokking}]
\small
\begin{verbatim}
"Name": "ntk_alignment_grokking",
"Title": "NTK-Output Alignment in Grokking: Tracking Feature Learning
Dynamics in Algorithmic Tasks",
"Experiment": "1. Implement a function to compute the NTK-output alignment:
the cosine similarity between the NTK's top eigenvector and the output
gradient. 2. Modify the training loop to compute and store this alignment
metric every 100 steps. 3. Run experiments across all datasets, tracking
NTK-output alignment alongside validation accuracy and training loss. 4.
Plot the evolution of NTK-output alignment alongside the grokking curve. 5.
Analyze how the alignment changes before, during, and after grokking,
identifying any consistent patterns across different operations. 6.
Investigate correlations between sudden changes in alignment and the onset
of grokking. 7. Compare alignment dynamics for models that achieve grokking
vs. those that don't. 8. Experiment with using the alignment metric as an
early stopping criterion or to adjust learning rates dynamically. 9.
Discuss implications of findings for understanding feature learning and
generalization in grokking.",
"Interestingness": 9,
"Feasibility": 8,
"Novelty": 9,
"novel": true
\end{verbatim}
\end{tcolorbox}

\begin{tcolorbox}[breakable,colback=blue!5!white, colframe=blue!75!black, title=Idea 26/50 - \texttt{loss\_landscape\_grokking}]
\small
\begin{verbatim}
"Name": "loss_landscape_grokking",
"Title": "Loss Landscape Evolution in Grokking: Geometric Insights into
Algorithmic Learning",
"Experiment": "1. Implement functions to compute and visualize 2D loss
landscapes using filter-wise normalization. 2. Modify the training loop to
save model checkpoints at key points: start of training, 25%
just before grokking (based on validation accuracy), during grokking, and
after grokking. 3. For each checkpoint, compute and store 2D loss landscape
visualizations. 4. Define quantitative metrics for loss landscape
characteristics: (a) local smoothness (average gradient magnitude), (b)
global convexity (ratio of loss at edges to center), (c) barrier height
(maximum loss along minimum loss path). 5. Run experiments across all
datasets, generating loss landscapes and computing metrics at key points.
6. Create side-by-side comparisons of loss landscapes at different stages
of training for each operation. 7. Analyze how loss landscape metrics
change before, during, and after grokking. 8. Compare loss landscape
evolution between operations that grok quickly vs. slowly. 9. Investigate
correlations between changes in loss landscape metrics and the onset of
grokking.",
"Interestingness": 9,
"Feasibility": 8,
"Novelty": 8,
"novel": true
\end{verbatim}
\end{tcolorbox}

\begin{tcolorbox}[breakable,colback=blue!5!white, colframe=blue!75!black, title=Idea 27/50 - \texttt{neural\_collapse\_grokking}]
\small
\begin{verbatim}
"Name": "neural_collapse_grokking",
"Title": "Neural Collapse in Grokking: Investigating Feature Geometry
During Algorithmic Learning",
"Experiment": "1. Modify Transformer to output final layer features. 2.
Implement functions to compute class means and covariances. 3. Calculate
simplified neural collapse metrics: (a) average cosine similarity between
class means, (b) ratio of within-class to between-class variances. 4. Track
these metrics every 500 steps during training. 5. Run experiments on
modular arithmetic and permutation datasets. 6. Plot neural collapse
metrics alongside grokking curves. 7. Analyze changes in metrics before,
during, and after grokking. 8. Compare neural collapse dynamics between
operations that grok quickly vs. slowly. 9. Visualize class mean
trajectories in 2D/3D using PCA. 10. Discuss implications for understanding
both grokking and general neural network learning dynamics.",
"Interestingness": 9,
"Feasibility": 6,
"Novelty": 9,
"novel": true
\end{verbatim}
\end{tcolorbox}

\begin{tcolorbox}[breakable,colback=blue!5!white, colframe=blue!75!black, title=Idea 28/50 - \texttt{data\_augmentation\_grokking}]
\small
\begin{verbatim}
"Name": "data_augmentation_grokking",
"Title": "Data Augmentation in Grokking: The Impact of Input
Transformations on Algorithmic Learning",
"Experiment": "1. Implement task-specific augmentations: (a) For modular
arithmetic: add random offsets (mod p) to inputs. (b) For permutations:
apply random permutations to inputs and outputs. 2. Modify GroupDataset to
apply augmentations with 0%
for each augmentation level across all datasets. 4. Track metrics: time to
grokking (99%
'augmentation generalization gap' (difference between augmented and non-
augmented validation accuracy). 5. Plot learning curves and generalization
gaps for each augmentation level. 6. Analyze the correlation between
augmentation level and grokking speed. 7. Compare attention patterns
between augmentation levels to understand representation changes. 8.
Discuss implications for designing data augmentation strategies in
algorithmic learning tasks.",
"Interestingness": 9,
"Feasibility": 9,
"Novelty": 8,
"novel": true
\end{verbatim}
\end{tcolorbox}

\begin{tcolorbox}[breakable,colback=blue!5!white, colframe=blue!75!black, title=Idea 29/50 - \texttt{emergent\_grokking}]
\small
\begin{verbatim}
"Name": "emergent_grokking",
"Title": "Emergent Abilities in Grokking: Investigating Scale-Dependent
Algorithmic Learning",
"Experiment": "1. Modify existing datasets to include 'simple' and
'complex' versions (e.g., mod sum with small vs. large primes). 2. Adjust
Transformer class to scale from tiny (1 layer, 64 dim) to medium (4 layers,
512 dim). 3. For each operation, train models of increasing size, tracking
grokking time and performance on both simple and complex versions. 4.
Implement a generalization test for each operation (e.g., mod sum with even
larger primes). 5. Plot learning curves for different model sizes,
highlighting grokking points. 6. Create heatmaps of model size vs.
operation complexity, showing grokking time and generalization test
results. 7. Perform statistical analysis to identify significant jumps in
performance across model sizes, using metrics such as accuracy increase
rate and time to reach 99%
behavior patterns across different operation types.",
"Interestingness": 9,
"Feasibility": 8,
"Novelty": 9,
"novel": true
\end{verbatim}
\end{tcolorbox}

\begin{tcolorbox}[breakable,colback=blue!5!white, colframe=blue!75!black, title=Idea 30/50 - \texttt{functional\_modularity\_grokking}]
\small
\begin{verbatim}
"Name": "functional_modularity_grokking",
"Title": "Functional Modularity in Grokking: Analyzing Emergent
Specialization in Transformer Networks During Algorithmic Learning",
"Experiment": "1. Implement functions to track weight update patterns and
attention focus for each layer and head. 2. Modify the training loop to
compute and store these metrics at regular intervals. 3. Define a
'functional modularity score' based on the consistency of weight updates
and attention patterns for specific input types. 4. Run experiments across
all datasets, tracking the functional modularity score alongside existing
metrics. 5. Plot the evolution of functional modularity alongside the
grokking curve. 6. Analyze how functional modularity changes before,
during, and after grokking. 7. Visualize the most consistent patterns at
different stages of training and interpret their functions. 8. Compare
functional modularity dynamics between different operations and model
sizes. 9. Investigate correlations between functional modularity and
grokking speed or generalization performance.",
"Interestingness": 9,
"Feasibility": 8,
"Novelty": 9,
"novel": true
\end{verbatim}
\end{tcolorbox}

\begin{tcolorbox}[breakable,colback=blue!5!white, colframe=blue!75!black, title=Idea 31/50 - \texttt{information\_compression\_grokking}]
\small
\begin{verbatim}
"Name": "information_compression_grokking",
"Title": "Information Compression in Grokking: Analyzing Representational
Dynamics During Algorithmic Learning",
"Experiment": "1. Modify Transformer class to include a bottleneck layer
(smaller dimension linear layer) after the encoder. 2. Implement function
to compute activation sparsity (%
bottleneck layer. 3. Update training loop to compute and store activation
sparsity and gradient magnitudes of the bottleneck layer at regular
intervals. 4. Run experiments with different bottleneck sizes (e.g., 25%
50%
to grokking, final validation accuracy, activation sparsity, and gradient
magnitudes. 6. Plot activation sparsity and gradient magnitude evolution
alongside grokking curves for each bottleneck size. 7. Analyze how these
metrics change before, during, and after grokking. 8. Test generalization
by evaluating models on slightly out-of-distribution examples (e.g., larger
numbers in modular arithmetic). 9. Investigate correlation between optimal
compression (measured by activation sparsity) and grokking speed,
generalization performance.",
"Interestingness": 9,
"Feasibility": 8,
"Novelty": 8,
"novel": true
\end{verbatim}
\end{tcolorbox}

\begin{tcolorbox}[breakable,colback=blue!5!white, colframe=blue!75!black, title=Idea 32/50 - \texttt{critical\_learning\_periods\_grokking}]
\small
\begin{verbatim}
"Name": "critical_learning_periods_grokking",
"Title": "Critical Learning Periods in Grokking: Temporal Dynamics of
Algorithmic Understanding",
"Experiment": "1. Modify the training loop to support 'intervention
periods' where learning rate is increased by 5x for 100 steps. 2. Implement
a sliding window intervention strategy, with windows of 500 steps, starting
every 250 steps. 3. Run experiments for each window across all datasets and
three model sizes (small, medium, large), including a control group with no
interventions. 4. Track metrics: time to grokking, final validation
accuracy, and 'intervention impact' (area under the validation accuracy
curve for 500 steps post-intervention). 5. Plot learning curves
highlighting intervention windows and their impacts. 6. Create heatmaps
visualizing intervention impact across time windows and model sizes for
each operation. 7. Analyze how intervention timing affects grokking across
different operations and model sizes. 8. Compare attention patterns
immediately before and after impactful interventions. 9. Investigate
whether certain operations or model sizes have more pronounced critical
periods than others. 10. Discuss implications for curriculum design in
machine learning and potential applications in continual and transfer
learning.",
"Interestingness": 9,
"Feasibility": 7,
"Novelty": 9,
"novel": true
\end{verbatim}
\end{tcolorbox}

\begin{tcolorbox}[breakable,colback=blue!5!white, colframe=blue!75!black, title=Idea 33/50 - \texttt{simplicity\_bias\_grokking}]
\small
\begin{verbatim}
"Name": "simplicity_bias_grokking",
"Title": "Simplicity Bias in Grokking: Analyzing Weight Matrix Complexity
During Algorithmic Learning",
"Experiment": "1. Modify AbstractDataset to include two complexity levels
for each operation (e.g., small vs. large prime for modular arithmetic,
short vs. long permutations). 2. Implement a function to compute the
effective rank of weight matrices using singular value decomposition. 3.
Update the training loop to compute and store the effective rank for each
layer every 500 steps. 4. Run experiments across all datasets and both
complexity levels, tracking effective rank alongside existing metrics. 5.
Plot the evolution of effective rank alongside grokking curves for each
complexity level and operation. 6. Analyze how effective rank changes
before, during, and after grokking, and how this relates to task
complexity. 7. Investigate correlations between effective rank dynamics and
grokking speed or generalization performance. 8. Compare effective rank
patterns across different operations and model sizes. 9. Contrast effective
rank dynamics between operations that grok quickly versus those that grok
slowly or fail to grok. 10. Experiment with using effective rank as an
indicator for the onset of grokking.",
"Interestingness": 9,
"Feasibility": 8,
"Novelty": 9,
"novel": true
\end{verbatim}
\end{tcolorbox}

\begin{tcolorbox}[breakable,colback=blue!5!white, colframe=blue!75!black, title=Idea 34/50 - \texttt{lucky\_initializations\_grokking}]
\small
\begin{verbatim}
"Name": "lucky_initializations_grokking",
"Title": "Lucky Initializations in Grokking: Identifying and Analyzing
Favorable Starting Points for Algorithmic Learning",
"Experiment": "1. Implement a function to generate and store 50 random
initializations for the Transformer model. 2. Modify the training loop to
support training from stored initializations and different learning rates.
3. For each dataset, train models from the 50 initializations with 3
learning rates, tracking 'grokking efficiency' (ratio of validation
accuracy to training steps at 99%
initializations (top 20%
characteristics of lucky initializations: weight distribution statistics,
layerwise norms, and attention pattern initialization. 6. Implement a
function to visualize the loss landscape around initial points using
filter-wise normalization. 7. Compare lucky initializations across
different operations to identify common patterns. 8. Develop a simple
predictor for initialization 'luckiness' based on identified
characteristics. 9. Test transfer of lucky initializations across tasks and
learning rates.",
"Interestingness": 9,
"Feasibility": 9,
"Novelty": 9,
"novel": true
\end{verbatim}
\end{tcolorbox}

\begin{tcolorbox}[breakable,colback=blue!5!white, colframe=blue!75!black, title=Idea 35/50 - \texttt{relative\_attention\_grokking}]
\small
\begin{verbatim}
"Name": "relative_attention_grokking",
"Title": "Relative Positional Attention and Its Impact on Grokking in
Algorithmic Learning",
"Experiment": "1. Modify the DecoderBlock class to support two attention
types: standard (current) and relative positional. 2. Implement relative
positional attention, ensuring it works with the existing sequence length.
3. Update the Transformer class to accept an attention_type parameter. 4.
Run experiments for both attention types across all datasets, tracking:
time to grokking (99%
training loss, grokking transition sharpness (rate of validation accuracy
increase), and post-grokking stability (variance in validation accuracy
after reaching 99%
highlighting grokking points and transition periods. 6. Visualize and
compare attention patterns between the two mechanisms at key stages: pre-
grokking, during grokking transition, and post-grokking. 7. Analyze how
relative positional attention affects grokking behavior, transition
sharpness, and stability for each operation type compared to standard
attention. 8. Investigate correlations between attention type and grokking
speed or post-grokking stability. 9. Discuss implications for designing
transformers for specific algorithmic tasks based on findings.",
"Interestingness": 9,
"Feasibility": 8,
"Novelty": 8,
"novel": true
\end{verbatim}
\end{tcolorbox}

\begin{tcolorbox}[breakable,colback=blue!5!white, colframe=blue!75!black, title=Idea 36/50 - \texttt{grokking\_task\_interference}]
\small
\begin{verbatim}
"Name": "grokking_task_interference",
"Title": "Grokking and Task Interference: Exploring the Stability of
Algorithmic Understanding",
"Experiment": "1. Modify the training loop to support learning two modular
arithmetic operations sequentially (e.g., addition then multiplication). 2.
Implement a task scheduler that switches between tasks at regular
intervals. 3. Create a 'dual-task evaluation' function to assess
performance on both tasks simultaneously. 4. Track metrics: time to
grokking for each task, performance on the first task while learning the
second, and a 'grokking stability' score (maintenance of >95%
task 1 while learning task 2). 5. Run experiments with different task
switching frequencies. 6. Analyze how grokking on one task affects learning
speed and grokking on the subsequent task. 7. Visualize attention patterns
before and after introducing the second task to understand representation
changes. 8. Investigate the correlation between grokking speed on the first
task and stability of that understanding when learning the second task. 9.
Compare results with a baseline of learning both tasks simultaneously to
isolate the effects of sequential learning.",
"Interestingness": 9,
"Feasibility": 8,
"Novelty": 9,
"novel": true
\end{verbatim}
\end{tcolorbox}

\begin{tcolorbox}[breakable,colback=blue!5!white, colframe=blue!75!black, title=Idea 37/50 - \texttt{attention\_inductive\_bias\_grokking}]
\small
\begin{verbatim}
"Name": "attention_inductive_bias_grokking",
"Title": "Inductive Biases in Attention Mechanisms: Their Impact on
Grokking in Algorithmic Learning",
"Experiment": "1. Modify DecoderBlock class to support two attention
mechanisms: standard dot-product and additive (Bahdanau). 2. Implement
these attention mechanisms, ensuring compatibility with existing
architecture. 3. Update Transformer class to accept an attention_type
parameter. 4. Select a subset of most illustrative datasets based on
preliminary experiments. 5. Run experiments for each attention type on
selected datasets, tracking: time to grokking (99%
final validation accuracy, training loss, and 'grokking transition
sharpness' (defined as the maximum rate of validation accuracy increase
over any 500-step window). 6. Implement a simple generalization test using
slightly out-of-distribution examples (e.g., larger numbers for modular
arithmetic). 7. Plot learning curves for each attention type, highlighting
grokking points and transition periods. 8. Analyze how different attention
mechanisms affect grokking behavior, transition sharpness, and
generalization performance for each operation type. 9. Visualize attention
patterns for each mechanism at key stages: pre-grokking, during grokking
transition, and post-grokking. 10. Discuss implications for designing
transformers with appropriate inductive biases for specific types of
algorithmic learning tasks.",
"Interestingness": 9,
"Feasibility": 9,
"Novelty": 9,
"novel": true
\end{verbatim}
\end{tcolorbox}

\begin{tcolorbox}[breakable,colback=blue!5!white, colframe=blue!75!black, title=Idea 38/50 - \texttt{gradient\_dynamics\_grokking}]
\small
\begin{verbatim}
"Name": "gradient_dynamics_grokking",
"Title": "Gradient Dynamics in Grokking: Analyzing Information Flow
Efficiency During Algorithmic Learning",
"Experiment": "1. Modify the training loop to compute gradient statistics
(sparsity and magnitude distribution) for each layer. 2. Implement
functions to calculate gradient sparsity (%
magnitude percentiles. 3. Update training process to store these metrics
every 500 steps. 4. Run experiments across all datasets, tracking gradient
metrics alongside existing performance metrics. 5. Plot the evolution of
gradient sparsity and magnitude distributions alongside grokking curves. 6.
Analyze how gradient dynamics change before, during, and after grokking. 7.
Compare gradient patterns between operations that grok quickly vs. slowly.
8. Investigate correlations between changes in gradient dynamics and
grokking speed or generalization performance. 9. Visualize gradient flow
patterns at key stages: pre-grokking, during grokking transition, and post-
grokking.",
"Interestingness": 9,
"Feasibility": 8,
"Novelty": 8,
"novel": true
\end{verbatim}
\end{tcolorbox}

\begin{tcolorbox}[breakable,colback=blue!5!white, colframe=blue!75!black, title=Idea 39/50 - \texttt{adaptive\_curriculum\_grokking}]
\small
\begin{verbatim}
"Name": "adaptive_curriculum_grokking",
"Title": "Adaptive Curriculum Learning in Grokking: Optimizing Example
Difficulty for Efficient Algorithmic Understanding",
"Experiment": "1. Modify AbstractDataset to include a difficulty scoring
function (e.g., input magnitude for modular arithmetic, cycle length for
permutations). 2. Implement adaptive sampling strategy: start with easiest
20%
level exceeds 90%
tracking difficulty of selected examples. 4. Run experiments comparing
adaptive curriculum, random sampling, and static curriculum (increasing
difficulty linearly) across all datasets. 5. Track metrics: time to
grokking, final validation accuracy, learning trajectory smoothness, and
example difficulty distribution over time. 6. Analyze relationship between
difficulty progression and grokking onset. 7. Visualize learning curves and
difficulty progression for each strategy. 8. Compare consistency and speed
of grokking across different random seeds for each strategy. 9. Analyze
computational efficiency by comparing total number of examples needed to
achieve grokking for each strategy. 10. Compare attention patterns at key
points (pre-grokking, during grokking, post-grokking) across strategies to
understand how adaptive curriculum affects internal representations.",
"Interestingness": 9,
"Feasibility": 8,
"Novelty": 9,
"novel": true
\end{verbatim}
\end{tcolorbox}

\begin{tcolorbox}[breakable,colback=blue!5!white, colframe=blue!75!black, title=Idea 40/50 - \texttt{task\_structure\_grokking}]
\small
\begin{verbatim}
"Name": "task_structure_grokking",
"Title": "Task Structure and Grokking: Investigating the Relationship
Between Algorithmic Complexity and Learning Dynamics",
"Experiment": "1. Modify AbstractDataset to include a
'structural_complexity' score based on: a) number of unique outputs, b)
input-output correlation, c) algebraic degree for modular operations or
cycle structure for permutations. 2. Extend existing dataset classes to
include a wider range of operations (e.g., modular addition,
multiplication, exponentiation; simple and complex permutations). 3. Run
experiments across all operations, tracking time to grokking, final
validation accuracy, and learning curve smoothness. 4. Plot grokking
metrics against structural complexity scores, comparing trends between
modular arithmetic and permutation tasks. 5. Analyze correlation between
structural complexity and grokking behavior. 6. Compare attention patterns
and gradient flows across tasks of different complexity. 7. Implement a
generalization test where models trained on simpler structures are
evaluated on more complex ones. 8. Discuss implications for neural network
learning on structured vs. unstructured tasks in general machine learning
contexts.",
"Interestingness": 9,
"Feasibility": 9,
"Novelty": 9,
"novel": true
\end{verbatim}
\end{tcolorbox}

\begin{tcolorbox}[breakable,colback=blue!5!white, colframe=blue!75!black, title=Idea 41/50 - \texttt{numerical\_base\_grokking}]
\small
\begin{verbatim}
"Name": "numerical_base_grokking",
"Title": "Numerical Base and Grokking: How Input Representation Affects
Pattern Recognition in Algorithmic Learning",
"Experiment": "1. Modify AbstractDataset and modular arithmetic dataset
classes to support binary and decimal bases. 2. Implement functions to
convert between bases and adjust the encode/decode methods. 3. Update the
Transformer class to handle variable input lengths. 4. Run experiments for
binary and decimal bases on modular addition and multiplication tasks. 5.
Track metrics: time to grokking (99%
accuracy, training loss, and 'cross-base generalization' (accuracy when
testing on the other base). 6. Plot learning curves for each base,
highlighting grokking points. 7. Compare learning curves for binary (0-3)
vs decimal (0-9) to isolate base effects from sequence length. 8. Analyze
how different bases affect grokking speed and pattern recognition. 9.
Compare attention patterns across bases at key stages: pre-grokking, during
grokking, and post-grokking. 10. Discuss implications for choosing input
representations in mathematical machine learning tasks.",
"Interestingness": 9,
"Feasibility": 9,
"Novelty": 9,
"novel": true
\end{verbatim}
\end{tcolorbox}

\begin{tcolorbox}[breakable,colback=blue!5!white, colframe=blue!75!black, title=Idea 42/50 - \texttt{activation\_function\_grokking}]
\small
\begin{verbatim}
"Name": "activation_function_grokking",
"Title": "Activation Functions and Grokking: Investigating the Role of Non-
linearity in Algorithmic Learning and Generalization",
"Experiment": "1. Modify the DecoderBlock class to support multiple
activation functions (ReLU, GELU, Tanh). 2. Update the Transformer class to
accept an activation_type parameter, allowing for both uniform and hybrid
activation setups. 3. Run experiments comparing the baseline (GELU) with
ReLU, Tanh, and a hybrid setup (ReLU in lower layers, Tanh in upper layers)
across all datasets. 4. Track metrics: time to grokking (99%
accuracy), final validation accuracy, training loss, 'grokking transition
sharpness', and gradient flow statistics. 5. Plot learning curves for each
activation setup, highlighting grokking points and transition periods. 6.
Visualize decision boundaries at different training stages for each
activation setup. 7. Analyze how different activation functions affect
grokking behavior, transition sharpness, and final performance for each
operation type. 8. Compare hidden representations (using t-SNE) across
activation setups at key stages: pre-grokking, during grokking transition,
and post-grokking. 9. Investigate the relationship between activation
function properties and the trade-off between memorization and
generalization. 10. Discuss implications for choosing activation functions
in tasks requiring pattern discovery and generalization beyond algorithmic
learning.",
"Interestingness": 9,
"Feasibility": 9,
"Novelty": 9,
"novel": true
\end{verbatim}
\end{tcolorbox}

\begin{tcolorbox}[breakable,colback=blue!5!white, colframe=blue!75!black, title=Idea 43/50 - \texttt{phase\_transition\_grokking}]
\small
\begin{verbatim}
"Name": "phase_transition_grokking",
"Title": "Grokking as a Phase Transition: Characterizing Critical Behavior
in Algorithmic Learning",
"Experiment": "1. Implement functions to track key metrics: validation
accuracy, training loss, gradient norm, and weight norm. 2. Modify training
loop to compute and store these metrics every 100 steps. 3. Run experiments
across all datasets, with finer-grained tracking (every 10 steps) around
the suspected grokking point. 4. Implement analysis tools to detect sudden
changes or discontinuities in metrics. 5. Plot all metrics on a single,
multi-axis graph to visualize potential phase transitions. 6. Calculate
susceptibility using fluctuations in validation accuracy near the grokking
point. 7. Analyze scaling behavior of susceptibility to identify critical
exponents, if any. 8. Compare phase transition characteristics across
different operations and model sizes. 9. Investigate whether manipulating
learning rate or gradient clipping can induce or prevent grokking phase
transitions.",
"Interestingness": 9,
"Feasibility": 8,
"Novelty": 9,
"novel": false
\end{verbatim}
\end{tcolorbox}

\begin{tcolorbox}[breakable,colback=blue!5!white, colframe=blue!75!black, title=Idea 44/50 - \texttt{effective\_dimension\_grokking}]
\small
\begin{verbatim}
"Name": "effective_dimension_grokking",
"Title": "Effective Dimension Dynamics in Grokking: Analyzing
Representational Complexity During Algorithmic Learning",
"Experiment": "1. Implement functions to compute the rank and top-k
singular values of weight matrices. 2. Modify the training loop to compute
and store these metrics every 500 steps for each layer. 3. Run experiments
across all datasets, tracking rank and singular value distributions
alongside existing performance metrics. 4. Implement a simple MLP baseline
that doesn't exhibit grokking for comparison. 5. Plot the evolution of rank
and singular value distributions alongside grokking curves for both
Transformer and MLP models. 6. Analyze how these metrics change before,
during, and after grokking in the Transformer, contrasting with the MLP. 7.
Compare rank dynamics between operations that grok quickly vs. slowly. 8.
Investigate correlations between changes in rank/singular values and
grokking speed or generalization performance. 9. Visualize the relationship
between these metrics and other performance indicators at different stages
of training.",
"Interestingness": 9,
"Feasibility": 8,
"Novelty": 9,
"novel": true
\end{verbatim}
\end{tcolorbox}

\begin{tcolorbox}[breakable,colback=blue!5!white, colframe=blue!75!black, title=Idea 45/50 - \texttt{representation\_entropy\_grokking}]
\small
\begin{verbatim}
"Name": "representation_entropy_grokking",
"Title": "Representation Entropy in Grokking: Tracking the Simplification
of Learned Concepts",
"Experiment": "1. Implement a function to compute the entropy of the
model's internal representations. 2. Modify the Transformer class to output
intermediate representations. 3. Update the training loop to compute and
store the representation entropy every 500 steps. 4. Run experiments across
all datasets, including configurations that lead to successful grokking and
those that don't (e.g., by varying model size or learning rate). 5. Track
entropy alongside existing performance metrics. 6. Plot the evolution of
representation entropy alongside grokking curves for both successful and
unsuccessful cases. 7. Analyze how representation entropy changes before,
during, and after grokking in successful cases, and compare with
unsuccessful cases. 8. Investigate correlations between changes in
representation entropy and grokking speed or generalization performance. 9.
Visualize the relationship between entropy and other performance indicators
at different stages of training. 10. Plot entropy distributions across
different layers of the model to understand how different parts contribute
to concept simplification.",
"Interestingness": 9,
"Feasibility": 8,
"Novelty": 8,
"novel": true
\end{verbatim}
\end{tcolorbox}

\begin{tcolorbox}[breakable,colback=blue!5!white, colframe=blue!75!black, title=Idea 46/50 - \texttt{mutual\_information\_grokking}]
\small
\begin{verbatim}
"Name": "mutual_information_grokking",
"Title": "Mutual Information Dynamics in Grokking: Tracing Information Flow
During Algorithmic Learning",
"Experiment": "1. Modify Transformer class to output representations from
input embedding, middle layer, and final layer. 2. Implement MINE (Mutual
Information Neural Estimation) for efficient mutual information
approximation. 3. Update training loop to compute and store mutual
information estimates between input-middle, input-output, and middle-output
every 500 steps. 4. Run experiments across all datasets, tracking mutual
information alongside existing performance metrics. 5. Plot the evolution
of mutual information alongside grokking curves and generalization gap. 6.
Analyze how mutual information changes before, during, and after grokking,
particularly in relation to the generalization gap. 7. Compare mutual
information dynamics between operations that grok quickly vs. slowly. 8.
Investigate correlations between changes in mutual information and grokking
speed or generalization performance.",
"Interestingness": 9,
"Feasibility": 8,
"Novelty": 8,
"novel": true
\end{verbatim}
\end{tcolorbox}

\begin{tcolorbox}[breakable,colback=blue!5!white, colframe=blue!75!black, title=Idea 47/50 - \texttt{lottery\_tickets\_grokking}]
\small
\begin{verbatim}
"Name": "lottery_tickets_grokking",
"Title": "Lottery Tickets in Grokking: Sparse Subnetworks and Sudden
Generalization",
"Experiment": "1. Implement iterative magnitude pruning for the Transformer
model. 2. Modify training loop for train-prune-reset cycles. 3. For each
dataset, run experiments with pruning levels of 50%
iterations. 4. Track metrics: time to grokking, final validation accuracy,
training loss, and 'grokking efficiency' (ratio of time to grokking for
sparse vs. dense network). 5. Plot learning curves for each pruning level,
highlighting grokking points. 6. Compare sparse network structures that
achieve grokking across operations. 7. Analyze correlation between pruning
level and grokking efficiency. 8. Implement simple MLP baseline without
grokking for comparison. 9. Visualize weight distributions of winning
tickets pre- and post-grokking.",
"Interestingness": 9,
"Feasibility": 9,
"Novelty": 8,
"novel": false
\end{verbatim}
\end{tcolorbox}

\begin{tcolorbox}[breakable,colback=blue!5!white, colframe=blue!75!black, title=Idea 48/50 - \texttt{architecture\_inductive\_bias\_grokking}]
\small
\begin{verbatim}
"Name": "architecture_inductive_bias_grokking",
"Title": "Architectural Inductive Biases and Grokking: Comparing Sudden
Generalization Across Neural Network Types",
"Experiment": "1. Implement simplified 1D CNN and LSTM model classes
compatible with existing sequence-based datasets. 2. Modify training loop
to support multiple model types. 3. Run experiments comparing Transformer,
1D CNN, and LSTM models across modular arithmetic datasets. 4. Track
metrics: time to grokking, final validation accuracy, training loss, and
architecture-specific indicators (attention patterns for Transformer,
filter activations for CNN, forget gate activations for LSTM). 5. Plot
learning curves for each architecture, highlighting grokking points. 6.
Analyze how different architectures affect grokking behavior, speed, and
final performance for each operation type. 7. Compare internal
representations (using t-SNE) across architectures at key stages: pre-
grokking, during grokking transition, and post-grokking. 8. Investigate the
relationship between architectural inductive biases and the trade-off
between memorization and generalization in modular arithmetic tasks.",
"Interestingness": 9,
"Feasibility": 8,
"Novelty": 8,
"novel": true
\end{verbatim}
\end{tcolorbox}

\begin{tcolorbox}[breakable,colback=blue!5!white, colframe=blue!75!black, title=Idea 49/50 - \texttt{shortcut\_learning\_grokking}]
\small
\begin{verbatim}
"Name": "shortcut_learning_grokking",
"Title": "Shortcut Learning and Grokking: The Interplay Between Surface
Patterns and Deep Understanding in Algorithmic Learning",
"Experiment": "1. Modify AbstractDataset to include operation-specific
shortcuts: for modular arithmetic, make the result always even if the first
operand is even; for permutations, always swap the first two elements. 2.
Implement a function to gradually remove these shortcuts over training by
reducing their frequency. 3. Update the training loop to apply the shortcut
removal function. 4. Add a 'shortcut reliance' metric: the accuracy
difference between shortcut-following and shortcut-violating examples. 5.
Run experiments with varying shortcut removal rates across datasets. 6.
Track metrics: time to grokking, final validation accuracy, shortcut
reliance over time, and performance on a shortcut-free test set. 7. Plot
learning curves and shortcut reliance alongside grokking curves. 8. Analyze
how shortcut presence and removal affect grokking timing and quality. 9.
Compare attention patterns between models trained with and without
shortcuts at key stages.",
"Interestingness": 9,
"Feasibility": 9,
"Novelty": 9,
"novel": true
\end{verbatim}
\end{tcolorbox}

\begin{tcolorbox}[breakable,colback=blue!5!white, colframe=blue!75!black, title=Idea 50/50 - \texttt{grokking\_forgetting\_complexity}]
\small
\begin{verbatim}
"Name": "grokking_forgetting_complexity",
"Title": "Grokking and Forgetting: The Interplay of Task Complexity and
Sudden Generalization in Algorithmic Learning",
"Experiment": "1. Modify ModSumDataset to support multiple complexity
levels (e.g., modular addition with increasing prime moduli). 2. Update the
training loop to gradually introduce higher complexity levels while
continuously evaluating on all levels. 3. Implement a 'multi-complexity
evaluation' function to assess performance across all complexity levels
simultaneously. 4. Track metrics: time to grokking for each complexity
level, performance on lower complexity levels when grokking occurs on a
higher level, and a 'complexity forgetting score' (decrease in accuracy on
lower complexity levels). 5. Analyze the correlation between grokking
events and performance changes on other complexity levels. 6. Compare
internal representations (using cosine similarity of hidden states) across
complexity levels before and after grokking events. 7. Investigate trends
in grokking speed across increasing complexity levels. 8. Plot learning
curves for all complexity levels simultaneously, highlighting grokking
points and potential forgetting events. 9. Visualize the evolution of
representation similarities over time using heatmaps.",
"Interestingness": 9,
"Feasibility": 9,
"Novelty": 9,
"novel": true
\end{verbatim}
\end{tcolorbox}